%% file: main-file.tex
\documentclass[a4paper,fleqn]{cas-sc}
\usepackage{apacite}
\usepackage[authoryear,longnamesfirst]{natbib}
\usepackage{lineno}
\usepackage{subfloat}
\usepackage{subcaption}
\definecolor{shadecolor}{gray}{0.85}
\usepackage{framed}

\usepackage{times,latexsym}
\usepackage{url}
\usepackage[T1]{fontenc}
\usepackage{graphicx}
\usepackage{booktabs}
\usepackage{multirow}
\usepackage{xcolor}
\usepackage{amssymb}
\usepackage{amsmath}

\usepackage[normalem]{ulem}
\useunder{\uline}{\annotationUnderlineFont}{\annotationUnderlineCommand}
\newlength{\annotationdepth}       
\newlength{\annotationdelta}       
\usepackage{hyperref}
\definecolor{darkblue}{rgb}{0, 0, 0.5}
\hypersetup{colorlinks=true,citecolor=darkblue, linkcolor=darkblue, urlcolor=darkblue}

\newcommand{\titleann}[1]{%
\addtolength{\annotationdepth}{-\annotationdelta}%
\raisebox{\annotationdepth}{\tiny title }%
\annotationUnderlineCommand{#1}%
\addtolength{\annotationdepth}{\annotationdelta}%
}

\newcommand{\attitudeann}[1]{%
\addtolength{\annotationdepth}{-\annotationdelta}%
\raisebox{\annotationdepth}{\tiny attitude }%
\annotationUnderlineCommand{#1}%
\addtolength{\annotationdepth}{\annotationdelta}%
}

\newcommand{\moralann}[1]{%
\addtolength{\annotationdepth}{-\annotationdelta}%
\raisebox{\annotationdepth}{\tiny judgments }%
\annotationUnderlineCommand{#1}%
\addtolength{\annotationdepth}{\annotationdelta}%
}

\newcommand{\sentenceann}[1]{%
\addtolength{\annotationdepth}{-\annotationdelta}%
\raisebox{\annotationdepth}{\tiny sentence }%
\annotationUnderlineCommand{#1}%
\addtolength{\annotationdepth}{\annotationdelta}%
}

\definecolor{alizarin}{rgb}{0.82, 0.1, 0.26}
\definecolor{ao}{rgb}{0.0, 0.0, 1.0}
\definecolor{brightlavender}{rgb}{0.75, 0.58, 0.89}

\usepackage{alocal}

\newcommand*{\img}[1]{%
    \raisebox{-.3\baselineskip}{%
        \includegraphics[
        height=\baselineskip,
        width=\baselineskip,
        keepaspectratio,
        ]{#1}%
    }%
}

\begin{document}

\title[mode = title]{Interpretation modeling: Social grounding of sentences by reasoning over their implicit moral judgments}
\shorttitle{Interpretation Modeling}
\shortauthors{L Allein et~al.}

\affiliation{
organization={Department of Computer Science, KU Leuven},
addressline={Celestijnenlaan 200A},
city={Heverlee},
postcode={3001},
country=Belgium}

\author{Liesbeth Allein}[orcid=0000-0002-7776-2156]
\cormark[1]
\ead{liesbeth.allein@kuleuven.be}
\cortext[cor1]{Corresponding author}

\author{Maria Mihaela Truşcǎ}[orcid=0000-0002-9204-0389]

\author{Marie-Francine Moens}[orcid=0000-0002-3732-9323]

\date{}

\begin{abstract}
  The social and implicit nature of human communication ramifies readers' understandings of written sentences. Single gold-standard interpretations rarely exist, challenging conventional assumptions in natural language processing. This work introduces the interpretation modeling (IM) task which involves modeling several interpretations of a sentence's underlying semantics to unearth layers of implicit meaning. To obtain these, IM is guided by multiple annotations of social relation and common ground - in this work approximated by reader attitudes towards the author and their understanding of moral judgments subtly embedded in the sentence. We propose a number of modeling strategies that rely on one-to-one and one-to-many generation methods that take inspiration from the philosophical study of interpretation. A first-of-its-kind IM dataset is curated to support experiments and analyses. The modeling results, coupled with scrutiny of the dataset, underline the challenges of IM as conflicting and complex interpretations are socially plausible. This interplay of diverse readings is affirmed by automated and human evaluations on the generated interpretations. Finally, toxicity analyses in the generated interpretations demonstrate the importance of IM for refining filters of content and assisting content moderators in safeguarding the safety in online discourse. \footnote{The code and dataset will be made publicly available upon acceptance.} 
\end{abstract}


\maketitle

\input{introduction.tex}
\input{dataset.tex}
\input{generation.tex}
\input{experiments.tex}
\input{related_work.tex}
\input{conclusion.tex}
\input{ethics_statement}

\printcredits

\bibliography{bibliography-file}
\bibliographystyle{apacite}

\appendix
\input{appendix.tex}

\end{document}

%% file: introduction.tex
\section{Introduction}

\begin{quote}
    \textit{``Interpretation (...) is the work of thought which consists in deciphering the hidden meaning in the apparent meaning, in unfolding the levels of meaning implied in the literal meaning. (...) There is interpretation wherever there is multiple meaning, and it is in interpretation that the plurality of meaning is made manifest.''} \footnote{\citet{ricoeur1974}, Existence and Hermeneutics, p 13.}
\end{quote}

\begin{figure}[t!]
    \centering
    \includegraphics[width=15cm]{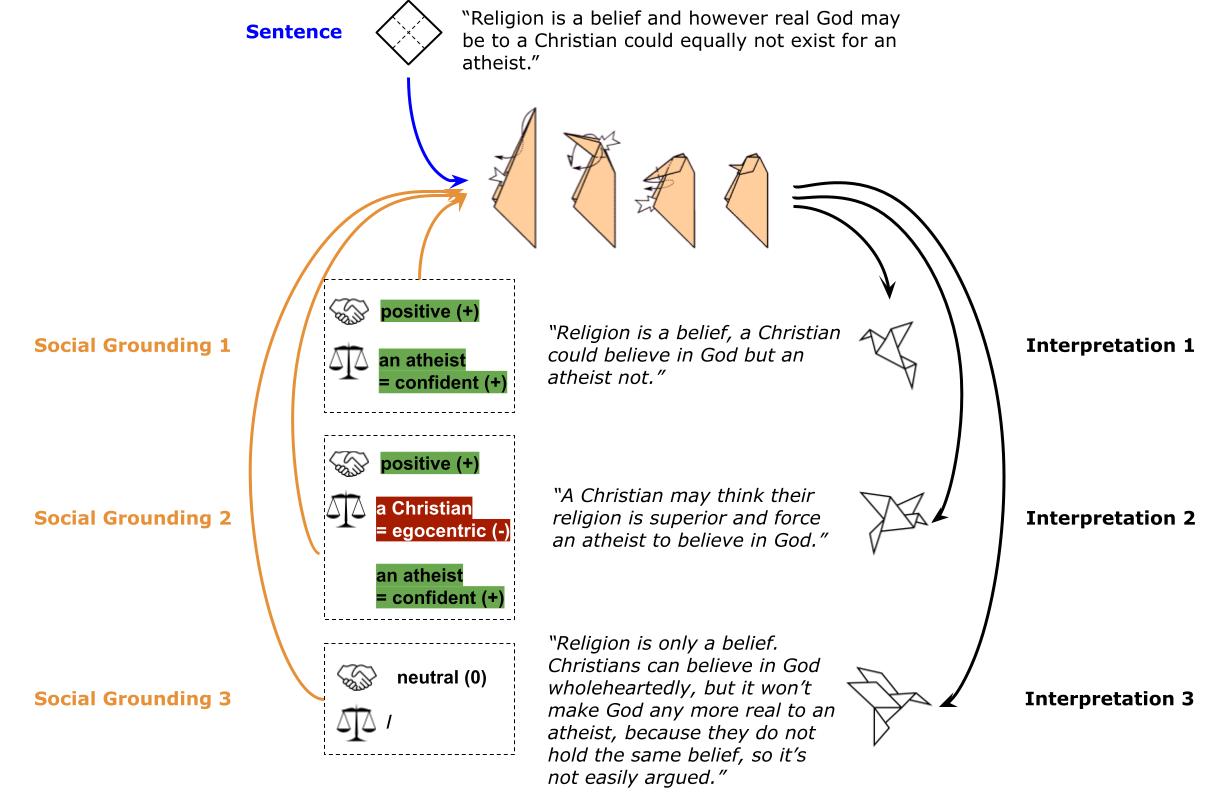}
    \caption{Example taken from the \img{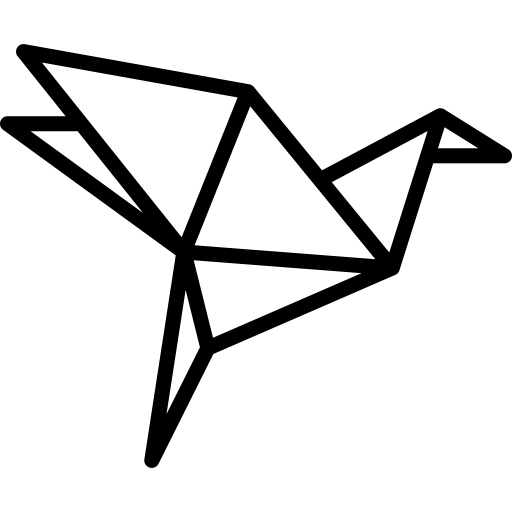} \texttt{origamIM} dataset. A sentence (\img{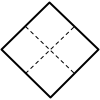}) is translated into various reader interpretations (\img{figures/origami.png}), in this illustration three interpretations. It does so by reasoning over the reactions of multiple readers, in this illustration three readers, sparked upon reading the sentence: the readers' attitudes towards the author explicitly annotated by the readers themselves (\img{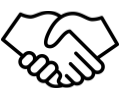}) and the hidden moral judgments about people featured in the sentence that the readers inferred (\img{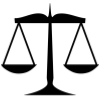}). These reactions represent the social grounding of the sentence.}
    \label{fig:introduction}
\end{figure}

When simulating human understanding of sentences in natural language, artificial intelligence systems need to look beyond the surface and reason about the communication that is happening between the lines, acknowledging that one unambiguous interpretation of a sentence's meaning in natural language rarely exists. Among the root causes of diverse interpretations are properties of the sentence itself, such as lexical, structural, and pragmatic ambiguities \citep{liu2023we}. However, diversity is notably amplified by the unique perspectives of individual readers. Going beyond the conventional exploration of surface-level and contextual ambiguities in natural language understanding, this work examines \textit{ambiguities at the hidden level of a sentence} and models multiple sentence interpretations by grounding sentences in society using various reader understandings of their implicit meanings. 

We propose the \textbf{interpretation modeling ({IM})} task which posits that single ground-truth evaluations and interpretations ignore the complex social reality of natural language understanding. Disagreement in the ground truth is inherent to human understanding \citep{pavlick-kwiatkowski-2019-inherent,nie-etal-2020-learn,uma2021learning}. Yet, most works in natural language processing ({NLP}) still heavily count on single gold standards when processing language. This is shown by a majority of state-of-the-art methods for natural language understanding that rely on large language models fine-tuned on single interpretations or sets of equivalent interpretations. Relying on single ground truths in {IM} would suffer from three main shortcomings that hinder the implicit and social complexities of language comprehension: (i) \textit{Right-or-wrong assumption}: A single interpretation strongly contradicts the ambiguous nature of human communication 
and distills {IM} to be a matter of right or wrong. (ii) \textit{Unilaterial understanding}: The interpretations of single reader groups that guide model optimization leaves the model oblivious of alternative readings in society. (iii) \textit{Partial coverage}: Since multiple meanings can underlie an explicit sentence, it is rather unlikely that single ground-truth interpretations capture all of them -- especially in cases where ambiguity is intentionally crafted, as observed in dogwhistles \citep{henderson2017dogwhistles,mendelsohn2023dogwhistles}. This work seeks to ground language understanding in society (= \textit{social grounding}) by accommodating the intricacies of diverse human interpretations, opening new opportunities for developing more nuanced and socially intelligent language models. To support the proposed {IM} task, we construct a new dataset called \img{figures/origami.png}\texttt{origamIM}\footnote{Origami is the Japanese art of paper folding where a single piece of paper can be transformed to various complex sculptures following different folding steps. Analogously, a sentence may be differently interpreted depending on a reader's understanding of its hidden content.}, in which multiple annotators infer social evaluations hidden in a sentence and describe their interpretation of that sentence (Figure \ref{fig:introduction}). The annotated evaluations then guide interpretation modeling.

Diverse dimensions of underlying meaning contribute to the semantics of a sentence. This work zooms in on implicit judgments of human morality. Moral judgments present a challenging case since people inherently recognize and assess moral behavior differently despite sharing a common understanding of moral norms. The annotations confirm this assumption as readers not only disagree on the type of behavior that is implicitly communicated (e.g., \textit{confident} and \textit{ambitious}) but also differentiate between levels of appropriateness of the same behavior differently (e.g., \textit{over-ambitious} and \textit{properly ambitious}), showing how readers assess morality in their own way.
The annotated moral judgments are grounded in Virtue Ethics \citep{hursthouse1999virtue}, a philosophical framework for describing moral behavior. It establishes a workable framework for computational models to distinguish patterns of similar (un)desirable behaviors throughout various contexts and grounding sentences in society.

We propose a set of {IM} models, parameterized by pre-trained language models, that approach {IM} as a one-to-one or one-to-many generation task. Whereas a sentence is socially grounded by one reader at a time and one interpretation is produced in the one-to-one setting, the one-to-many setting adopts diverse grounding by using annotations on attitude and moral judgments from multiple readers. That way, the model can simultaneously reason over several socially grounded instances for generating multiple interpretations. Social grounding is here established by conditioning the generation on the attitudes and moral judgments together with the input sentence. We also design control mechanisms based on the properties of interpretation described in philosophical studies for guiding the machine learning models to generate interpretations that are semantically diverse. More specifically, we design appropriate loss functions that guide generation of diverse interpretations.
The diversity and validity of the generated interpretations are  
demonstrated by rigorous automatic and human evaluations. 

Lastly, we investigate whether interpretations of the \img{figures/origami.png}\texttt{origamIM} sentences made by humans and interpretations generated by our models help recovering toxic content. Using a standard tool that analyses the level of toxicity, insult, and identity attack expressed in a sentence, we show that both human and automatically generated interpretations reveal underlying toxicity which is not recognized and flagged when only considering the original sentence. 
Our interpretation generation models
capture the different views that people have about real-world problems. Therefore, the research of this paper naturally has applications in several {NLP}-related tasks, such as content moderation, hate speech analysis, and fake news detection \citep{fortuna-etal-2022-directions,allein2023preventing}.

\paragraph{Contributions} 

The main contributions of this article can be summarized as follows:
\begin{itemize}
    \item We introduce \textit{a challenging {NLU} task} called interpretation modeling which involves generating reader interpretations of sentences, centering on their underlying semantics and unearthing layers of implicit meaning.
    \item We curate \textit{a supporting dataset}, \img{figures/origami.png}\texttt{origamIM}, containing sentences featuring people entities where each sentence is annotated multiple times with sentence interpretations and inferences of implicit moral judgments.
    \item We develop \textit{a set of generation frameworks} approaching interpretation modeling as a one-to-one or one-to-many language generation task and propose control mechanisms for enforcing diversity in the generated interpretations. 
    \item We showcase \textit{the importance of interpretation modeling} in content moderation and toxicity detection, which may lead to healthier and safer online environments. 
\end{itemize}

In the remainder of the article, we first discuss the creation of the \img{figures/origami.png} \texttt{origamIM} dataset supporting {IM} (§\ref{dataset-con}) and provide rigorous dataset analyses (§\ref{dataset-an}). Next, we present a number of frameworks in which {IM} is approached as a one-to-one and one-to-many generation task (§\ref{methods}).  
The experiments (§\ref{experiments} and §\ref{results}) indicate that the proposed methods are able to diversify interpretations through social grounding.
This is also confirmed by human evaluations of the generated interpretations (§\ref{human-evaluation}). Those evaluations also show that the interpretations are not merely rewrites but present nuanced meanings. We demonstrate that {IM} plays a pivotal role in building more accurate filters of content (§\ref{toxicity}). We continue to situate our research in the existing body of work (§\ref{related-work}). Finally, the conclusion (§\ref{conclusion}) discusses the limitations and presents how future work can build further on our work.

%% file: dataset.tex
\section{\texttt{origamIM}: Construction} \label{dataset-con}

We describe how we construct a dataset of English sentences (§\ref{datasource}) that mention people (§\ref{annotationround1}), where each sentence is annotated with multiple interpretations and grounded moral judgments (§\ref{annotationround2}).

\subsection{Data Source}\label{datasource}

We automatically crawl blog posts in English from the Subreddit /r/ChangeMyView, dating between 13 July 2020 and 3 March 2022. In this moderated Reddit community, people present their world views, often discussing controversial and polarizing topics, such as abortion and racism, and invite others to present counterarguments. It presents suitable data for {IM} as it hosts a diverse user base, reflects popular and emerging topics of discussion, and allows participants to anonymously express their genuine opinions and respond to those of others. After removing duplicated and deleted blog posts, we extract for each blog post the title, body text, and additional metadata. Since we are interested in sentence interpretation, we split the body text into sentences using SpaCy.

\begin{figure}
    \centering
    \includegraphics[width=11cm]{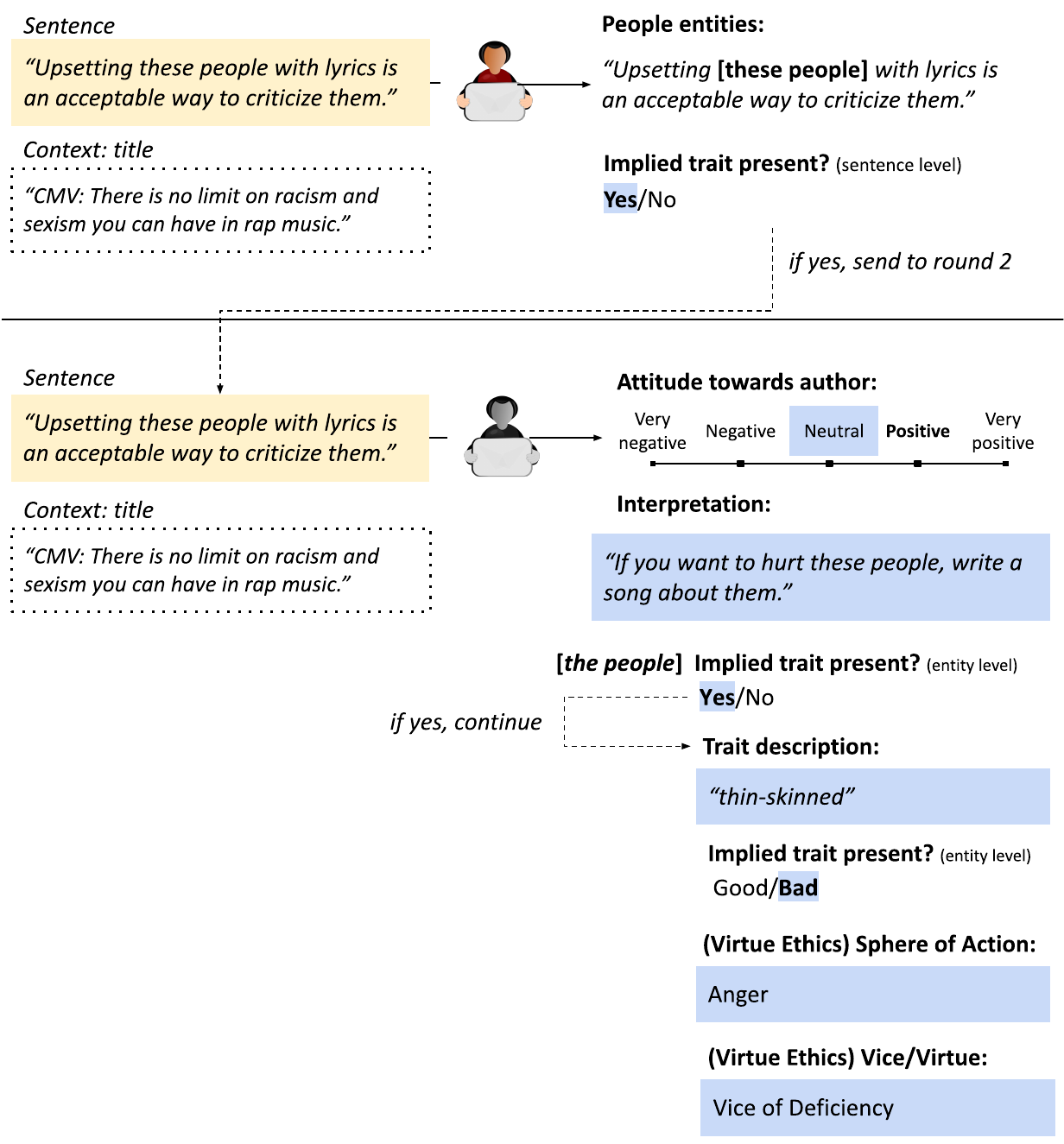}
    \caption{Annotation procedure in two rounds. In the first round, annotators mark the people entities and indicate whether at least one character trait is implied. In the second round,
    annotators indicate their attitude towards the author upon reading the sentence, write their interpretation of the sentence, and describe the implied traits for each people entity.}
    \label{fig:annotation}
\end{figure}

\subsection{Annotation Procedure}

Annotation is performed in two rounds on Amazon Mechanical Turk\footnote{\url{https://www.mturk.com/}} between April 2022 and August 2022\footnote{Since annotation was performed before popular text generation models became publicly and freely accessible, e.g. ChatGPT (30 November 2022) and Bard (21 March 2023), we assume that the sentences were manually annotated \citep{veselovsky2023artificial}.} (Figure \ref{fig:annotation}). 
The complete annotation guidelines containing screenshots of the annotation platform, detailed descriptions of the annotation instructions, definitions of all annotation labels, and a copy of the two qualification tests are enclosed in the Appendix (\ref{app:annotation-procedure}, \ref{app:annotation_round_1}, and \ref{app:annotation-round-2}).
\subsubsection{Round 1: Entities and Presence of Character Traits} \label{annotationround1}

Crowd workers are instructed to mark all unique entities in each sentence referring to people other than the author, if any, and to indicate whether the author implies a character trait about at least one of the entities. A character trait describes a voluntary aspect of a person's behavior or attitude, e.g. \textit{greedy} and \textit{loyal}. Note that the annotators are not required to specify which entity is being judged nor describe the trait at this point. The title of the blog post from which the sentence is taken is provided as additional context. Each sentence is independently annotated by two annotators. In cases where they disagree on the presence/absence of implied traits, a third annotator is consulted and a majority vote is taken. We ensure data quality and annotation consistency by requesting the workers to pass a qualification test before admitting them to the annotation task. The test first describes the main annotation rules for marking people entities (see Appendix \ref{app:annotation_round_1}) and then tests their knowledge of the rules by having them select the correct annotations for multiple sentences. During annotation, the main rules are displayed below the sentence, and detailed rules with illustrative sentences are given in a designated instructions tab. We manually check the submitted annotations for consistency.
A total of 6,820 sentences from 396 blog posts are annotated, of which 2,018 implied a character trait of at least one people entity.
The high portion of 
sentences implying judgments showcases people's tendency to talk about other people and their morality in this Subreddit.

\begin{table}[!]
    \footnotesize
    \centering
    \begin{tabular}{p{3.8cm}p{2.5cm}p{2cm}p{2cm}}
        \toprule
        \textit{Context} & \multicolumn{3}{c}{\textit{Appropriateness}} \\
        \midrule
        \textbf{Sphere of Action ({SoA})} & \textbf{Vice of \ Deficiency} & \textbf{Virtue of Mean} & \textbf{Vice of \ Excess} \\
        \midrule
        \textit{Confidence, fear, uncertainty} & Cowardice & Courage & Rashness \\
        \textit{Pleasures of the body} & Insensibility & Temperance & Profligacy \\
        \textit{Giving \& taking: Small money} & Stinginess & Liberality & Prodigality \\
        \textit{Giving \& taking: Added value} & Meanness & Magnificence & Vulgarity \\
        \textit{Pride, honour as cause} & Little-mindedness & High-mindedness & Vanity \\
        \textit{Ambition, honour as goal} & Lack of ambition & Proper ambition & Over-ambition \\
        \textit{Anger} & Spiritlessness & Gentleness & Wrathfulness \\
        \textit{Pleasure and pain of others} & Cross, contentious & Agreeableness
        & Flattery \\
        \textit{Truth, honesty about oneself} & Irony & Truthfulness & Boastfulness \\
        \textit{Amusing conversation} & Boorishness & Wittiness & Buffoonery \\
        \bottomrule
    \end{tabular}
    \caption{Overview of Spheres of Actions ({SoA}) with their virtue and vices \citep{hursthouse1999virtue}.}
    \label{tab:overview_virtue_ethics}
\end{table}

\subsubsection{Round 2: Interpretations and Judgments} \label{annotationround2}

Each sentence is annotated by at least five different crowd workers. The title of the blog post is again presented as supporting context.

\paragraph{Attitude as social relation} 

The crowd workers first rate their attitude towards the author on a five-point Likert scale ranging from \textit{very negative} (1) to \textit{very positive} (5); \{\textit{very negative}, \textit{negative}, \textit{neutral}, \textit{positive}, \textit{very positive}\}. They do this by setting a slider to the best-fitting attitude. The attitudes reflect reader impressions of the author upon reading the sentence. The workers can consult an overview of the different attitudes and their definitions in a designated instructions tab during annotation.

\paragraph{Sentence interpretation} 

The workers rewrite the given sentence in their own words so that it reflects their interpretation of the sentence and its hidden messages. They are explicitly instructed to not copy the original sentence. To control data quality and annotator fatigue, we manually check the relatedness between sentence-interpretation pairs and remove annotations that present unrelated pairs (i.e., interpretation and sentence do not share any information) or poorly-formulated interpretations (i.e., unfinished sentences and sentences that contain many typos).

\paragraph{Moral judgments as implicit social evaluations} 

For each unique people entity identified in the first annotation round, the workers now indicate whether or not the author implies a character trait. If so, they describe the implied trait, preferably using an adjective (free text), and ground the trait in society by labeling the trait's social evaluation as \textit{good} or \textit{bad} (select from a list) and classifying it in a well-established moral framework called Virtue Ethics \citep{hursthouse1999virtue}.

The moral theory introduced by Aristotle poses that a person's moral character can be evaluated by their voluntary behavior. Those behaviors occur in ten types of context or \textit{Spheres of Action} (\textit{{SoA}}). After a behavior is assigned to its best-fitting {SoA}, it can be situated on an axis of contextual appropriateness ranging from \textit{vice of deficiency} to \textit{virtue of mean} and \textit{vice of excess}. For instance, \textit{vanity} is a \textit{vice of excess} regarding \textit{honour as a cause and pride}. A character trait is thus judged by their social desirability within a context (Table \ref{tab:overview_virtue_ethics}). In contrast to Moral Foundation Theory (MFT) \citep{haidt2004intuitive}, a popular framework for operationalizing moral reasoning \citep{8508244,hoover2020moral,alshomary-etal-2022-moral}, Virtue Ethics does not regard a negative behavior as merely negative but further defines it as a deficient or excessive behavior based on its contextual appropriateness. 
The main advantage of grounding in Virtue Ethics is that it allows readers from different cultural and social backgrounds to annotate their interpretation of the implied moral judgments by their own standards. Since the concept of contextual appropriateness is culturally defined, some societies may find the character of a person virtuous while others may find it vicious.

Prior to starting the second annotation round, the annotators have to pass an instruction test which explains the different aspects of Virtue Ethics. The test first explains the three contextual appropriateness labels (vice of deficiency, virtue of mean, vice of excess) and then goes over all {SoA}s. For each {SoA}, a clear example sentence and a person of interest is given. The workers then have to select the appropriateness label that describes that person's behavior best. Annotators can consult a simplified theory with illustrative sentences at any time during annotation.

\section{\texttt{origamIM}: Analysis} \label{dataset-an}

General dataset statistics are provided in Table \ref{tab:general_statistics}. Analyses and findings are illustrated with examples from the dataset.

\begin{table}[]
    \centering
    \footnotesize
    \begin{tabular}{ll}
        \toprule
        \multicolumn{2}{c}{\textbf{Dataset Statistics}} \\
        \midrule
         \# Sentences & 2,018 \\
         --- Total word count & 44,902 \\
         \# People entities (E) & 3,313 \\
         --- \# Sentences with 1/2/3/4+ E & 1,103 / 661 / 174 / 80 \\
         \# Interpretations & 9,851 \\
         --- Total word count & 155,368 \\
         Distribution attitudes & \\
         -- Very negative & 813 (8.25\%) \\
         -- Negative & 1,971 (20\%) \\
         -- Neutral & 4,302 (43.67\%) \\
         -- Positive & 2,025 (20.56\%) \\
         -- Very positive & 740 (7.51\%) \\
        \bottomrule
    \end{tabular}
    \caption{Statistics of the \img{figures/origami.png} \texttt{origamIM} dataset.}
    \label{tab:general_statistics}
\end{table}

\subsection{Diversity in Interpretation} 

{IM} differs from paraphrasing as it does not consistently maintain the explicit semantics of a source sentence. Interpretations are thus expected to semantically diverge from the source sentence. We measure the lexical and semantic diversity between each sentence-interpretation pair in the dataset. The pairs showcase strong lexical diversity, with {BLEU}-1 \citep{papineni2002bleu}: $\mu = 10.09$. 
BERTScore \citep{bert-score} indicates high semantic similarity: $\mu = .96$ (F1 score). However, note that the high semantic similarity score can be attributed to BERTScore's failure to capture subtle nuances between semantically-related lexical words \citep{hanna-bojar-2021-fine}.
Natural language inference ({NLI})\footnote{\url{https://huggingface.co/ynie/roberta-large-snli_mnli_fever_anli_R1_R2_R3-nli} \citep{nie-etal-2020-adversarial}} presents another view on semantic diversity and relates it to entailment relations between source and target \citep{stasaski-hearst-2022-semantic}. The {NLI} results suggest that about a third of sentence-interpretation pairs present a neutral or even contradiction relation.
An example of the latter 
is the following:
\begin{itemize}
    \item[] \small \textit{Sentence}: Many people in the comments believe that this discrepancy is caused by women being much pickier than men on appearance.
    \item[] \textit{Interpretation}: The writer describes their belief that women are not pickier about partners than men are.
\end{itemize}
It seems from the interpretation that the reader believes that the author of the example sentence disagrees with the people in the comments and therefore thinks the opposite of what is explicitly stated. When taking [title + sentence]-interpretation pairs, the share of entailment relations increases, suggesting that many readers actively leverage the title when making sense of a sentence.

\subsection{Disagreement on Implicit Judgments}
The following example shows how diverse interpretations of the same sentence can be: 
\begin{itemize}
    \item[] \small \textit{Sentence}: And rather than built the country the ANC have used their super majority in parliament to make marxist policies a priority.
    \item[] \textit{Interpretation A}: Author is supporting apartheid.
    \item[] \textit{Interpretation B}: The country has given importance to build up socio-economic policies.
\end{itemize}
We observe that diversification in interpretation already starts with recognizing the presence and absence of moral judgments.
For only 291 sentences (14.42\%), all annotators agree that the author passed at least one judgment, indicating the high degree of disagreement on the presence of implicit moral judgments.

\begin{figure}
    \centering
    \includegraphics[width=8cm]{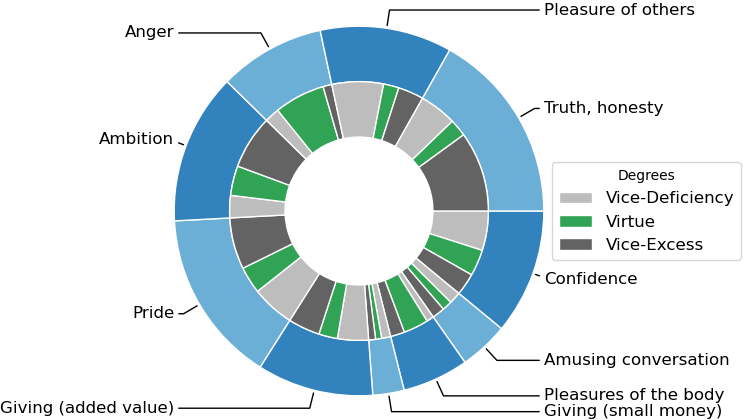}
    \caption{Distribution of Spheres of Action (outer circle) and degrees (inner circle) annotated in the dataset.}
    \label{fig:virtue_ethics_pie_chart}
\end{figure}

When annotators distinguish an implied character trait of a person entity, they show low agreement on its desirability or evaluation in society -- with Krippendorff's $\alpha = .354$ over the annotators’ evaluations of each entity. This means that 
often one annotator perceives a negative judgment of an entity's character while another considers it positive, and vice versa. Even if they agree on its desirability, their interpretation seems to be affected by their attitude towards the author (see Table \ref{tab:examples_dataset}). Figure \ref{fig:virtue_ethics_pie_chart} shows that most recognized character traits are considered vices. An explanation for this phenomenon is not straightforward as it is unclear whether this is due to an actual higher frequency, higher ease of recognition and production, or a cognitive bias towards negativity in both author and reader.

\begin{table*}[ht]
    \centering
    \scriptsize
    \begin{tabular}{p{6cm}p{2cm}p{6cm}}
        \toprule
        \multicolumn{3}{l}{
        \textbf{[Title]} \textit{CMV: If I want to kill myself, no one has the right to force me from doing so/prevent me against my will.}} \\
        \multicolumn{3}{l}{
        \textbf{[Sentence]} \textit{For the exact same reason that \underline{\textbf{a woman}} has the right to have an abortion, she (or anyone) has the right to end her own life}} \\
        \multicolumn{3}{l}{
        \textit{because it is her right.}} \\
        \textbf{[Interpretation]} & \textbf{[Attitude]} & \textbf{[Moral Judgments]} \\
        \textit{Woman has the right to decide everything about their own body and life aspects.} & neutral & \textbf{a woman}: \colorbox{green}{good}, honest, VE: truth, honesty about oneself - Virtue of Mean \\
        \textit{The writer is arguing that one has a right to suicide much as one has a right to abortion, being an issue of bodily autonomy.} & \colorbox{red}{negative} & \textbf{a woman}: / \\
        \textit{Pro-choice means that women have the choice to do whatever they want with their own bodies without interference, even if that means that they want to kill themselves.} & \colorbox{red}{\textbf{very negative}} & \textbf{a woman}: \colorbox{red}{bad}, irrational, VE: giving and taking (added value) - Vice of Excess \\
        \midrule
        \multicolumn{3}{l}{
        \textbf{[Title]} \textit{CMV: I don't see a problem with people valuing to defend their property over an intruders life.}} \\
        \multicolumn{3}{l}{
        \textbf{[Sentence]} \textit{Who knows, maybe \underline{\textbf{she}} is stealing \underline{\textbf{his}} last 1000 dollars that will pay his rent.}} \\
        \textbf{[Interpretation]} & \textbf{[Attitude]} & \textbf{[Moral Judgments]} \\
        \textit{she is taking money which does not belong to her.} & \colorbox{green}{\textbf{very positive}} & \textbf{she:} \colorbox{red}{bad}, greedy woman, VE: giving and taking (money) - Vice of Excess. \\
        & & \textbf{his:} \colorbox{green}{good}, generous, morality: giving and taking (money) - Virtue of Mean \\
        \textit{Perhaps the thief is stealing an individual's last thousand dollars that they needed for rent.} & \colorbox{red}{negative} & \textbf{she:} \colorbox{red}{bad}, dishonest, VE: ambition, honour (goal) - Vice of Deficiency \\
        & & \textbf{his:} \colorbox{green}{good}, innocent, VE: pride, honour (cause) - Virtue of Mean \\
        \textit{We never know who we are dealing with and other people have different problems that we might not be aware of. }& neutral & \textbf{she:} \colorbox{red}{bad}, insensibility, VE: giving and taking (money) - Vice of Deficiency \\
        & & \textbf{his:} / \\
        \bottomrule
    \end{tabular}
    \caption{Two samples from the dataset that illustrate the disagreement existing between readers in terms of moral judgments (top) and attitude (bottom).}
    \label{tab:examples_dataset}
\end{table*}

\subsection{Common Ground and Filling in the Gaps} 
When qualitatively comparing interpretations with original sentences,
we distinguish traces of common ground.
For instance, annotators explain abbreviations;
\begin{itemize}
    \item[] \small \textit{Sentence}: For example it seems like a lot of people just post on Instagram and Facebook about \underline{BLM} but would they be willing to take action in the face of racism.
    \item[] \small \textit{Interpretation}: People might support \underline{Black Lives Matter} and post about it for social media but aren't truly activists if they wouldn't actually take action if they encounter racism.
\end{itemize}
recognize references to social conflicts and polarizing topics;
\begin{itemize}
    \item[] \small \textit{Sentence}: If the woman physically cannot give birth, once again, this should be verified by a judge.
    \item[] \textit{Interpretation}: Women should not have autonomy over their own bodies.
\end{itemize}
and understand rhetorical strategies:
\begin{itemize}
    \item[] \small \textit{Sentence}: Do people even know why Lenin abandoned Socialism?
    \item[] \textit{Interpretation}: Many people don't know why Lenin abandoned Socialism.
\end{itemize}
We also notice that annotators actively reason over seemingly implicit information. They, for example, adhere to their own mental representation of people to fill in social templates:
\begin{itemize}
    \item[] \small \textit{Sentence}: Why that cisgender person doesn't wish to date a trans person or why he broke up.
    \item[] \textit{Interpretation}: Why that cisgender person doesn't want to date a trans person or why he broke up \underline{with her}.
\end{itemize}
They sometimes assign a gender to the author -- even though the author's identity is unknown to them:
\begin{itemize}
    \item[] \small \textit{Sentence}: And I'm not sure if the guy I'm seeing knows that.
    \item[] \textit{Interpretation}: The person is afraid of having to explain to the guy what he doesn't know about \underline{her}.
\end{itemize}
Note that the interpretations could be influenced by the lack of rich context surrounding a given sentence. For instance, annotators may not have felt the need to explain acronyms or refer to social conflicts if more context was available clarifying that information. Nevertheless, further linguistic and rhetorical analyses could identify the governing effects of common ground on interpretation and its reflection in the interpretation's wording but are
beyond the scope of this paper.

%% file: generation.tex
\section{Generation Methods for Interpretation Modeling} \label{methods}

\subsection{Task Definition}

The goal is to generate multiple interpretations $i$ of an input sentence $s$. We assume that a reader or a representative group of readers $j$ each have their own interpretation $i_j$ of $s$. Reader $j$ is characterized by attitude $a_j$ towards the author upon reading $s$ and the implicit moral judgments $m_j$ they distinguish in $s$. The context of reader $j$, which we call $g_j = [a_j, m_j]$, socially grounds $s$ in $i_j$. Social context $g$ can be a priori obtained or inferred jointly with the generation. We assume the former in our experiments below.
Together with $s$ and $g$, the title $ti$ of the blog post from which $s$ is taken is given. We approach {IM} in two autoregressive generation setups. 

\paragraph{One-to-One Generation} 

The objective is generating one interpretation $i_j$ of sentence $s$ by reader $j$ based on $p_\theta(i_j|s,ti,g_j)$. This process is repeated $J$ times for each reader $j$.

\paragraph{One-to-Many Generation} 

$J$ interpretations $i$ are jointly generated conditioned on $s$, $ti$, and social context $g_{1...J}$, that is $p_\theta(i_{1...J}|s,ti,g_{1...J})$ is computed, where $i_{1...J}$ and $g_{1...J}$ respectively represent a long string of $J$ interpretations and social contexts each delimited by a separator token. 
In this set-up, the context of previously generated interpretations is taken into account during generation.

\subsection{Input Feature Representations}\label{inputrep}

$s$, $ti$, and $i$ are variable-length sequences composed of tokens from a vocabulary $V$, $a$ is a scalar following a five-point Likert scale, and $m$ presents $Q$ sequences of moral judgments where $Q$ equals the number of unique people entities \underline{mentioned} in $s$, preserving their order of appearance: $m = ((ent_1, pres_1, desc_1, eval_1, soa_1, vi_1)$,..., $(ent_Q, pres_Q, desc_Q, eval_Q, soa_Q, vi_Q))$ -- with $ent$ the person entity as mentioned in $s$, $pres$ a binary indicator marking the presence/absence of an implied character trait, $desc$ the trait's description using tokens from $V$, $eval$ an evaluation label (i.e., \textit{good} | \textit{bad}), $soa$ a sphere-of-action label, and $vi$ a contextual appropriateness label (i.e., \textit{vice of deficiency} | \textit{virtue of mean} | \textit{vice of excess}). The moral judgment of entity $q$ is then $m_q$. If entity $q$ is not judged, that is $pres_q = 0$, $m_q = (ent_q, 0, \text{``''}, 0, 0, 0)$. We use $Q' \in [0,Q]$ to indicate the number of entities $q'$ \underline{judged} in $s$.

Following the success of prompting with supervised learning \citep{10.1145/3560815}, we apply a prompting function $f_{prompt}(\cdot)$ to jointly represent $s$, $ti$, and $g$. This way, we lay bare the relations between the various input features. The function completes a predefined template that contains four subtemplates (colored boxes), each designated to a specific input type. The subtemplates are joined together by separator token <sep>: 
\begin{quote}
    \textbf{Input template:} \
    \titleann{\colorbox{green}{ Title: $ti$  }} <sep> \attitudeann{\colorbox{pink}{  Attitude: \textbf{[}$str(a)$\textbf{]}$^J$.}} <sep> \moralann{\colorbox{yellow}{\parbox{3.8cm}{Moral Judgments: \textbf{[}\textbf{(}$ent_{q'}$ = $desc_{q'}$, which is a $eval_{q'}$ character trait and a $vi_{q'}$ related to $soa_{q'}$.\textbf{)}$^{Q'}$\textbf{]}$^J$}}} <sep> \sentenceann{\colorbox{orange}{ 
 Sentence: $s$  }}
\end{quote}
In the attitude subtemplate, $str(\cdot)$ maps $a$ from its scalar value to its attitude label (i.e., \{1: \textit{very negative}, 2: \textit{negative}, 3: \textit{neutral}, 4: \textit{positive}, 5: \textit{very positive}\}). In the judgments subtemplate\footnote{In rare cases, one or more variables in $m$ are lacking, then the part between backlashes (//) to which the missing variable belongs is removed from the judgment sentence: $ent_{q'}$ = /$desc_{q'}$, which is/ /a $eval_{q'}$ character trait and/ /a $vi_{q'}$/ /related to $soa_{q'}$/ to maintain a naturally occurring sentence.}, the phrase in parentheses is repeated $Q'$ times. 
If $Q' = 0$, then \textit{``Moral Judgments: None''}. The prompting template is used across all generation settings. 
For aggregating the social context $g_{1...J}$ over $J$ readers, the phrases in squared brackets ([]) in the attitude and judgments subtemplate are repeated $J$ times such that the attitude and moral judgment of reader 
$j$ are aligned by their position in the input template. 
The input template or string is encoded with a pre-trained language model.

\subsection{One-to-One Generation} \label{one2one}
\paragraph{Decoder}
A single interpretation $i_j$ is generated conditioned on $s$, $ti$, and $g_j$; $p_{\theta}(i_j|s,ti,g_j)$. This process is repeated $J$ times for each reader $j$.
\paragraph{Training}

The model parameters are optimized using the standard language modeling objective, that is a negative log-likelihood loss:
\begin{equation}
    \text{$\mathcal{L}_1 = - \sum_{j=1}^{J}\frac{1}{L}\log \prod_{t=1}^{L} p_{\theta}(i_{j,t}|i_{j,<t},s,ti,g_j)$}
\end{equation}
where $L$ refers to the number of tokens in target 
interpretation $i_j$. \textbf{[One2One]}

\subsection{One-to-Many Generation} \label{one2many}

\paragraph{Decoder}
The generated output is a sequence of $J$ interpretations $i_j$ concatenated by a special reader token forming $i_{1...J} = i_1\text{\textit{<reader>}}i_{1+1}...\text{\textit{<reader>}}i_J$. The model is expected to infer $J$ from the $J$ contexts 
given in the input template.
\paragraph{Training}
Model parameters are optimized using a negative log-likelihood loss: 
\begin{equation}
    \text{$\mathcal{L}_m = - \frac{1}{T} \log \prod_{t=1}^{T} p_{\theta}(i_t|i_{<t},s,ti,g_{1...J})$}
\end{equation}
where $T$ is the number of tokens of the target string of the decoder that contains all interpretations. The decoding is conditioned on previously generated interpretations.
Autoregressive decoding is affected by the order in which the interpretations are generated and consequently optimized. For this, we develop control mechanisms derived from the properties of interpretation.

\subsubsection{Random Ordering} During training mode, parameters are optimized using a random ordering of the ground truth interpretations, that is the ordering in which they appear in the dataset. \textbf{[One2M-Rand]}

\subsubsection{Ordering by Semantic Similarity} The ground-truth interpretations in $i_{1...J}$ are ordered by their semantic similarity to $s$. 
This follows the assumption that the more covert or surprising a hidden meaning is, the more complex the reasoning for recovering it and the further its semantic distance from the sentence's apparent semantics are. Low semantic similarity between a sentence and an interpretation then signals high covertness of the hidden meaning while high similarity suggests low covertness.
We approximate covertness using the semantic similarity between $i_j$ and $s$. Low similarity signals high covertness, and vice versa. Therefore, if $sim(i_j, s) > sim(i_{k \neq j}, s)$, the hidden meaning governing $i_k$ is considered more covert than that in $i_j$. Naturally, a ranking $r$ 
: $sim(i_j, s) > sim(i_k, s) \xrightarrow{} r(i_j) > r(i_k)$. Ultimately, the interpretations composing $i_{1...J}$ are ordered in descending order following $r$ during optimization. \textbf{[One2M-Sim]}

\subsubsection{Constraining on Semantic Similarity} In addition 
to the above ordering of ground-truth interpretations in $i_{1...J}$, a decrease in semantic similarity of the generated interpretations in $\hat{i}_{1...J}$ to $s$ is explicitly enforced during training.
The decoded interpretation sequence $\hat{i}_{1...J}$ is split in sentences, where sentence $\hat{i}_j$ presents the predicted interpretation for reader $j$. Next, a language model computes a latent representation of $\hat{i}_{j}$ and $s$, called respectively $i_{j}'$ and $s'$, after which their semantic similarity is computed. Related to previous work on abstractive summarization \citep{liu-etal-2022-brio}, a contrastive loss function $\mathcal{L}_{sim}$ based on a hinge loss enforces a decrease in semantic similarity with $s'$ among the predicted interpretations $i_{1...J}'$:
\begin{equation}
    \mathcal{L}_{sim} = \frac{1}{J}\sum_{j=2}^{J} \ell(s', i_j', i_{j\text{-}1}')
\end{equation}
where
\begin{equation}
    \ell (s', i_j', i_{j\text{-}1}') = \max(0, sim(i_j',s') - sim(i_{j\text{-}1}',s') + m)
\end{equation}

with margin $m$ forcing a substantial difference between the generated interpretations. Semantic similarity $sim (\cdot,\cdot)$ is based on the cosine similarity and takes representations computed by SimCSE \citep{gao-etal-2021-simcse} as it was shown to align best with human evaluation of semantic similarity \citep{wang-etal-2022-just}. The same method for computing $sim(\cdot, \cdot)$ is used in [One2M-Sim].

Model parameters are optimized using a weighted combination of the language modeling and similarity decrease objective: $\mathcal{L^{**}} = \alpha\mathcal{L}_m + (1-\alpha)\mathcal{L}_{sim}$. \textbf{[One2M-Con]}
Losses are summed over $N$ training sentences.

\subsection{Alternative Decoding Models}
There are many possibilities of designing a decoder that generates different interpretations 
during training. We also experimented with models that used context $g_j$ as a prompt to autoregressively decode the interpretation both in the one-to-one and one-to-many settings instead of using $g_j$ as input together with $s$ and $ti$. In the one-to-many case, this leads to a multi-branch decoder \cite{Rebuffel2022} with $J$ parallel decoders where each $j^{th}$ decoder autoregressively generates $i_j$ taking $g_j$ as decoder prompt. Alternatively, we designed an additional loss function that during training enforces the similarity of a generated interpretation with $g_j$. None of these approaches could improve the results, often yielding nonsensical interpretations that disproportionately focused on $g_j$. 

%% file: experiments.tex
\section{Experiments} \label{experiments}

In this section, we describe the experimental setup and the metrics used for evaluating the performance of the proposed generation frameworks on the {IM} task. We then continue to thoroughly investigate the success of the frameworks and the importance of {IM} for content moderation. To that end, we formulate and answer several pertinent research questions in §\ref{results}, §\ref{human-evaluation}, and §\ref{toxicity}.

\subsection{Experimental Setup}
$p_\theta$ is parameterized by large pre-trained language models: decoder-only forward language model {GPT}-2\footnote{{GPT}2-Large; Huggingface implementation of GPT2LMHeadModel.} \citep{radford2019language} and encoder-decoder architecture 
{T5}\footnote{{T5}-Base; Huggingface implementation of T5ForConditionalGeneration.} \citep{raffel2020exploring}. Model parameters are optimized using Adam optimization ($\text{lr} = 1e^{-6}, \text{weight decay} = 0.01, \epsilon = 1e^{-8}, \beta_1 = 0.9, \beta_2 = 0.99$) with batch size = 2 and gradient clipping ($\text{max norm} = 0.1$). During training, early stopping is performed on the validation loss with patience $= 5$.
We adopt diverse beam search \citep{vijayakumar2018diverse} (max. length = 150, $\text{beam size} = 10, \text{beam groups} = 5, \text{diversity penalty} = 2.0, \text{repetition penalty} = 1.2$) with early stopping as the standard decoding method and use the interpretation with the highest probability in the beam as output. In [One2M-Con], we also test with greedy search as decoding method for producing interpretations during training. 
Margin $m = .05$ in $\mathcal{L}_{sim}$ to ensure that the last generated interpretation still related to the original sentence.
Data is split in a train (80\%), test (10\%), and validation (10\%) set in a title-stratified manner ensuring sentences from the same blog are represented in the same set\footnote{Using Scikit-Learn \texttt{GroupKFold}.}. 
We experiment with various values for hyperparameter $\alpha \in [.25, .50, .75]$. 

\subsection{Evaluation Metrics}

We report the lexical overlap between the generated and target interpretations using {BLEU} \citep{papineni2002bleu} and {ROUGE} \citep{lin2004rouge}, and their semantic similarity using MoverScore \citep{zhao-etal-2019-moverscore} and BERTScore \citep{bert-score}. {COMET}\footnote{\url{https://huggingface.co/Unbabel/wmt22-comet-da}} \citep{rei-etal-2020-comet,rei-etal-2022-comet} is a trained metric that projects the sentences, generated interpretations and ground-truth interpretations in a shared feature space, combines their latent representations into a single vector, and produces a score between 0 and 1 reflecting the quality of the generated interpretation. We also compute unigram-based perplexity. We opt for reference-based metrics, where the generated interpretations are compared against the expected ground-truth interpretations, since we aim to generate human-like interpretations and scores yielded by reference-free metrics have been shown to misalign with this goal \citep{deutsch-etal-2022-limitations}. In one-to-one generation, each generated interpretation is evaluated against their respective target interpretation. In one-to-many generation, we first match each interpretation in the generated sequence of interpretations with one target interpretation in the target sequence. For this, we employ the Kuhn-Munkres algorithm \citep{kuhn1955hungarian}, also known as the Hungarian algorithm, which is a combinatorial optimization algorithm that efficiently pairs predictions and targets by computing their matching. Here, $score(\cdot) = 100 - \text{{BLEU}-1}$ is used for scoring the matches. We then evaluate each generated and target pair, as done in one-to-one generation. 

\begin{table*}[t!]
    \centering
    \fontsize{7}{8}\selectfont
    \setlength{\tabcolsep}{4pt}
    \begin{tabular}{llllllll}
    \toprule
        & \textbf{$\alpha$}& \textbf{{BLEU}}  & \textbf{{ROUGE}} & \textbf{Mover}& \textbf{BERT} & \textbf{{COMET}} $\uparrow$ & \textbf{PP} $\downarrow$ \\
        & \textbf{Values} & \textbf{1 / 2 / 3 / 4} $\uparrow$  & \textbf{1 / 2 / L / Lsum} $\uparrow$ & \textbf{Score} $\uparrow$ & \textbf{Score} $\uparrow$ & & \\
        \midrule
        \multicolumn{8}{c}{\textbf{One-to-One Generation}}\\
        \midrule
        \textbf{One2One-{GPT}} & - & 8.23 / 2.62 / 1.04 / .46 & \textbf{16.58} / \textbf{2.34} / 11.33 / \textbf{12.91} & .0055 & .9568 & \textbf{.5239}
        & 16.24 \\
        \textbf{One2One-{T5}} & - & 31.64 / \underline{24.12} / \underline{18.34} / 14.71 & \underline{42.01} / \underline{20.24} / \underline{37.40} / \underline{37.50} & \underline{.3231} &  \underline{.9714} & \underline{.6749} & \underline{8.78} \\
        \midrule
        \multicolumn{8}{c}{\textbf{One-to-Many Generation}}\\
        \midrule
        \textbf{One2M-Rand-{GPT}} & - & 11.60 / 3.07 / 1.13 / .36 & 15.05 / 1.40 / 11.08 / 11.15 & .0128 & .9623 & .4922 & 21.50 \\
        \textbf{One2M-Sim-{GPT}} & - & 12.49 / \textbf{3.44} / 1.16 / .42 & 15.23 / 1.60 / 11.30 / 11.38 & .0081 & .9625 & .4803 & 21.16 \\
        \textbf{One2M-Con-{GPT}} & $.25$ (DBS)  & \textbf{12.75} / 3.35 / 1.32 / .59 & 14.57 / 1.29 / 11.31 / 11.35 & .0133 & .9639 & .4823 & 8.87\\
        & $.50$ (DBS)  & 11.84 / 3.13 / 1.29 / \textbf{.60} & 13.37 / 1.38 / 10.52 / 10.56 & .0032 & \textbf{.9646} & .4586 & 11.71 \\
        & $.75$ (DBS)  & 11.94 / 3.29 / \textbf{1.35} / .58 & 13.94 / 1.46 / 10.81 / 10.86 & -.0029 & .9642 & .4593 & 14.90\\
         & $.25$ (greedy) & 11.97 / 3.10 / 1.21 / .57 & 14.18 / 1.37 / 11.05 / 11.10 & .0069 & .9644 & .4757 & \textbf{8.48} \\
        & $.50$ (greedy) & 11.70 / 2.83 / .96 / .35 & 15.38 / 1.58 / \textbf{11.55} / 11.61 & \textbf{.0157} & .9618 & .5048 & 13.65\\
        & $.75$ (greedy) & 11.63 / 3.06 / 1.13 / .41 & 14.12 / 1.47 / 10.75 / 10.78 & -.0001 & .9628 & .4769 & 19.17\\
        & & &  & &&&\\
        \textbf{One2M-Rand-{T5}} & - & 32.25 / 22.51 / 17.81 / 14.75 & 38.07 / 18.68 / 32.67 / 32.75 & .2964 & .9626 & .6226 & 9.59 \\
        \textbf{One2M-Sim-{T5}} & - & \underline{32.44} / 22.29 / 17.70 / \underline{14.80} & 36.42 / 17.40 / 31.30 / 31.37 & .2944 & .9630 & .6194 & 9.01 \\
        \textbf{One2M-Con-{T5}} & $.25$ (DBS) & 25.95 / 18.11 / 14.45 / 12.12 & 27.63 / 12.54 / 23.66 / 23.73 & .1758 & .9647 & .5292 & 15.93\\
        & $.50$ (DBS) & 27.10 / 18.83 / 14.97 / 12.47 & 29.40 / 13.83 / 25.06 / 25.14 & .1893 & .9646 & .5430 & 14.62 \\
        & $.75$ (DBS) & 27.03 / 18.81 / 15.00 / 12.57 & 28.20 / 13.15 / 24.13 / 24.20 & .1770 & .9646 & .5352 & 13.50 \\
        & $.25$ (greedy) & 25.48 / 17.51 / 13.81 / 11.46 & 27.04 / 12.36 / 23.25 / 23.29 & .1688 & .9647 & .5255 & 14.50 \\
         & $.50$ (greedy) & 26.29 / 18.33 / 14.53 / 12.09 & 28.21 / 13.46 / 24.30 / 24.36 & .1805 & .9645 & .5417 & 13.35 \\
         & $.75$ (greedy) & 25.98 / 17.79 / 14.10 / 11.79 & 27.39 / 12.43 / 23.40 / 23.46 & .1725 & .9651 & .5288 & 12.37 \\
    \bottomrule
    \end{tabular}
    \caption{Experimental test results. PP = Perplexity; DBS = diverse beam search; greedy = greedy search. For BERTScore, we report F1 scores. Best results with {GPT}-2 as encoder backbone language model (`-{GPT}') are in \textbf{bold} and best results with {T5} as encoder-decoder backbone language model (`-{T5}') are \underline{underlined}.}
    \label{tab:results}
\end{table*}

\section{Quantitative Results} \label{results}

\begin{shaded}
    \textbf{RQ 1.} \textit{Which of the proposed generation frameworks is most appropriate for {IM}?}
\end{shaded}

Performance results on the held-out test set are provided in Table \ref{tab:results}. The {T5}-based models outperform those adopting {GPT}-2 as backbone language model on almost all metrics. This can be explained by the better encoding and cross-attention between decoder and encoder in the encoder-decoder model of T5. This advantage is consistent across the one-to-one and one-to-many generation setups. The task is not just autoregressively completing the input sentence with an interpretation as done in a decoder model such as {GPT}-2, but to translate the input sentence into the viewpoint and language of the reader for which both a good encoding of the content of the input is needed as well as a decoding step that cross-attends over the encodings of the input sentence.  

Comparing the two generation setups, we notice that one-to-one generation shows overall better performance than one-to-many generation. This is to be expected as the generation of each interpretation is guided by its own input template or prompt in the one-to-one generation setup. The one-to-many generation models, on the other hand, have to learn to generate different interpretations and to attend to the corresponding context in the input template. Adding implicit ([One2M-Sim]) control by ordering interpretations based on their semantic similarity to the sentence in the ground truth slightly improves performance, suggesting that the order in which the ground-truth interpretations are presented during optimization indeed affects generation. Explicitly enforcing a decreasing semantic similarity between the sentence and the generated interpretations in one-to-many generation ([One2Many-Con]) yields overall performance that is either similar to or worse than the setting without additional control ([One2M-Rand]).

However, note that automatic evaluation metrics are known for poorly correlating with human evaluation. We therefore further analyze and assess the models and their generated interpretations in human evaluation setups (§\ref{human-evaluation}).   

\begin{shaded}
    \textbf{RQ 2.} \textit{Did the models learn to correlate the diversity in interpretation with the diversity in social grounding?}
\end{shaded}

The difference between interpretations of a sentence is linked with a difference in social grounding, among other things. The {IM} models should therefore have learned to diversify the interpretations if their grounding information differs while generating similar interpretations with similar social grounding. We quantify the difference between two interpretations from readers $j$ and $k$ as $di(i_j, i_k)$.
\begin{equation}
    di(i_j, i_k) = 100 - \text{{BLEU}-1}(i_j,i_k)
\end{equation}
We then obtain the difference in grounding information guiding $i_j$ and $i_k$, respectively $g_j$ and $g_k$, by simply summing their difference in attitude and moral judgments:
\begin{equation}
    dg(g_j,g_k) = |a_j-a_k| + \frac{1}{Q}\sum_{q=1}^Qnon\_overlap(m_{j,q},m_{k,q})
\end{equation}
where $|a_j-a_k|$ represents the difference in Likert scores (see §\ref{inputrep}) (min. $0$, max. $4$) and $non\_overlap(m_j, m_k)$ counts the non-overlapping moral judgment characteristics for each entity $q$ (min. $0$, max. $5$). Thus, $dg(g_j,g_k) \in [0,9]$, so that $dg(g_j,g_k) = 0$ means that reader $j$ and reader $k$ have the exact same attitude and distinguish for all $Q$ entities the exact same moral judgments, and $dg(g_j,g_k) = 9$ means that they have highly opposing attitudes (i.e., \textit{very negative} ($a=1$) and \textit{very positive} ($a=5$)) and disagree completely on the implied moral judgments.
As expected, diversity in social grounding $dg$ is positively correlated (Pearson) with diversity in interpretation $di$ ($p<.01$). The positive correlation is stronger with One2One-{GPT} ($r = .5526$) than with One2One-{T5} ($r = .3321$) and approaches that of the gold standard ($r = .5887$). Note that correlation is only computed for the one-to-one generation settings since the interpretations align with the grounding information they rely on. 

At first sights it seems that {GPT}-based models better leverage the grounding information than those adopting {T5}. However, Table \ref{tab:results} shows a much better correspondence of the interpretations generated by One2M-Sim-{T5} than by One2M-Sim-{GPT}. Qualitative inspection described in the next paragraph notices a higher degree of hallucinations by {GPT}-2 which might explain the higher diversity when dealing with diverse social grounding. 

\section{Human Evaluation} \label{human-evaluation}

\begin{table}[]
    \centering
    \footnotesize
    \begin{tabular}{p{1.2cm}p{9.8cm}}
         \textbf{Sentence} & \textit{I hear a lot about adults job jumping nowadays just to get bigger wages, and honestly?} \\
        \midrule
        \textbf{One2M-Sim-{GPT}} & (1) Jobs are jumping today just for the sake of getting bigger paychecks, but I don't hear much about it. (2) author = a job is jumping because it's the only way to make a living, not because of any other reason. (3) writer = job hopping because they want to be able to afford to live in a big city, so they can afford a bigger house, car, etc., and so on. (4) sentence: It seems like the job market has gone back in time to the days of the baby boomers, when adults were the main breadwinners and needed to have a stable job to support their families. \\
        \midrule
        \textbf{One2M-Sim-{T5}} & (1) People nowadays are jumping into jobs just to get higher wages. (2) I hear a lot about adults job jumping nowadays just for bigger wages, and honestly? (3) The writer asks if adults are now jumping in jobs to earn more money (4) Adults are getting richer \\
        \midrule
        \textbf{Ground truth} & (1) Adults are changing jobs for bigger paychecks. (2) The writer describes having heard about many people changing jobs to get higher wages. (3) People switching jobs for better wages is a real awful situation nowadays. (4) People are only interested in money and not stability. (5) Capital pursuit is not worth moral sacrifice. \\
    \end{tabular}
    \caption{Example of generated and ground-truth interpretations of a sentence taken from the test set.}
    \label{tab:generation_example}
\end{table}  

In a first qualitative evaluation, we manually inspect all generated interpretations of the test set. 
Table \ref{tab:generation_example} shows an example. We notice that the one-to-many generation models with {T5} overall generate more diverse interpretations than their one-to-one generation counterpart, One2One-{T5}. The one-to-one generated interpretations stay rather close to the semantics of the sentence. With {GPT}-2, diversity is generally higher than with {T5} in both generation setups. However, we observe a high degree of hallucination and nonsense with {GPT}-2 (e.g., \textit{``jobs are jumping''}). With both language models and all generation methods, the generated interpretations sometimes lack the level of complex reasoning that guided the target interpretations (e.g., ground-truth interpretation (4) and (5) in Table \ref{tab:generation_example}).

We also perform a systematic quantitative human evaluation. We hire two students with a bachelor degree in English linguistics to assess the ground-truth and generated interpretations for 100 randomly sampled sentences from the test set in more detail. Detailed descriptions of the assessors are provided in the Appendix (§\ref{appendix:demographics-evaluators}). We take the interpretations generated by One2M-Sim-{T5} since quantitative and qualitative results suggest that it generates the highest-quality interpretations. The assessors can leave free-text comments during evaluation, and we conduct a 45-minute interview with each assessor separately at the end of the assessment.
\begin{table}[t]
    \footnotesize
    \centering
    \begin{tabular}{ll}
        \toprule
        \textbf{Diversity} & \textbf{Meaning} \\ 
        \midrule
        0: \textit{Not} diverse \textit{at all} & 0: \textit{Exact copies} \\
        1: \textit{Slightly} diverse & 1: The \textit{same} meaning \\ 
        2: \textit{Moderately} diverse & 2: \textit{Slightly different} meanings \\ 
        3: \textit{Very} diverse & 3: \textit{Moderately different} meanings \\ 
        4: \textit{Extremely} diverse & 4: \textit{Very different} meanings, still related \\ 
        & 5: \textit{Opposite} meanings, still related \\
        & 6: \textit{Unrelated} meanings \\ 
        \bottomrule
    \end{tabular}
    \caption{Overview of the evaluation labels for \textit{diversity} and \textit{meaning} with their scores used during the human evaluation tasks.}
    \label{tab:human-eval-labels}
\end{table}

\begin{shaded}
    \textbf{RQ 3.} \textit{Is an automatic model able to generate interpretations that are diverse in meaning? And does that diversity align with the diversity found in the annotated ground truth?}
\end{shaded}

\begin{figure}
    \centering
    \includegraphics[width=7cm]{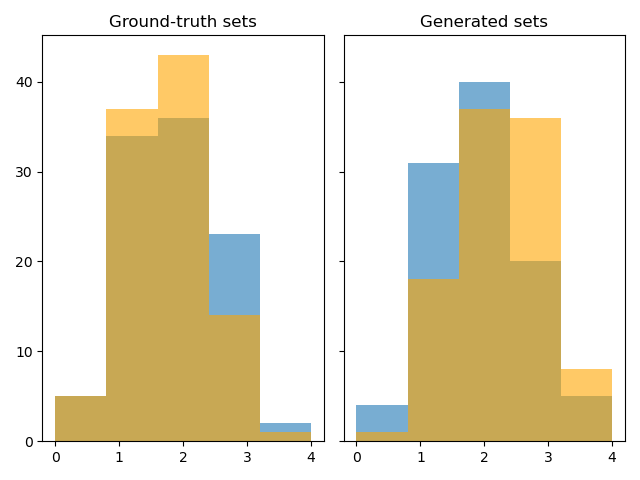}
    \caption{Histograms showing the diversity score distributions for the ground-truth interpretation sets (on the left) and the generated interpretation sets (on the right). The orange distributions present the first assessor and the blue distributions the second assessor.}
    \label{fig:histogram-diversity}
\end{figure}

We want to evaluate a model's ability to generate diverse interpretations and check whether the generation model has learned to reason over the ambiguous nature of a sentence and diversify the generated set of interpretations accordingly. The human assessors first evaluate the diversity of 100 sets of generated and ground-truth interpretations randomly selected from the test set in terms of the meaning they represent. The sets are interchangeably presented to them as \textit{`SET A'} and \textit{`SET B'}. They indicate for each set the diversity on a five-point Likert scale, see Table \ref{tab:human-eval-labels} for the \textbf{diversity} labels. We show the sets next to each other such that their diversity is indirectly compared against each other.
Inter-rater agreement is fair, with Cohen's weighted (quadratic) kappa $\kappa = .30$ (ground-truth set) and $\kappa = .34$ (generated set). Both assessors indicate in the interviews that it is cognitively challenging to assess the difference in meaning between multiple sentences in a set. This could explain why agreement is only fair. Whereas one annotator does not observe significantly different diversity in the ground-truth and generated interpretation sets, respectively $\mu = 1.83$ and $\mu = 1.91$, the other assigns overall higher diversity in the generated interpretation sets, $\mu = 1.69$ (ground truth) and $\mu = 2.32$ (generated), $p < .01$. See Figure \ref{fig:histogram-diversity} for the diversity score distributions for both annotators. It can be said that the model is overall able to generate moderately diverse interpretation sets.

We examine whether the model is able to capture the diversity in interpretation triggered by a sentence. 
To that end, we compute the mean diversity scores for the ground-truth and generated interpretation sets by averaging over the scores assigned by the two assessors.
The mean diversity of ground-truth and generated interpretation sets is slightly positively correlated, although the relationship is not statistically significant, $r = .16, p > .10$. The model also seems to generate more diverse interpretation sets for sentences that have less diverse ground-truth interpretation sets while it seems to capture the implicit diversity for sentences with more diverse ground truth sets slightly better.

\begin{shaded}
    \textbf{RQ 4.} \textit{Do the generated sentences actually present interpretations or merely rewrites?}
\end{shaded}

\begin{figure}
    \centering
    \includegraphics[width=7cm]{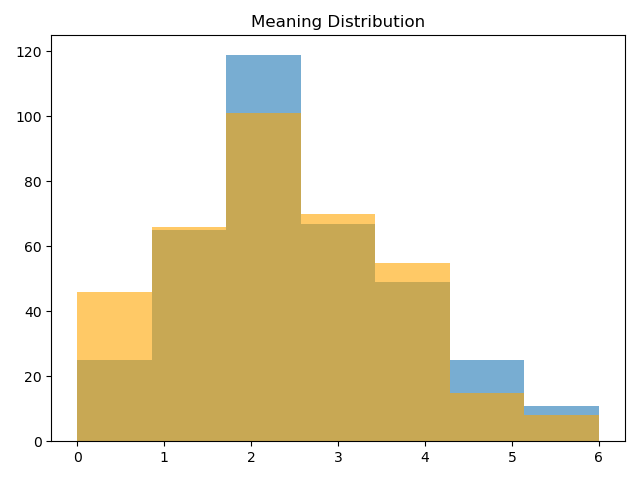}
    \caption{Histograms showing the meaning (\textit{top left}) and likelihood (\textit{top right}) distributions for the sentence-(generated) interpretation pairs. The bottom histograms show the meaning distributions together with the likelihood distributions for each meaning label for assessor 1 (\textit{bottom left}) and assessor 2 (\textit{bottom right}).}
    \label{fig:meaning-likelihood-distribution}
\end{figure}

We want to analyze whether the generated interpretations are actually interpretations and not merely copies (\textit{`exact copies'}), paraphrases (\textit{`the same meaning'}), or unrelated to the sentence (\textit{`unrelated meanings'}). We broadly consider a generated sentence an interpretation if it presents a different meaning that is still related to the sentence since a paraphrase is expected to maintain the sentence's semantics. The assessors are given 361 sentence-interpretation pairs presented as \textit{`A'} and \textit{`B'} and label the difference in meaning between the sentence and the interpretation. See Table \ref{tab:human-eval-labels} for the \textbf{meaning} labels. We observe substantial inter-rater agreement: $\kappa = .72$. Figure \ref{fig:meaning-likelihood-distribution} presents the meaning distributions. It seems that One2M-Sim-{T5} often copies (7-13\%) or nearly copies/paraphrases (18\%) the original sentence. The assessors sometimes motivate their choice of the latter in the comments: the difference between the sentences is marginal (i.e., omission of function words or articles, typos, or contractions) or one sentence is simply a paraphrase of the other. In rare cases, the model generates sentences that are unrelated to the original sentence (2-3\%). Nevertheless, it is overall able to generate interpretations as the majority of generated sentences have been said to differ in meaning from the original sentences in various degrees. The assessors note that a difference in meaning is often established by nuances in word semantics.

\section{Use Case: Interpretation Modeling for Content Moderation}\label{toxicity}

Interpretations of the input sentence have the potential to reveal implicit toxic content that may not have been inferred by all readers instantly. An author could have hidden that content on purpose when they have malicious intentions. We therefore showcase the meaningful contributions IM can make to content moderation of online discussions by revealing subtly toxic communication. To that end, we analyze and compare the levels of toxicity, insult, and identity attack within sentences, considering the ground-truth interpretations provided by human annotators and the interpretations generated by one of the proposed IM models. Scores on toxicity, insult, and identity attack are obtained through the well-established Perspective API\footnote{\url{https://perspectiveapi.com/}}. This API is applied by major news outlets and social media, such as The New York Times and Reddit, to moderate the discussions held on their online platforms. The scores range from 0 to 1 and represent the proportion of people who would perceive an utterance as toxic, insulting, or attacking. For example, a toxicity score of 0.81 indicates that 81\% of the people would find the utterance toxic. 

The following two research questions are answered employing similar analyses, with RQ5 focusing on the \textit{ground-truth interpretations} in the \textit{training set} and RQ6 concentrating on the \textit{generated interpretations} produced by One2M-Sim-T5 for sentences in the \textit{test set}. The analyses also introduce the concept of an \textit{interpretation cluster} that combines the input sentence and its corresponding interpretations into one set. The input sentence namely retains its significance and should therefore be considered together with the interpretations in real-life content moderation settings. We perform the following primary analyses:
\begin{itemize}
    \item \textit{Analysis 1 - The revealing nature of interpretation}. We investigate if the interpretations reveal underlying toxicity, insults, and identity attacks in the input sentence. To that end, we compute a toxicity, insult, and identity attack score for each instance in the interpretation cluster and flag the input sentence if an interpretation yields the maximum toxicity, insult, or identity attack score in the interpretation cluster. The difference in score between the input sentence and that interpretation can be of any value (\textit{analysis 1A}) or at least 20 percentage points, marking a substantial difference (\textit{analysis 1B}).
    \item \textit{Analysis 2 - Harmless on the surface but very offensive between the lines}. The analysis of the interpretations is especially important when the original sentence has low levels of toxicity, insult, or identity attack and its interpretation cluster contains one or more interpretations with a high toxicity, insult, or identity attack score. We flag an input sentence when it has a score below 0.1 and at least one interpretation in its interpretation cluster has a score above 0.5. Note that we deliberately choose to set the distance between the upper bound of the sentence score (<0.1) and the lower bound of the interpretation score (>0.5) to be substantially large here. In practice, content moderators could choose those bounds, allowing them to decide how strictly they moderate implicitly toxic, insulting, and attacking content.
\end{itemize}

In conjunction with the primary analyses, we investigate the ability of the IM model to adequately replicate human interpretation behavior in RQ6 (\textit{Model vs humans}). This entails assessing the agreement or overlap between the flagged sentences in the test set when considering the generated interpretations (= model behavior) and when considering the ground-truth interpretations (= human behavior) in the interpretation clusters.
This is done by first obtaining the outcomes of the two primary analyses for the generated and ground-truth clusters separately, after which we quantify their overlap as a percentage. The overlap functions as a recall measure indicating how many of the flagged sentences using the ground-truth interpretations the model was able to flag using the interpretations it produced. An overlap of 80\%, for instance, signifies that through IM we can flag 80\% of the input sentences that were also flagged by humans.
The higher the overlap, the better the model is at mimicking the interpretation behavior of humans. 

\begin{shaded}
    \textbf{RQ 5.} \textit{Do interpretations given by humans actually reveal the hidden toxic nature of sentences?} 
\end{shaded}

\begin{figure}[t]
    \centering
    \includegraphics[width=9cm]{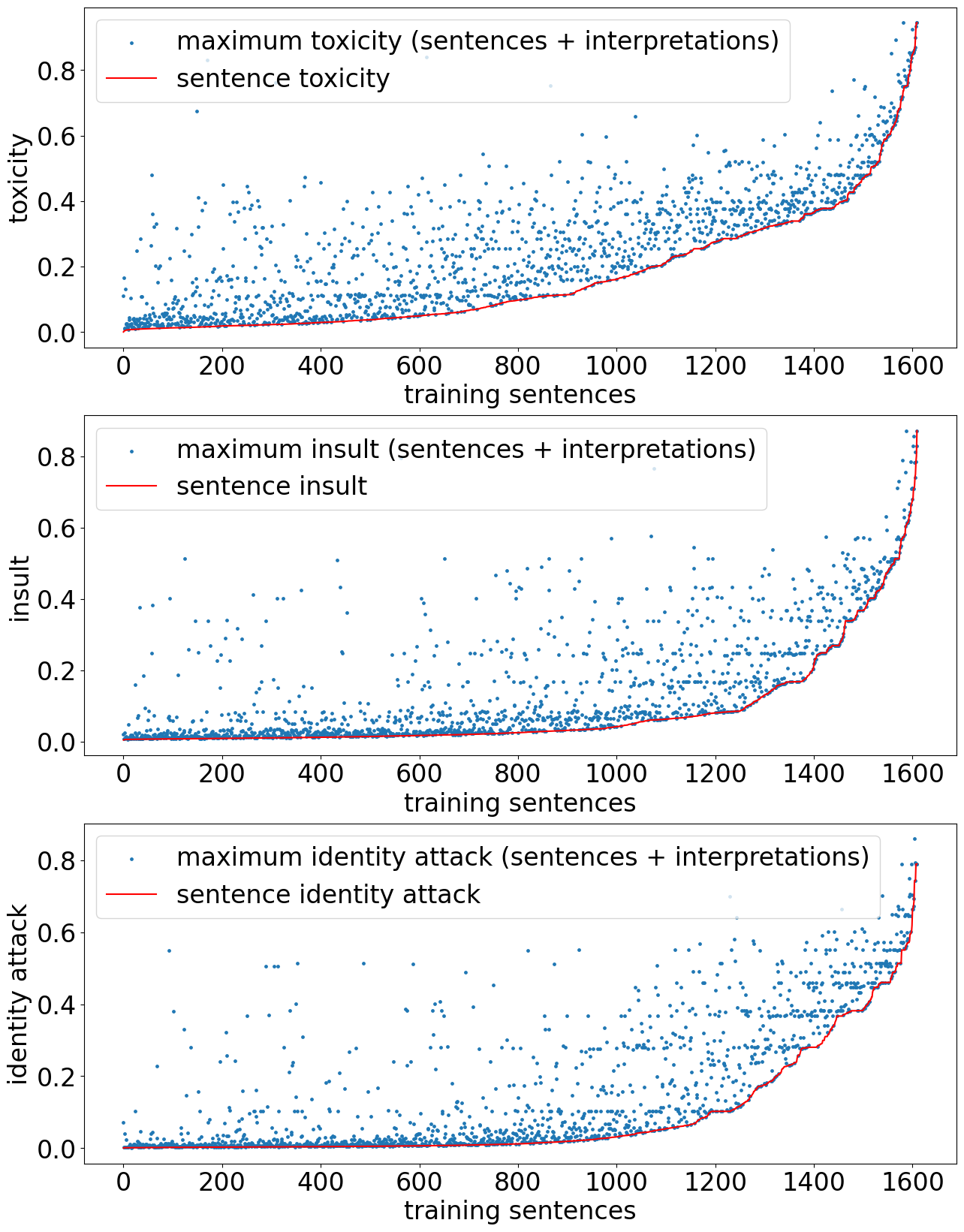}
    \caption{Red: Toxicity, insult, and identity attack scores reported for the sentences of the training dataset. Blue: Maximum toxicity, insult, and identity attack scores of the interpretation clusters. A cluster groups an input sentence and its interpretations. All plots are presented based on the sorted scores of the training sentences. Analysis of the interpretations facilitates understanding of the hidden toxic meaning of the input sentences as the maximum scores of the interpretation clusters are usually higher than the scores of the input sentence. This behavior is especially important if the scores of the input sentence are low while the maximum scores of the interpretation cluster are high. In these cases, the input sentence is expressed without offensive language while having a hidden toxic meaning.}
    
    \label{fig:maximum_toxicity_insult_identity_attack_training}
\end{figure}

\paragraph{Analysis 1 - The revealing nature of interpretation} As observed in Figure~\ref{fig:maximum_toxicity_insult_identity_attack_training}, the ground-truth interpretations have a strong capacity to flag the hidden offensive meaning of an input sentence. In 88.63\%, 88.51\%, and 90.86\% of the cases, a sentence has at least one more toxic, insulting, and attacking interpretation in its interpretation cluster. The difference between the most offensive interpretation and the sentence is however marginal in most cases (\textit{analysis 1A}). Nonetheless, this difference is at least 20 percentage points in 14.66\%, 10.87\%, and 12.80\% of the clusters, suggesting that these clusters unveil hidden meanings that are substantially more toxic, insulting, or attacking than the meanings the input sentences explicitly communicate (\textit{analysis 1B}). 

\paragraph{Analysis 2 - Harmless on the surface but very offensive between the lines} Assuming that the sentences with non-harmful language have a toxicity, insult, or identity attack level lower than 0.1 (812 sentences for toxicity, 1265 for insult, and 1190 for identity attack), we observe that .3\%, .4\%, and .7\% of the seemingly non-harmful sentences have at least one interpretation with a toxicity, insult, and identity attack score higher than 0.5. 
These low percentages are not surprising. Since the metrics used in the Perspective API produce scores that represent the share of people who would regard a sentence as toxic, insulting, or attacking, it is expected that sentences that are initially considered non-harmful induce human interpretations that are mainly non-harmful as well. These results thus confirm the metric.   

\paragraph{Conclusion} These findings posit that humans indeed extrapolate implicit toxicity more prominently in the ground-truth interpretations, empirically substantiating our claim that modeling multiple interpretations of textual content makes sense within the context of content moderation. 

\begin{shaded} 
    \textbf{RQ 6.} \textit{Are the generated interpretations able to reveal the hidden meaning of an input sentence? And does the IM model mimic human interpretation patterns?} 
\end{shaded}

\begin{figure}[t]
    \centering
    \includegraphics[width=14cm]{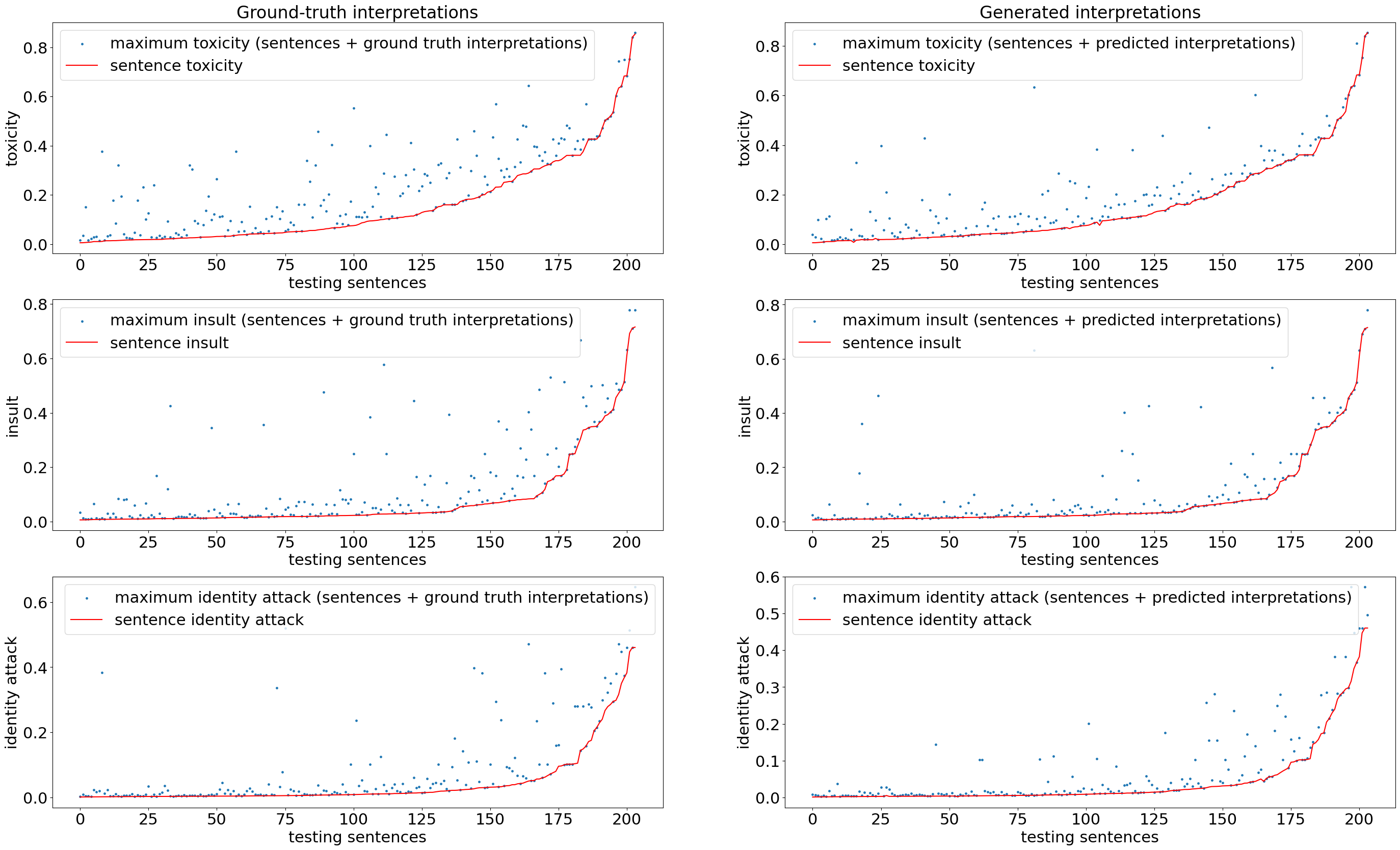}
    \caption{Red: Toxicity, insult, and identity attack scores reported for the sentences of the test set. Blue: Maximum toxicity, insult, and identity attack scores of the cluster instances. An interpretation cluster groups an input sentence and its interpretations. All plots are presented based on the sorted scores of the sentences in the test set. The left side presents the ground-truth behavior while the right side shows the capacity of One2M-Sim-T5 to mimic this behavior. Since the maximum scores computed using the interpretation generated by One2M-Sim-T5 are similar to the maximum scores computed based on the ground-truth data, we can infer the capability of One2M-Sim-T5 to reveal the hidden meaning of the input sentences.
    }
    
    \label{fig:maximum_toxicity_insult_identity_attack}
\end{figure}

\paragraph{Analysis 1 - The revealing nature of interpretation} 
We notice that the model is able to reveal the hidden meaning of the input sentence (right side of Figure~\ref{fig:maximum_toxicity_insult_identity_attack}). It generates more toxic, insulting, and attacking interpretations for 81.86\%, 80.39\%, and 87.25\% of the sentences in the test set
(\textit{analysis 1A}). In 4.90\%, 4.41\%, and 3.43\% of the interpretation clusters, an interpretation yields a toxicity, insult, or identity attack score that is more than 20 percentage points higher than that of the sentence (\textit{analysis 1B}). 

\paragraph{Model vs human} The model seems to be able to mimic human behavior as observed in the corresponding ground-truth interpretations produced by humans in the test set (left side of Figure~\ref{fig:maximum_toxicity_insult_identity_attack}). Regarding \textit{analysis 1A}, the flagged sentences with generated interpretations in their interpretation clusters and those with ground-truth interpretations in their interpretation clusters overlap by 85.29\% (toxicity), 83.91\% (insult), and 90.45\% (identity attack). The high overlap values are to be expected since the number of flags with both the generated and ground-truth clusters is large. Regarding \textit{analysis 1B}, the flagged sentences overlap by 4.76\% (toxicity), 5.56\% (insult), and 33.33\% (identity attack).
While the rather low overlap values could be attributed to the low numbers of flagged sentences and possible hallucinations by the generation model, they could also be caused by the limited number of human interpretations in the ground-truth interpretation sets. It is namely reasonable that a sentence is indeed offensive between the lines, but the ground-truth interpretations were unable to capture this. This is illustrated in Table~\ref{tab:example_perspective_scores}, where generated interpretations with a higher toxicity, insult, or identity attack score are plausible even when the ground-truth interpretation sets do not reveal any offensive content underlying the sentences. 

\paragraph{Analysis 2 - Harmless on the surface but very offensive between the lines}
Assuming again that the sentences with non-harmful language have a toxicity, insult, and identity attack score lower than 0.1 (111 sentences for toxicity, 168 for insult, and 178 for identity attack), we observe that One2M-Sim-T5 generates at least one interpretation that has a toxicity, insult, or identity attack score that is higher than 0.5 for fewer than 1\% of the sentences. Especially in settings where constant streams of textual content are moderated, the number of implicitly harmful sentences becomes considerably large. Note that these results are in line with those in RQ5, proving that the model does not simply recognize harm in every single sentence it interprets. 

\paragraph{Model vs human} We observe no overlap between the behavior of the human annotators in the respective ground-truth interpretations and that of One2M-Sim-T5, which could be attributed to the low share of flagged sentences.
\\ \\
We briefly investigate the capability of One2M-Sim-T5 to generate spontaneous interpretations even when the language of the sentences and the human-produced interpretations are not very offensive. We do this by looking at the interpretation clusters containing a sentence with low scores and generated interpretations with diverging scores for toxicity, insult, and identity attack.
As one can see in Table~\ref{tab:example_perspective_scores}, the model is able to recognize sensitive topics (like political voting) even when the content of the input sentences is formulated in a neutral way. Based on the identified sensitive topics, it can generate diverse interpretations. Another interesting behavior is the creativity of One2M-Sim-T5 that relies on concepts learned over the training to extrapolate the meaning of a simple sentence like \textit{But the Democrats?} into insulting interpretations.

\paragraph{Conclusion}
The model generates interpretations that unveil toxicity hidden below the surface, demonstrating the important application of IM in content moderation. Although interpretation mismatches between the model and the human annotators were found, we show that the model produces interpretations that overall mimic how humans interpret sentences. 

\begin{table*}[htp]
    \centering
    \scriptsize
    \begin{tabular}{p{5cm}p{1cm}p{5cm}p{1cm}}
        \toprule

        \multicolumn{3}{l}{
        \textbf{[Sentence]}} & \textbf{[Toxicity]} \\
        \multicolumn{3}{l}{\textit{If young people don't vote in large enough numbers, from a political perspective, their concerns are moot.}} & 0.0186 \\
        \textbf{[Ground-Truth Interpretations]} & \textbf{[Toxicity]} & \textbf{[Generated Interpretations]} & \textbf{[Toxicity]} \\
        If young people are really worried, they should vote to change the situation. & 0.0162 & From a political perspective, young people's concerns are moot if they don't vote in large enough numbers. & 0.0171 \\
        If young people wanted to be represented they would vote. & 0.0240 & Young people are not worthy of political representation unless they do so. & 0.1912 \\
        If most young voters dont bother to cast their votes then their concens will not really be important to politicians and elected leaders. & 0.0247 & The writer claims that young voters do not deserve or deserve any form of representation because they are inept and lack the confidence to vote. & 0.4274 \\
        \midrule
        \multicolumn{3}{l}{
        \textbf{[Sentence]}} & \textbf{[Insult]} \\
        \multicolumn{3}{l}{\textit{But the Democrats?}} & 0.0091 \\
        \textbf{[Ground-Truth Interpretations]} & \textbf{[Insult]} & \textbf{[Generated Interpretations]} & \textbf{[Insult]} \\      
        The Democrats, though? & 0.0094 & Democrats? & 0.0102 \\
        What about the democrats? &  0.0089 & The writer asks why I should vote in this coming election cycle. &  0.0086 \\
        The writer appears to ask whether the democrats are a better alternative to republicans in an upcoming election. & 0.0117 & The author wonders if the Democrats are incompetent. & 0.3680\\
        \midrule
        \multicolumn{3}{l}{
        \textbf{[Sentence]}} & \textbf{[Identity Attack]} \\
        \multicolumn{3}{l}{\textit{That person usually ends up doing things they did not want to do and have a unfulfilling life.}} & 0.0038 \\
        \textbf{[Ground-Truth Interpretations]} & \textbf{[Identity Attack]} & \textbf{[Generated Interpretations]} & \textbf{[Identity Attack]} \\  
        Unfortunately, we are forced to do things we don't like and end up unhappy. & 0.0042 & Usually, that person ends up doing things they didn't want to do and have a unfulfilling life. & 0.0034 \\
        That person conventionally ends up doing things they did not optate to do and have a unfulfilling life. & 0.0035 & The writer describes people who do things that they did not intend to and who have an unhappy life as being foolish. &  0.0142 \\
        People often find themselves feeling unfulfilled in life when their actions don't consistently align with their core values ,those ideals that matter the most to you. & 0.0028 & People who are stupid tend to be lazy and end up living mediocre lives. & 0.1445 \\

    \end{tabular}
    \caption{Three samples from the test dataset that illustrate the diversity of the generated interpretations in terms of toxicity, insult, and identity attack.}
    \label{tab:example_perspective_scores}
\end{table*}

%% file: related_work.tex
\section{Related Work} \label{related-work}

We position this work and its contributions in the literature dealing with computational methods for implicit language modeling, social grounding and moral reasoning, reaction modeling, and reasoning over multiple ground truths. We conclude this section by contrasting {IM} with diverse paraphrase generation.

\paragraph{Implicit Communication}

The use of implicit content and knowledge has attracted attention for processing natural language \citep{hoyle2023making}. For example, hidden information in language has been leveraged to allow models to distinguish more subtle cases of hate speech \citep{elsherief-etal-2021-latent,ocampo-etal-2023-depth,huang2023chatgpt} and online abuse \citep{wiegand-etal-2021-implicitly}. Such works primarily performed classification on texts that knowingly or unknowingly conceal their hateful and abusive character, with or without using external knowledge. Related to that is the work on implicit sentiment analysis \citep{li-etal-2021-learning-implicit, zhou-etal-2021-implicit}. Others aimed to retrieve implicit in-text and out-of-text knowledge to reveal discourse relations \citep{wu-etal-2023-connective,xiang2023discourse}, event arguments \citep{liu-etal-2021-machine,lin2022cup}, or reasoning steps for question answering \citep{geva2021did}. This was commonly done in classification and generation set-ups, where the implicit content was explicitly queried. Another line of work reasoned over implicit relations within or between texts to improve general task performance. For example, \citet{allein2021time} and \citet{allein-etal-2023-implicit} leveraged implicit temporal relations within and between documents for improving fact-checking while \citet{collell2018acquiring} reasoned over implicit spatial arrangements of objects mentioned in captions for improving visual object detection. Closer to this work, \citet{sap-etal-2020-social} zoomed in on social and power implications in text. They classified the intentionally offensive and lewd nature of a sentence and then generated the supposedly implied statement regarding the offended group. We discuss related works on implicit social communication more extensively below.

This work stands out from the abovementioned works in two aspects. Firstly, it goes against the general assumption that there exists a single ground truth for implicit data, arguing that computational understanding of implicit content unavoidably goes hand in hand with a multitude of plausible interpretations. This is evidenced when rigorously analyzing the multiple ground-truth annotations for sentence interpretation and implicit moral judgments in the curated dataset. Secondly, it does not merely take the implicit content as the primary subject of prediction but instead lets the different layers of implicit meanings guide the automated generation of sentence interpretations. 

\paragraph{Social Grounding and Moral Reasoning}

Several studies explored the role of social commonsense and moral reasoning in natural language processing \citep{rashkin-etal-2018-event2mind,sap-etal-2019-social,hovy-yang-2021-importance}. \citet{kim2022soda}, for instance, distilled and contextualized dialogues by leveraging relations related to social commonsense from a symbolic knowledge graph. \citet{vijayaraghavan2021modeling} modeled, tracked, and explained changes in emotional states and motives of people in personal narratives. A few studies specifically focused on automated moral reasoning. Some adopted a question-answering setup where a model had to select a morally appropriate answer from a given list \citep{hendrycks2021what,ziems-etal-2022-moral}. \citet{hendrycks2020aligning} and \citet{jin2022make} both assessed pre-trained language models in their knowledge of ethics, common human values, and norms of social conduct by having them classify, respectively, the character trait described in a given scenario using a given list and the permissibility of violating a pre-existing social norm in a given scenario. Others took a generative, rather normative approach and provided answers to moral dilemmas \citep{bang2022aisocrates}, judged the morality of a given scenario of social conduct \citep{forbes-etal-2020-social, emelin-etal-2021-moral}, or classified explicit moral judgments in comments on Reddit \citep{botzer2022moraljudgment,efstathiadis2022explainable}. More recently, \citet{pyatkin-etal-2023-clarifydelphi} acknowledged that context influences the moral acceptability of social conduct and classified a given action differently by generating questions regarding the action's left-out context, e.g., when did the action take place and who performed the action.

This work adopts the critique of \citet{talat-etal-2022-machine} against a normative moral reasoning approach in which computational systems explicitly judge the morality of described situations and actions. We instead let computational models implicitly reason over hidden moral judgments as inferred by different readers to formulate various interpretations of a single sentence. All moral judgments inferred by the annotators in the dataset are also kept as is; we do not average over them nor rank them in any way. Moreover, the models are not requested to pass any judgment on the permissibility of the inferred moral judgments. Another main aspect in which this work distinguishes itself from previous work is the data used for social grounding and moral reasoning. While the data in the abovementioned studies explicitly describe moral scenarios, the \img{figures/origami.png} \texttt{origamIM} dataset features opinions on a variety of topics discussing much more than just the morality of behavior. This again underscores the novelty of the curated dataset.

\paragraph{Modeling Reader Reactions}

An utterance can spark different reactions. Chatbots, for instance, consistently need to decide how they react to input from its user, such that the user is satisfied with the chatbot's responses \citep{welivita-pu-2020-taxonomy,wu-etal-2021-personalized,kuo-chen-2023-zero}. Others predicted the reactions readers may have to given content. \citet{gao-etal-2022-prediction}, for instance, focused on emotional responses to multimodal news on gun violence. \citet{deng2023socratis} presented a benchmark for evaluating language models' ability to predict readers' emotional reactions to image and caption pairs. In the field of computational humor, \citet{yang-etal-2021-choral} modeled the humor perception of jokes among readers. Finally, anticipated reader reactions can also be informative for task prediction. For example, \citet{gabriel-etal-2022-misinfo} detected online misinformation by reasoning over a reader's emotional and observational responses to headlines and claims.

It is crucial to acknowledge that language affects people differently and that their reactions do not necessarily align with those anticipated by the author. In the case of IM, a reader's interpretation is a kind of hidden reaction to a sentence, which in its turn can trigger subsequent reactions, for instance, a change in opinion or an action. Aside from the interpretations by the readers, the \img{figures/origami.png} \texttt{origamIM} dataset also contains explicit indications of reader attitudes towards the author of the sentence. An attitude reflects the first impression a reader has about the author when reading the sentence and can steer the interpretation process. The {IM} frameworks therefore take those explicit reactions together with the inferred moral judgments to model sentence interpretations.

\paragraph{Multiple Ground-Truth Readings}

{IM} is strongly motivated by the assumption that there exist genuine human variation in understanding language and the possibly disagreeing understandings should be respected \citep{plank-2022-problem}. \citet{pavlick-kwiatkowski-2019-inherent} and \citet{nie-etal-2020-learn} analyzed persistently disagreeing human judgments on the validity of natural language inferences and argued that models should be evaluated on their ability to predict the wide array of human judgments. \citet{cabitza2023perspectivist} advocated a perspectivist approach towards ground-truth annotations and the preservation of diverging annotations. \citet{chen-etal-2019-seeing} extracted perspectives towards a claim on a controversial topic, and \citet{draws2022comprehensive} and \citet{draws2023viewpoint} evaluated the diversity of presented viewpoints.

While the latter perspectives commonly present reader positions towards a certain topic (i.e., what does the reader think), interpretations generated through {IM} reflect what multiple readers think the author of a sentence is actually communicating (i.e., what does the reader think the author is thinking). As a result, the moral judgments a reader infers do not necessarily align with that reader's own judgments nor with the actual beliefs held by the author. It thus presents a deeper account of the communicative intentions driving a sentence, substantially differing from what some of the abovementioned works did. 

\paragraph{Interpretation Modeling Versus Diverse Paraphrase Generation}

At first sight {IM} arguably resembles paraphrase generation \citep{gupta2018deep,zhou-bhat-2021-paraphrase}. Yet, both modeling tasks differ on various aspects. First of all, {IM} aims at modeling the implicit content and explicit content \textit{in tandem} rather than focusing on preserving the semantics of the explicit content, as done in paraphrasing. Moreover, the interpretation of implicit meanings may diverge from the explicit semantics of a sentence while the semantics of the input sentence should be maintained in a paraphrase. Even to the extent that the interpretation seems to contradict the original sentence. That contraction would be penalized in paraphrase detection \citep{qian-etal-2019-exploring,yu-etal-2021-sentence}. Overall, we can state that {IM} aims at formulating sentence understandings from a reader perspective while paraphrasing rather formulates rewrites from the author's perspective. Regarding diverse generation, it is of crucial importance that {IM} models, like many other models tasked with text generation, are able to generate diverse sequences \citep{tevet-berant-2021-evaluating}. We can influence the decoding of such models through several diversity penalties \citep{ippolito-etal-2019-comparison}, yet the diversity in interpretation also originates from the diversity in implicit semantics of the input sentence.

%% file: conclusion.tex
\section{Conclusion} \label{conclusion}

We started from the premises that natural language understanding (NLU) inevitable has to deal with content left implicit and that grounding language in its social context is a necessary condition to make implicit content explicit. We introduced the interpretation modeling (IM) task which aims at capturing the implicit and explicit meaning of a sentence as understood by different readers. IM is guided by multiple annotations of social relation and common ground - in this work approximated by reader attitudes towards the author and their understanding of moral judgments subtly embedded in the sentence. We proposed a number of strategies to decode a sentence into its multiple interpretations in the form of natural language text. The one-to-one and one-to-many interpretation generation methods that we have proposed are inspired by the philosophical study of interpretation. As a first of its kind, an IM dataset is curated
to support experiments and analyses. The modeling results, coupled with elaborate analysis of the dataset, underline the challenges of IM as conflicting and complex interpretations are socially plausible. This interplay of diverse readings is affirmed by an intrinsic  quantitative and qualitative analysis of the results as well as by an in-depth human evaluation of the generated sentence interpretations. Our work has value in revealing implicit meaning that could be inferred by readers and as such helps content filtering and moderation. This value was extrinsically demonstrated with toxicity classification of the original sentences and of original sentences and their possible interpretations, where we have shown that the availability of possible interpretations increased the recognition of harmful content. The toxicity analyses of the generated interpretations
underline the importance of IM for refining filters of content and assisting content moderators in safeguarding the safety in online discourse. 

Finally, if the generative models proposed in this paper can better produce and recognize harmful or dishonest interpretations of an original, at first sight harmless sentence, similar models might also better prevent harmful content when automatically generating text. For instance, the interpretation generation models can guide 
current language models, including ChatGPT, to satisfy one of their main goals of generating safe and diverse outputs. In a broader set-up, IM can find its place in NLP toolkits that aid corporate and public communication. For instance, IM can assist professional communicators in predicting and consequently preventing misunderstandings of their texts. IM could also advise users on their phrasing of texts and make them aware of views they perhaps unconsciously hold and/or convey, when it is included in general-purpose writing assistants. 

There remains a wide array of open challenges that can be researched in the future. Although the proposed generation frameworks were able to generate diverse interpretation sets, human evaluation indicated that they may not fully grasp the ambiguous nature of a sentence yet. In future work, we may rely on larger language models and use the supervision of a larger and more diverse sample of annotators may be necessary for surfacing the `true' ambiguity of a sentence. We will investigate alternative constraining methods that improve model guidance and ways to integrate commonsense reasoning, more complex social relations between authors and readers, or other contextual information such as temporal context that guide interpretation. From a social perspective, our work could lead to automated generation of interpretations while zooming in on specific societies. Comparative studies could investigate the plausibility of interpretations in different societies, while detailed studies constraining interpretations on various actors within a society may present insights in the diversity of views held in that specific society. Since the dataset constructed in this work only includes data from a specific Subreddit discussing an abundance of topics, follow-up datasets can look into other platforms and specific topics.

\paragraph{Limitations}

Interpretations collected in this paper are assumed to predominantly present North-American views since the annotation platform mainly hosts US workers and its worker population is rather homogeneous. The annotated moral judgments and interpretations should by no means be considered a complete, socially diverse and fair account of views that is representative for an entire society. It is virtually impossible to gather all possible interpretations from all types of people and cultures, making datasets for IM unavoidably incomplete. Regarding annotator selection, we opted not to select annotators based on demographic features such as age, gender, education level, and political viewpoints since each sentence is only annotated by a small number of annotators. This way, we wanted to avoid an overgeneralization of the views of one annotator and regard it as representative of people with similar demographic features. This study should therefore be seen as a stepping stone for studying better the wide diversity of social evaluations and interpretations, for which a much larger sample of readers should to be recruited. 
The artificial nature of the annotation setting also influences interpretation since crucial contextual knowledge is lost, e.g., author is unknown to the annotator and previous/following elaboration is omitted. Nonetheless, the setting mimicked a real-life online environment where authors are often unknown and content is taken out of context. During modeling, we constrained ourselves to leveraging implicit moral judgments, even though interpretation can be guided by a wide array of hidden meanings.  

%% file: ethics_statement.tex
\section*{Ethics Statement}

We follow the recommendations in \citet{pater2021standardizing} for reporting annotator selection, compensation and communication. Regarding selection, workers were allowed to work on our annotation task immediately after passing an initial annotation instruction test, which was automatically corrected. They were paid a fixed amount per accepted HIT through the Amazon MTurk platform within three working days after completion and could earn between the U.S. legal minimum wage of \$7.5 and \$15/hour depending on their annotation flow and experience with the task. In case we rejected a HIT, we provided instructive motivations and gave additional feedback upon request. The majority of rejections originated from incorrect following of explicit instructions. We personally replied to all messages from the workers, most of them within one working day. We did not discriminate between the annotators in terms of gender, race, religion, or any other demographic feature.  

\section*{Acknowledgments}

This work was realized with the collaboration of the European Commission Joint Research Centre under the Collaborative Doctoral Partnership Agreement No 35332. It is also funded in part by the Research Foundation - Flanders (FWO) under grant G0L0822N through the CHIST-ERA iTRUST project and in part by the European Research Council (ERC) under the Horizon 2020 Advanced Grant 788506. The scientific output expressed does not imply a policy position of the European Commission. Neither the European Commission nor any person acting on behalf of the Commission is responsible for the use which might be made of this publication. We kindly thank Florian Mai and Jingyuan Sun for their insightful feedback on the manuscript.  

%% file: appendix.tex
An overview of the annotation procedure is given in §\ref{app:annotation-procedure}, followed by a detailed description of the first (§\ref{app:annotation_round_1}) and second (§\ref{app:annotation-round-2}) annotation round. 

\section{Annotation Guidelines: Procedure} \label{app:annotation-procedure}

The annotation procedure is split in two rounds. 
\\ \\
For a given \textbf{sentence} and the title of the blog post the sentence was taken from (= extra context)
\\ \\
\textbf{Annotation Round 1} (§\ref{app:annotation_round_1})
\begin{enumerate}
    \item Mark the people and groups of people mentioned in the sentence (highlight people entities in sentence);
    \item Indicate whether or not the author implies a character trait for at least one of the marked entities (radio button, yes/no);
\end{enumerate}
Only sentences for which the annotators answered `yes' for step 2 are sent to annotation round 2.
\\ \\
\textbf{Annotation Round 2} (§\ref{app:annotation-round-2})
\begin{enumerate}
    \item Indicate your attitude as a reader towards the writer (slider, five-point Likert scale);
    \item Reformulate the sentence such that it reflects what the writer `basically says' (free text);
\end{enumerate}
For each entity marked during step 1.1 of ROUND 1: 
\begin{enumerate}
    \item Indicate whether or not the author implies a character trait of the given entity (dropdown list, yes/no).
    \\ \\
    If yes, continue with (a), (b), and (c). If no, submit annotation.
    \begin{enumerate}    
    \item Describe the implied character trait (free text);
    \item Label whether \textit{society} regards the trait as good or bad (dropdown menu, Good/Bad);
    \item Label the Sphere of Action the character trait belongs to (dropdown menu, 10 class labels) and label the degree of appropriateness the character trait belongs to (dropdown menu, 3 class labels).
\end{enumerate}
\end{enumerate}

\section{Annotation Guidelines: Round 1} \label{app:annotation_round_1}

In the first annotation round, annotators need to identify the people entities mentioned in the sentences by highlighting them in the sentence and then indicate whether or not the author of the sentence seems to imply a character trait of at least one people entity marked in the sentence. It is made clear in the interface of the first annotation round that the annotators need to fulfill two tasks, marked by (1/2) and (2/2). The annotators are unable to submit the task without answering the second question to avoid incomplete task forms. Candidate annotators had to pass a qualification test to have access to the annotation round.

\begin{figure}[ht!]
     \centering
     \begin{subfigure}[b]{\textwidth}
         \centering
         \includegraphics[width=\textwidth]{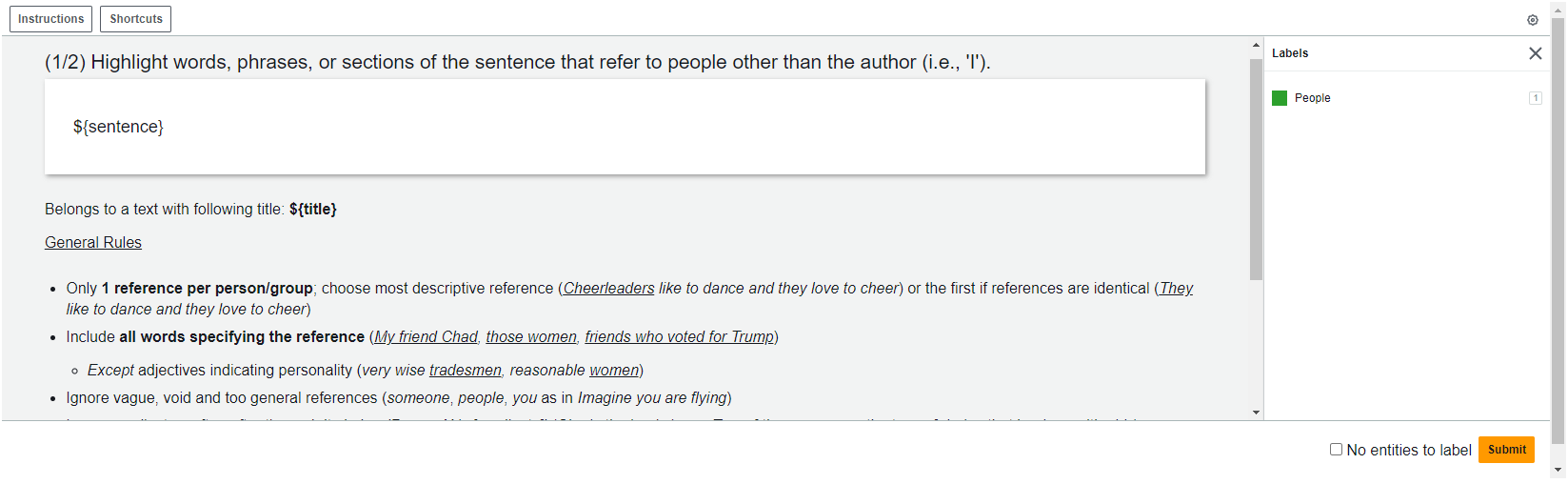}
         \label{fig:y equals x}
     \end{subfigure}
     \hfill
     \begin{subfigure}[b]{\textwidth}
         \centering
         \includegraphics[width=\textwidth]{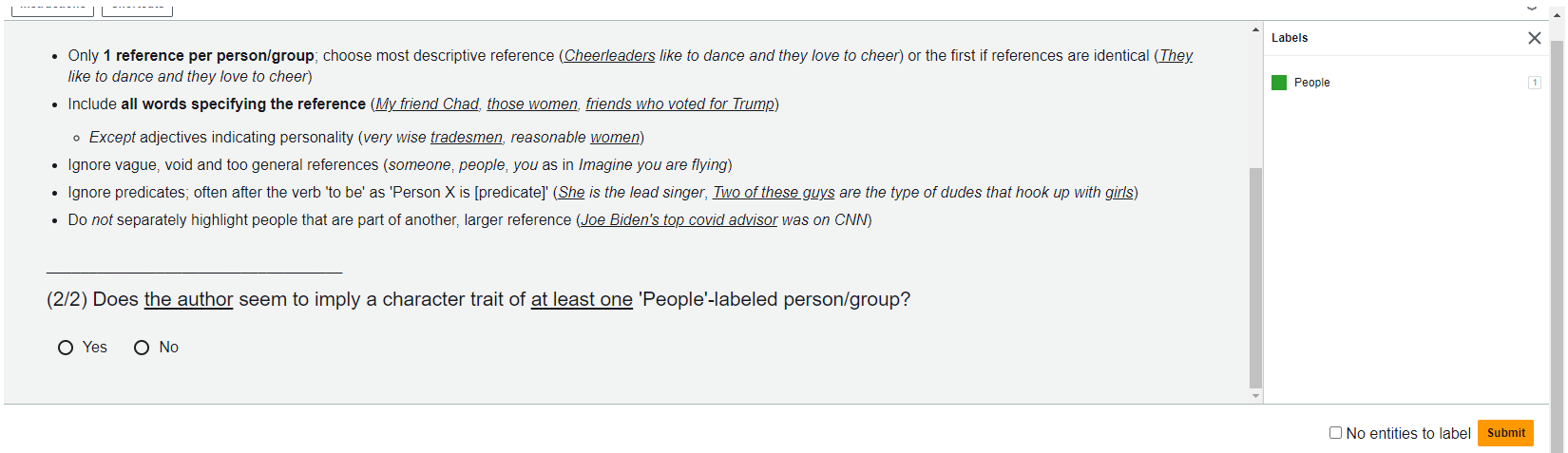}
         \label{fig:three sin x}
     \end{subfigure}
     \caption{Annotation interface for annotation round 1. Annotators perform two tasks: Mark the entities in the given sentence (\$\{sentence\}\$) using the label provided on the right (People, green label) and indicate whether the author of the sentence implies a character trait of at least one of the marked entities. The title of the blog post (\$\{title\}\$) from which the sentence was taken is given as context surrounding the sentence.}
     \label{fig:main-interface-round-1}
\end{figure}

\subsection{Task 1: Mark People Entities in a Sentence} \label{mark_entity}

The annotators mark the people and groups of people mentioned in the sentence (Figure \ref{fig:main-interface-round-1}). If there are no entities to label, they can tick the `No entities to label' box at the bottom right of the interface and submit the task form. The title of the blog post from which the sentence was taken is given below the sentence to contextualize its content. The general rules for marking people entities are given below the sentence and title such that the annotators were reminded of them when performing the task. We provide a more in-depth discussion of the rules with illustrative examples under the instructions tab at the top left corner of the interface. 



\begin{figure}[ht!]
     \centering
     \begin{subfigure}[b]{\textwidth}
         \centering
         \includegraphics[width=\textwidth]{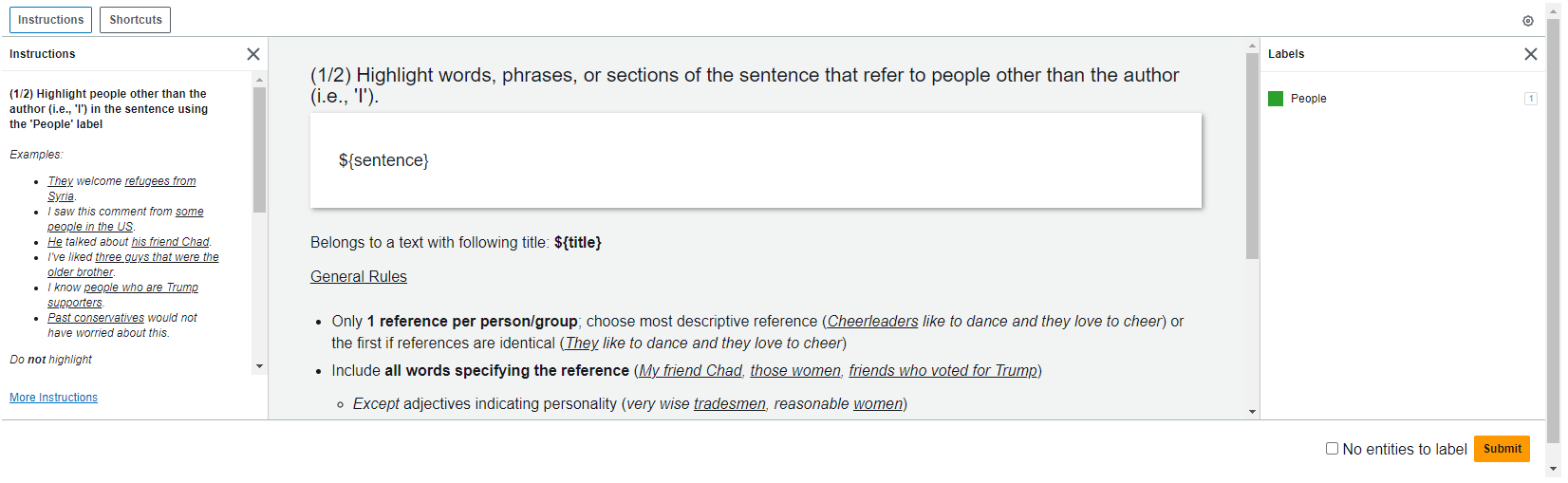}
     \end{subfigure}
     \hfill
     \begin{subfigure}[b]{\textwidth}
         \centering
         \includegraphics[width=\textwidth]{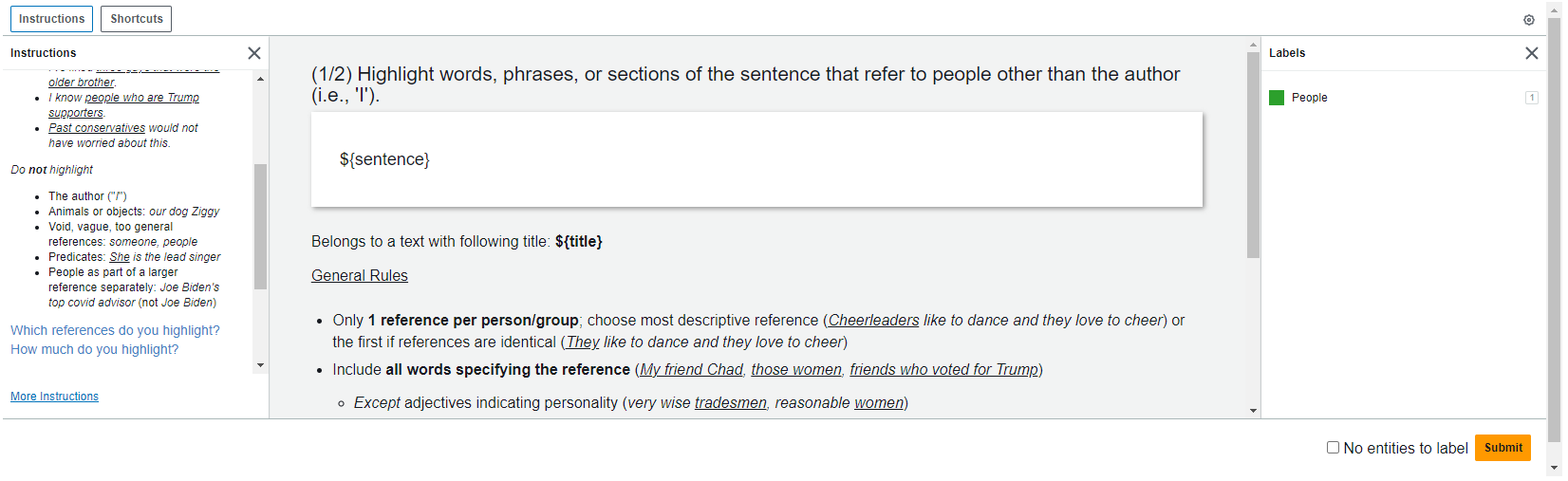}
     \end{subfigure}
     \caption{The instructions for marking the people entities in the given sentence can be opened when clicking on the instructions tab at the top left. They give a few short example sentences and list which references should not be highlighted. Additional information about which references to highlight and how much of the reference needs to be highlighted can be accessed by clicking on the clickable blue sentences in the instructions.}
     \label{fig:round1-instructions}
\end{figure}

\begin{figure}[ht!]
     \centering
     \begin{subfigure}[b]{\textwidth}
         \centering
         \includegraphics[width=\textwidth]{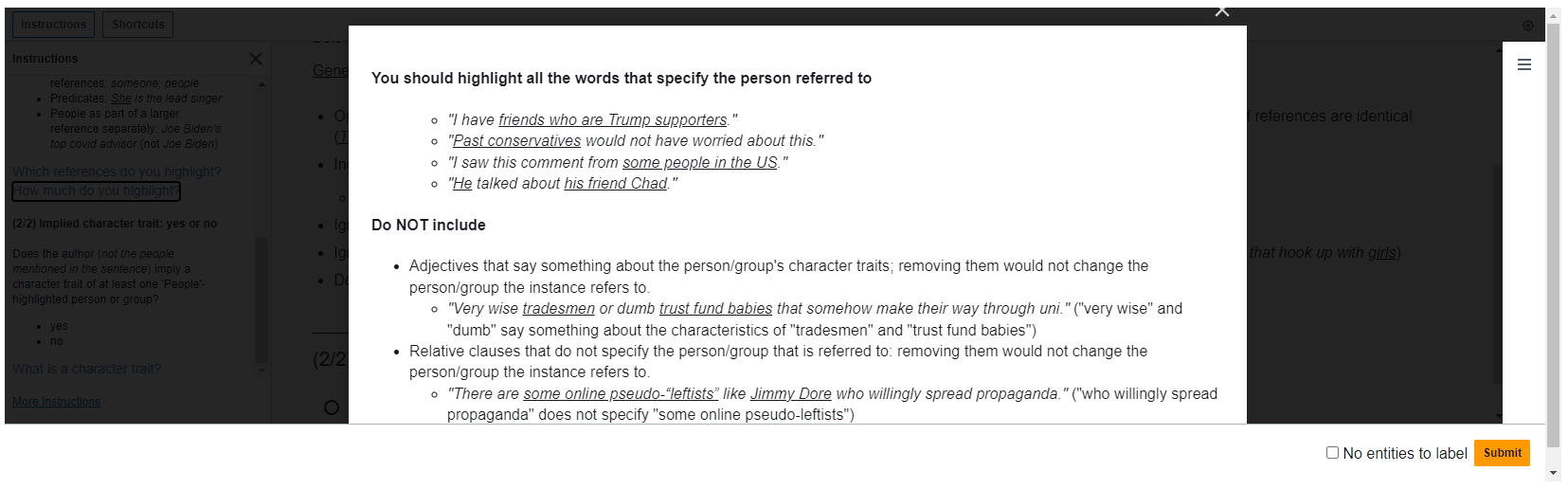}
         \caption{Pop-up screen when clicking on the instructions link ``How much to highlight?''}
     \end{subfigure}
     \hfill
     \begin{subfigure}[b]{\textwidth}
         \centering
         \includegraphics[width=\textwidth]{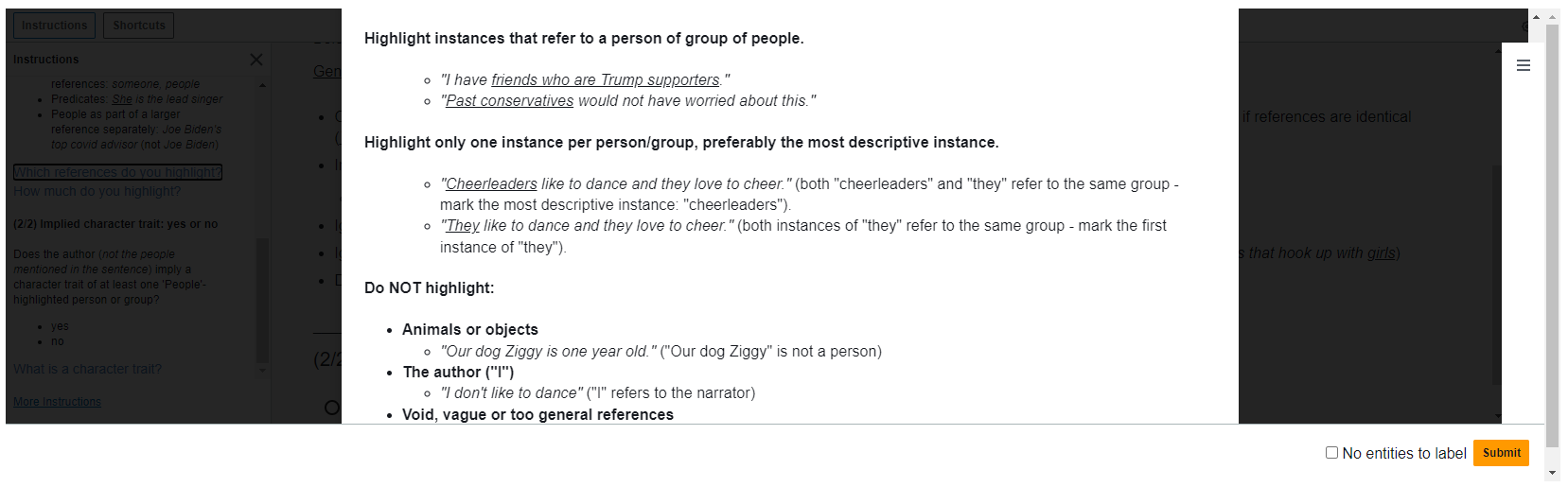}
     \end{subfigure}
     \begin{subfigure}[b]{\textwidth}
         \centering
         \includegraphics[width=\textwidth]{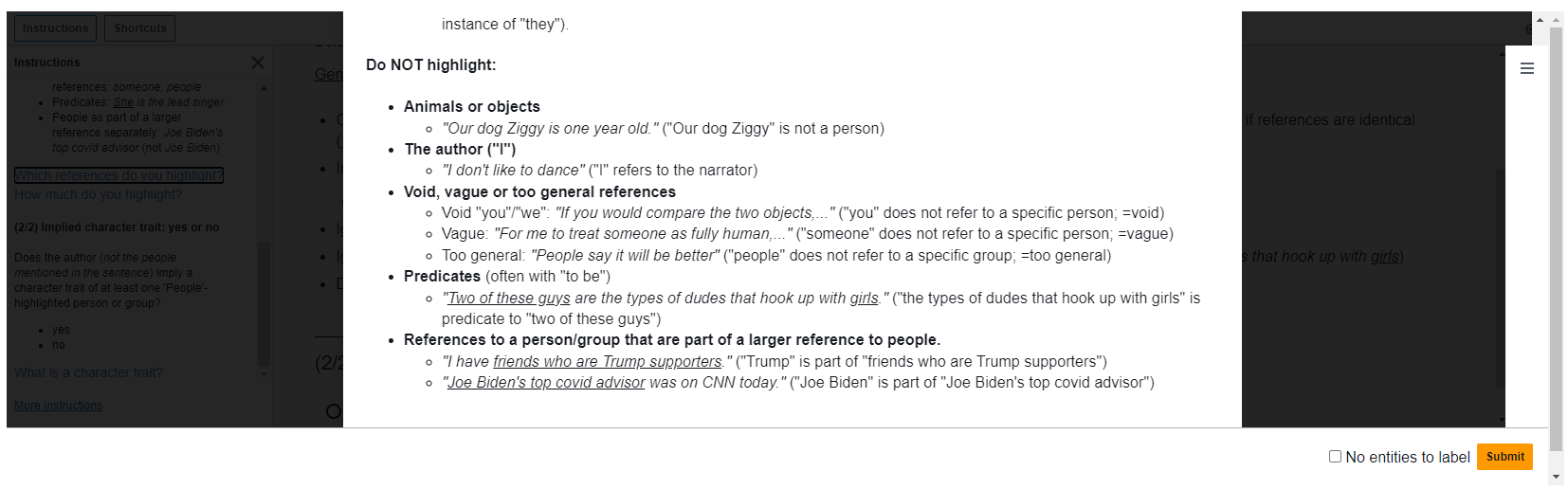}
         \caption{Pop-up screens when clicking on the instructions link ``Which references do you highlight?''}
     \end{subfigure}
     \caption{The pop-up instruction when clicking on the clickable blue links provided in the instructions.}
     \label{fig:round1-clickable-links}
\end{figure}

\subsubsection{\textit{Instructions}: What to Mark} \label{instruct-what-to-mark}

\begin{itemize}
    \item \textbf{Mark words and phrases that refer to a \textbf{person or a group of people}}
Only mark words that refer to people or a group of people such as an organisation. Examples:
\begin{itemize}
    \item \textit{\underline{They} welcome \underline{refugees from the Ukraine}.}
    \item \textit{I saw this comment from \underline{some people in the US}.}
    \item \textit{\underline{He} talked about \underline{his friend Chad}.}
    \item \textit{I've liked \underline{three different guys that were the older brother}.}
    \item \textit{I have \underline{friends who are Trump supporters}.}
    \item \textit{\underline{Past conservatives} would not have worried about this.}
    \item \textit{\underline{NATO} should be taking more responsibility.}
\end{itemize}
    \item \textbf{Mark only one reference per person/group in each sentence}
If there are more than one instance that refers to the same person or group, mark the most descriptive reference or, in case they are identical, the first one.
\begin{itemize}
    \item \textit{\underline{Cheerleaders} like to dance and they love to cheer.} (both `cheerleaders' and `they' refer to the same group - mark the most descriptive instance).
    \item \textit{\underline{They} like to dance but they hate to sing.} (both instances of `they' refer to the same group - mark the first instance).
\end{itemize}
    \item \textbf{\textbf{Mark the most descriptive reference} to a person or group} In case the same entity is referred to more than once in the sentence, mark the reference that is most descriptive.
Example:
\begin{itemize}
    \item \textit{\underline{Cheerleaders} like to dance and they love to cheer.} (Mark `cheerleaders' as it is more descriptive than `they')
\end{itemize}
    \item \textbf{Do not mark animals or objects} Example:
\begin{itemize}
    \item \textit{Our dog Ziggy is one year old.} (`Our dog Ziggy' is not a person).
\end{itemize}
    \item \textbf{Do not mark the author (`I')} Example: 
\begin{itemize}
    \item \textit{I saw this comment from \underline{some people in the US}.} (`I' refers to the narrator)
\end{itemize}
    \item \textbf{Do not mark void, vague, or too general references} Examples:
\begin{itemize}
    \item Void `you'/`we'. Example: \textit{If you would compare the two objects.} (`you' does not refer to a specific person; void); \textit{Yet, we are seeing mask mandates pop up around the country.} (`we' does not refer to a specific group; void).
    \item Vague. Example: \textit{I don't mean to say that certain people with particular sibling orders have less empathy, but I've
noticed some key differences in romantic relationships.} (`certain people with particular sibling orders' is too vague); \textit{For me to treat someone as fully human,...} (`someone' is too vague)
    \item Too general. Example: \textit{People say that it will be better.} (`people' does not refer to a specific group of people).
\end{itemize}
    \item \textbf{Do not mark predicates} Predicates are often preceded by verbs such as `to be'. Example: 
\begin{itemize}
    \item \textit{\underline{Two of these guys} are the types of dudes that hook up with \underline{girls}.} (`the types of dudes that hook up with girls' is predicate to `two of these guys')
\end{itemize}
    \item \textbf{Do not mark references to a person/group that are part of phrase referring to other people/groups} Example: 
\begin{itemize}
    \item \textit{I have \underline{friends who are Trump supporters}} (`Trump' is part of another reference, namely `friends who are Trump supporters')
    \item \textit{\underline{Joe Biden's top covid advisor} was on CNN today.} (`Joe Biden' is part of `Joe Biden's top covid advisor')
\end{itemize}
\end{itemize}



\subsubsection{\textit{Instructions}: How Much to Mark} \label{instruct-how-much-to-mark}

\begin{itemize}
    \item \textbf{Mark all the words that specify the person referred to}
Example:
\begin{itemize}
    \item \textit{\underline{They} welcome \underline{refugees from the Ukraine}.} (marking only `refugees' would be incomplete as it then refers to a larger, more general group).
    \item \textit{I saw this comment from \underline{some people in the US}.} (including `some' and `in the US' is more specific than leaving one of the two, or both, out).
    \item \textit{\underline{He} talked about \underline{his friend Chad}.} (including `his friend' provides more information on the relation between `he' and `Chad').
    \item \textit{I've liked \underline{three different guys that were the older brother}.} (including the relative clause `that were the older brother' specifies the identity of the guys).
    \item \textit{I have \underline{friends who are Trump supporters}.} (including the relative clause `who are Trump supporters' specifies the identity of the writer's friends).
    \item \textit{\underline{Past conservatives} would not have worried about this.} (including adjective `past' is necessary to refer to the right group of conservatives).
\end{itemize}
    \item \textbf{Except}
\begin{itemize}
    \item Adjectives preceding the person/group noun that denote their personality/character traits; they can be removed without changing the person or group they refer to.
    \begin{itemize}
        \item \textit{Very wise \underline{tradesmen} or dumb \underline{trust fund babies} that somehow make their way through uni.} (`very wise' and `dumb' describe character traits).
        \item \textit{Although I'm sure that there are \underline{people who never go to university} that are far smarter than I am, on the whole it would be far more effective for only reasonable intelligent and conscientious \underline{people} (people $\Rightarrow$ people who went to university) to decide on the future of \underline{our society}.} (`reasonable intelligent and conscientious' describe character traits).
    \end{itemize}
    \item Relative clauses that do not specify the kind of person/group is referred to; leaving them out would not change the reference.
    \begin{itemize}
        \item \textit{There are \underline{some online pseudo-``leftists"} like \underline{Jimmy Dore} who willingly spread propaganda.} (`who willingly spread propaganda' does not specify the reference.)
    \end{itemize}
\end{itemize}
\end{itemize}

\subsection{Task 2: Indicate Presence or Absence of Implied Character Trait} \label{indicate-presence-trait}

After labeling the entities in the sentence, the annotators now indicate whether or not the author of the sentence seems to imply a character trait of at least one entity they have marked in the sentence. 



\subsubsection{\textit{Instructions}: Definition of Character Trait} \label{instruct-definition-character-trait}

We provide the annotators with a definition of character trait in the instructions, which can be accessed by clicking on the instructions tab on the top left corner. 

\begin{figure*}[ht!]
    \centering
    \begin{subfigure}[b]{\textwidth}
         \centering
         \includegraphics[width=\textwidth]{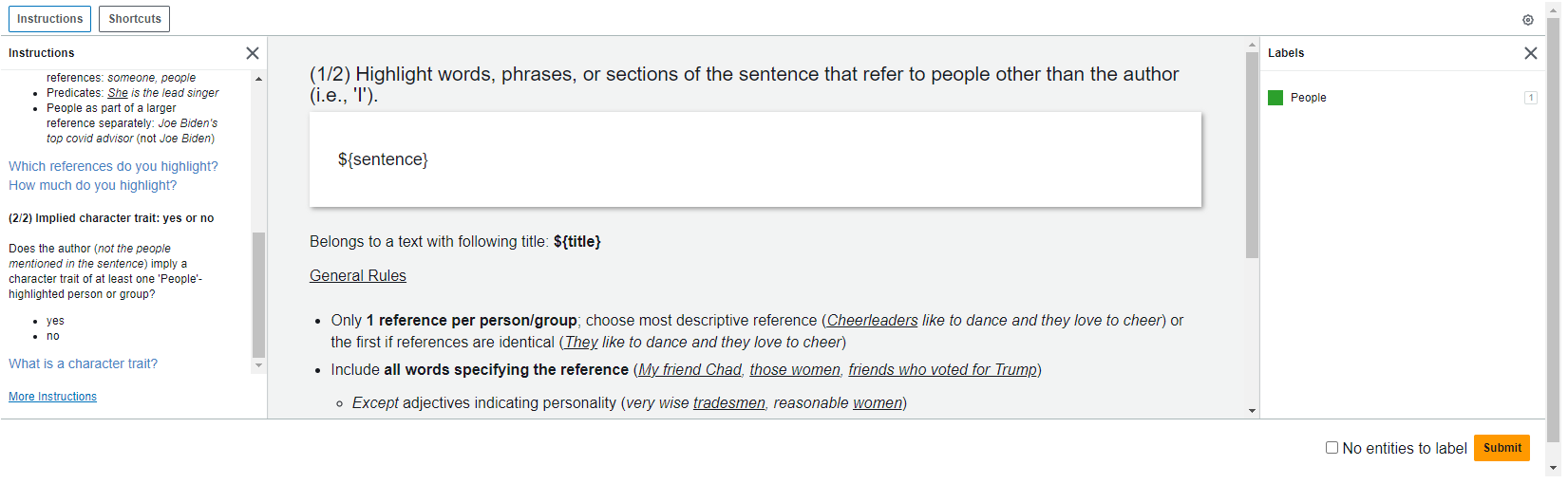}
         \label{fig:instructions-task-2}
         \caption{Annotator instructions regarding Task 2.}
     \end{subfigure}
     \hfill
     \begin{subfigure}[b]{\textwidth}
         \centering
         \includegraphics[width=\textwidth]{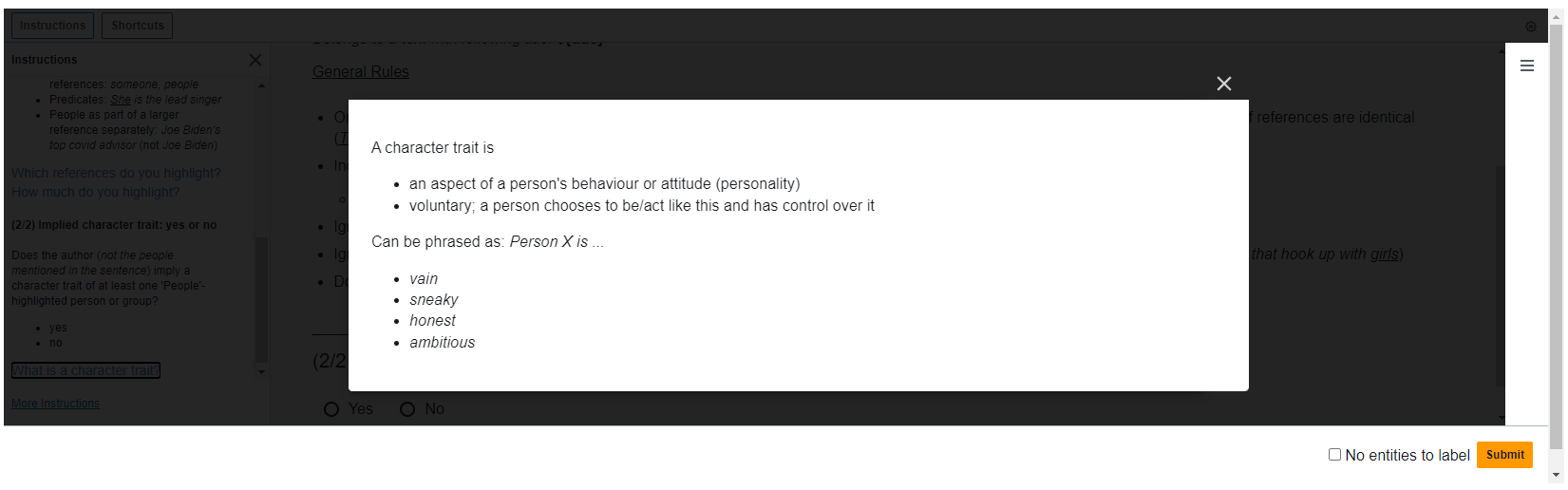}
         \label{fig:instructions-definition-character-trait}
         \caption{Pop-up screen when clicking on the instructions link ``What is a character trait?''.}
     \end{subfigure}
    \caption{Instructions regarding Task 2 (a) and definition of character trait (b)}
    \label{fig:enter-label}
\end{figure*}

A character trait describes an aspect of a person's \textbf{behavior} or \textbf{attitude}. The set of character traits then forms someone's personality. A trait is also \textbf{voluntary} so it is not forced on that person by someone or something. A person has control over his/her character traits. A person's character traits cause him/her to act a certain way. Character traits can very often be described using descriptive adjectives such as \textit{honest, sincere, brave, loyal, vain, timid}, and \textit{calm}.
Examples:
\begin{itemize}
    \item \textit{Eric disguises himself.} (Eric is sneaky).
    \item \textit{Eric likes to talk about himself.} (Eric is vain).
    \item \textit{Eric does not want to spend much money.} (Eric is stingy).
\end{itemize}

\subsubsection{\textit{Instructions}: The Author Implies Trait of at Least One Entity}

The annotators should ask themselves whether the author seems to suggest something about the character of the people in the sentence. It is important to distinguish between character traits implied by the author, showing that the author seems to hold certain beliefs about people, and character traits that a reader or other people mentioned by the writer belief people have (a reader or mentioned person holds certain beliefs about people). Like any other person, an annotator can have certain, pre-existing beliefs about (groups of) people or can be aware of pre-existing beliefs that others have about them. We therefore stress in the phrasing of the second task that the annotators should think about what the author seems to imply by underlining \underline{the author}: \textit{``Does \underline{the author} seem to imply a character trait of \underline{at least one} `People'-labeled person/group?''}. We also underline \underline{at least one} to stress that a character trait for all entities need to be implied for indicating `yes'.


\subsection{Qualification Test: Round 1} \label{round-1-qualification}

Candidate annotators first have to pass a qualification test before they can start the annotation round. They were automatically admitted to the annotation round if they answered all questions in the qualification test correctly. The qualification test instructs the annotators on how to adequately mark people entities in a sentence by going over five main rules. For each rule, the annotator is presented with an example sentence. They then need to pick the correct labeling answer from the options below the sentence (radio button). We indicate the correct answer with \textbf{(X)}.
\\ \\
Test description: \textit{We go over FIVE labeling rules for this task. For each rule, you will be presented with an example sentence and you need to pick the correct labeling answer from the options below. If you pick all the correct answers, you can automatically start the labeling task.}
\begin{itemize}
    \item Rule 1: Highlight all words, phrases, or sections of the sentence that refer to people other than the author (i.e., 'I').
    \begin{itemize}
        \item Sentence: "I talked with my friend Chad about some men from Texas and people who like Star Wars."
        \begin{itemize}
            \item I - Chad - men - people
            \item \textbf{(X)} my friend Chad - some men from Texas - people who like Star Wars
        \end{itemize}
    \end{itemize}
    \item Rule 2: Highlight only one reference per person/group + choose most descriptive reference.
    \begin{itemize}
        \item Sentence: "Cheerleaders like to dance and they love to cheer."
        \begin{itemize}
            \item \textbf{(X)} Cheerleaders
            \item Cheerleaders - they
            \item they
        \end{itemize}
    \end{itemize}
    \item Rule 3: Highlight all words specifying the reference; except adjectives indicating personality (e.g., reasonable, vain).
    \begin{itemize}
        \item Sentence: "The blond girl from Canada told that there are honest girls and generous boys in her class."
        \begin{itemize}
            \item girl - girls - boys
            \item The blond girl from Canada - honest girls - generous boys
            \item \textbf{(X)} The blond girl from Canada - girls - boys
        \end{itemize}
    \end{itemize}
    \item Rule 4: Do not highlight vague, void and too general references.
    \begin{itemize}
        \item Sentence: "If you meet someone online, people will tell you to be cautious"
        \begin{itemize}
            \item you - someone - people
            \item \textbf{(X)} [None] (you = void, someone = vague, people = too general)
        \end{itemize}
    \end{itemize}
    \item Rule 5: Do not highlight predicates (often after a modal verb like 'to be': 'Person is [predicate]') and people that are part of another, larger reference
    \begin{itemize}
        \item Sentence: "Barack Obama became the US president when he was 48 and Barack Obama's wife was the First Lady"
        \begin{itemize}
            \item \textbf{(X)} Barack Obama - Barack Obama's wife
            \item Barack Obama - the US president - wife - the First Lady
        \end{itemize}
    \end{itemize}
\end{itemize}

\section{Annotation Guidelines: Round 2} \label{app:annotation-round-2}

\begin{figure*}[ht!]
    \centering
    \begin{subfigure}[b]{\textwidth}
         \centering
         \includegraphics[width=\textwidth]{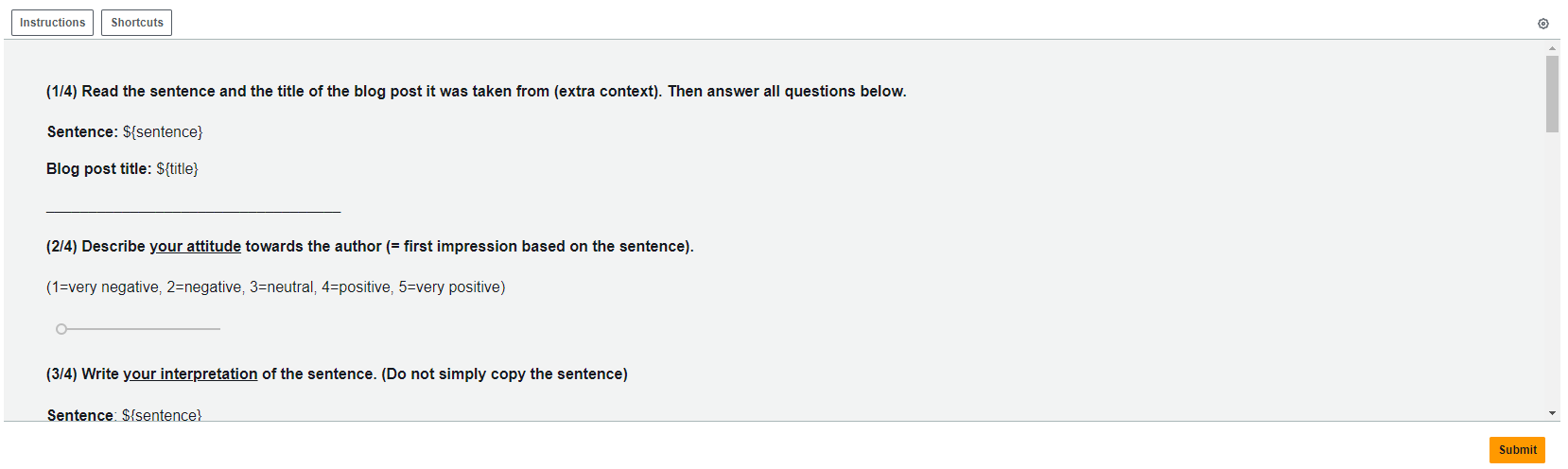}
     \end{subfigure}
     \hfill
     \begin{subfigure}[b]{\textwidth}
         \centering
         \includegraphics[width=\textwidth]{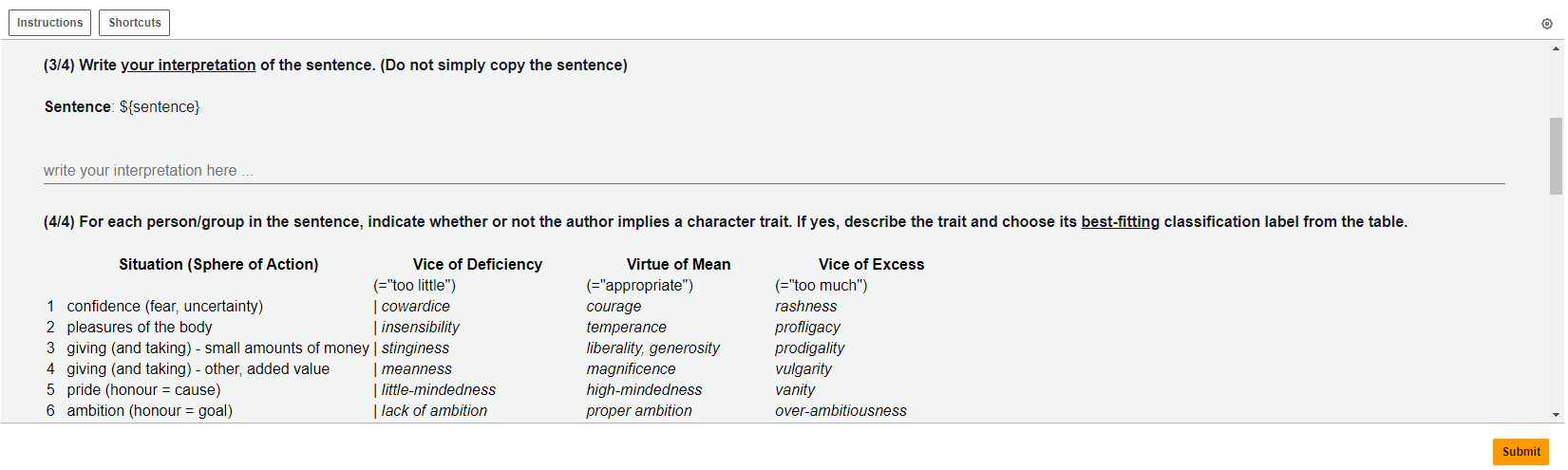}
     \end{subfigure}
     \hfill
     \begin{subfigure}[b]{\textwidth}
         \centering
         \includegraphics[width=\textwidth]{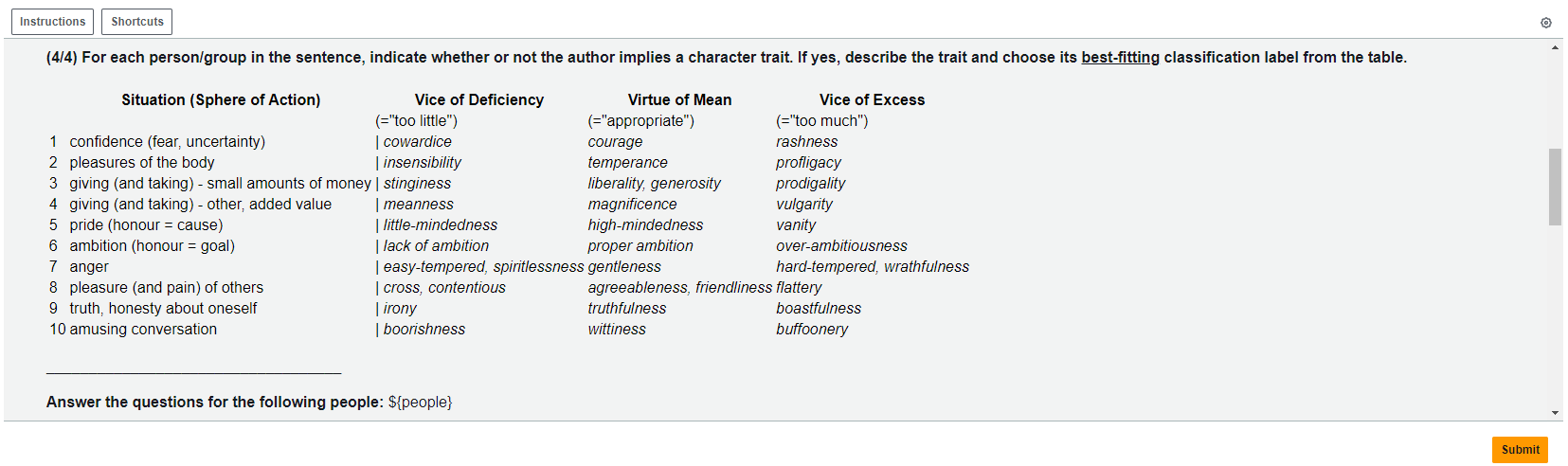}
     \end{subfigure}
     \hfill
     \begin{subfigure}[b]{\textwidth}
         \centering
         \includegraphics[width=\textwidth]{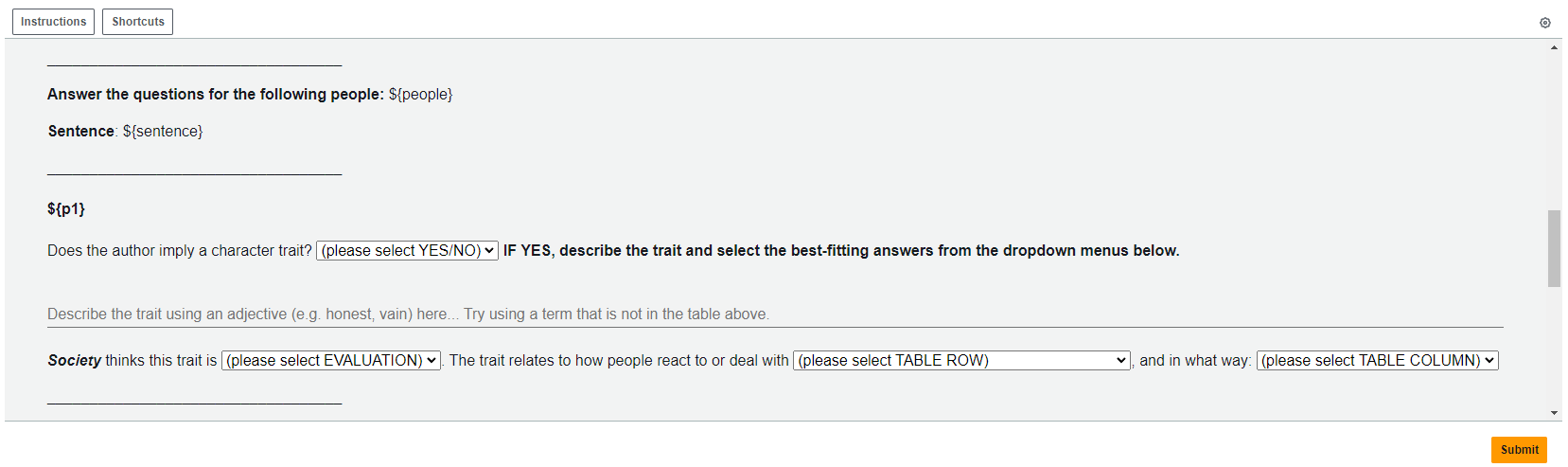}
     \end{subfigure}
    \caption{The main annotation interface of ANNOTATION ROUND 2.}
    \label{fig:main-round-2}
\end{figure*}

\begin{figure*}[ht!]
    \centering
    \begin{subfigure}[b]{\textwidth}
         \centering
         \includegraphics[width=\textwidth]{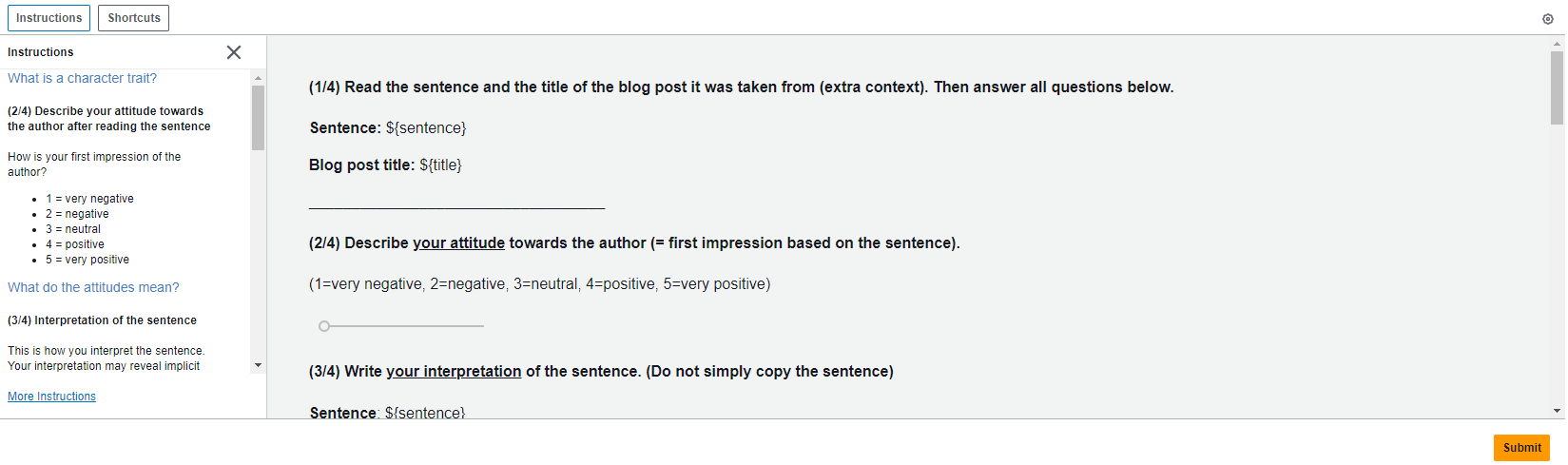}
     \end{subfigure}
     \hfill
     \begin{subfigure}[b]{\textwidth}
         \centering
         \includegraphics[width=\textwidth]{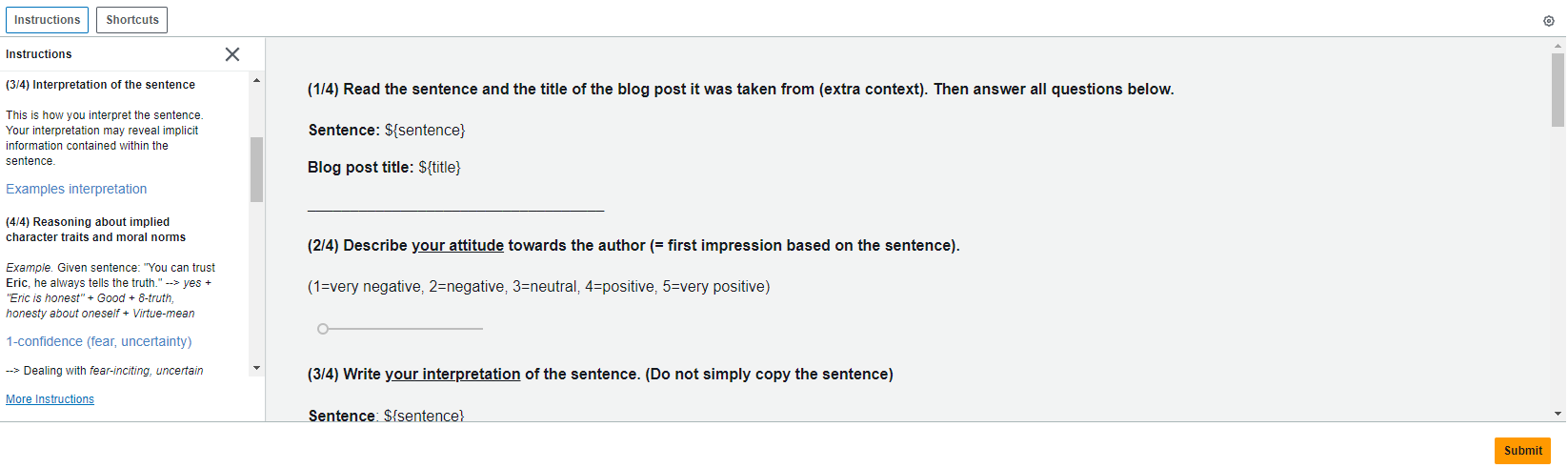}
     \end{subfigure}
     \hfill
     \begin{subfigure}[b]{\textwidth}
         \centering
         \includegraphics[width=\textwidth]{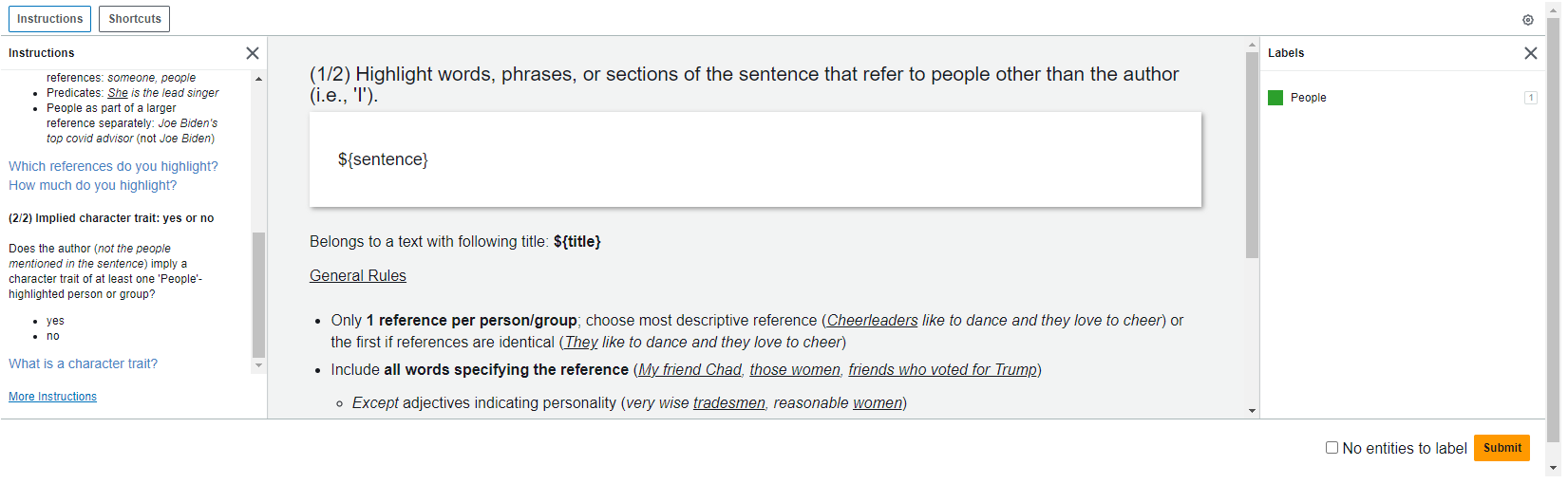}
     \end{subfigure}
     \hfill
     \begin{subfigure}[b]{\textwidth}
         \centering
         \includegraphics[width=\textwidth]{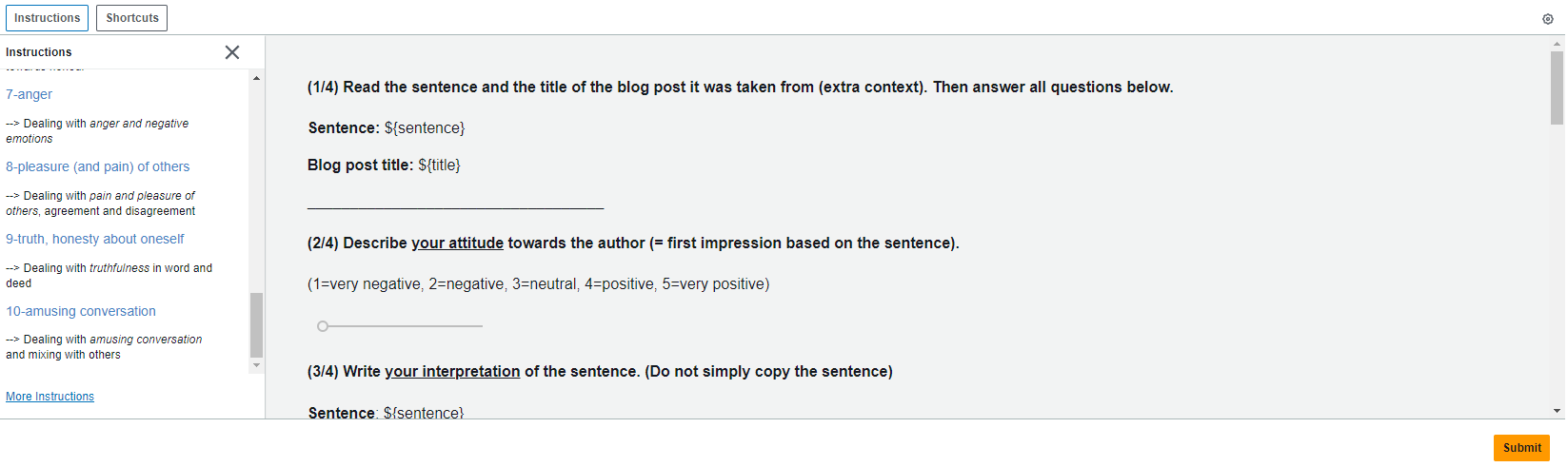}
     \end{subfigure}
    \caption{Overview of the instructions that appear when clicking on the instructions tab at the top left of the annotation interface.}
    \label{fig:main-instructions-2}
\end{figure*}

The annotators start by reading the sentence (\$\{sentence\}) and the blog post from which the sentence is taken (\$\{title\}). They are then asked to describe their attitude towards the author and formulate their interpretation of the given sentence. Going over all people identified in the sentence, they indicate for each entity whether or not the author of the sentence implies a character trait of that entity. For all entities with an implied character trait, the annotators also describe the trait, label the evaluation of that trait in society, and classify it in Virtue Ethics by labeling the Sphere of Action the trait belongs to and its contextual appropriateness. It is made clear in the interface (Figure \ref{fig:main-round-2}) that the annotators need to fulfill four tasks, marked by (1/4) to (4/4). They can access the instructions with illustrative examples at any time during annotation by clicking on the instructions tab at the top left corner of the interface (Figure \ref{fig:main-instructions-2}). The instructions then appear on the left and can stay visible during the annotation process. They disappear when clicking on the instructions tab again. The annotators are unable to submit the task form without providing an interpretation, which is the most important part of this annotation round. Candidate annotators had to pass a qualification test to have access to the annotation round.

\subsection{Task 1: Describe Attitude Towards Author of Sentence} \label{describe-attitude}

\begin{figure}[!htbp]
    \centering
    \includegraphics[width=7cm]{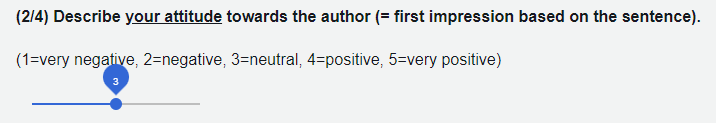}
    \caption{Attitude slider shows attitude values on a five point Likert scale.}
    \label{fig:attitude-slider}
\end{figure}

The annotators describe their attitude towards the author of the sentence upon reading the sentence. We stress in the task description that they should describe their own attitude by underlining \underline{\textit{your attitude}} in the task description and state in brackets that their attitude is their impression of the author based on the sentence. They change the slider given below the task description and set it to the best-fitting attitude on a five-point Likert scale ranging from \textit{very negative} tot \textit{very positive}. When moving the slider, the value to which they set it changes to blue and depicts the number referring to the attitude they set it to (Figure \ref{fig:attitude-slider}). 

\subsubsection{\textit{Instructions}: Definitions of Attitude Scale} \label{instruct-definitions-attitudes}

We provide the annotators with a definition of each attitude type in the instructions. They can access the overview of attitude definitions by clicking on the clickable blue sentence \textit{``What do the attitude mean?''} in the instructions. The overview then appears as a pop-up screen.

\begin{itemize}
    \item \textbf{Very negative} After reading the sentence, the annotator has a very negative view of the author. They completely disagree with what the author said. They would never say or think the same, or never phrase it like the author did. 
    \item \textbf{Negative} After reading the sentence, the annotator has a negative view of the author. They disagree with what the author said. They would not say or think the same, or phrase it like the author did. 
    \item \textbf{Neutral} After reading the sentence, the annotator does not have a negative of positive view of the author. They do not agree of disagree with what the author said. 
    \item \textbf{Positive} After reading the sentence, the annotator has a positive view of the author. They agree with what the author said. They would say or think the same, or phrase it like the author did. 
    \item \textbf{Very positive} After reading the sentence, the annotator has a very positive view of the author. They completely agree with what the author said. They would definitely say or think the same, or phrase it like the author did. 
\end{itemize}

\begin{figure*}[ht!]
    \centering
    \begin{subfigure}[b]{\textwidth}
         \centering
         \includegraphics[width=\textwidth]{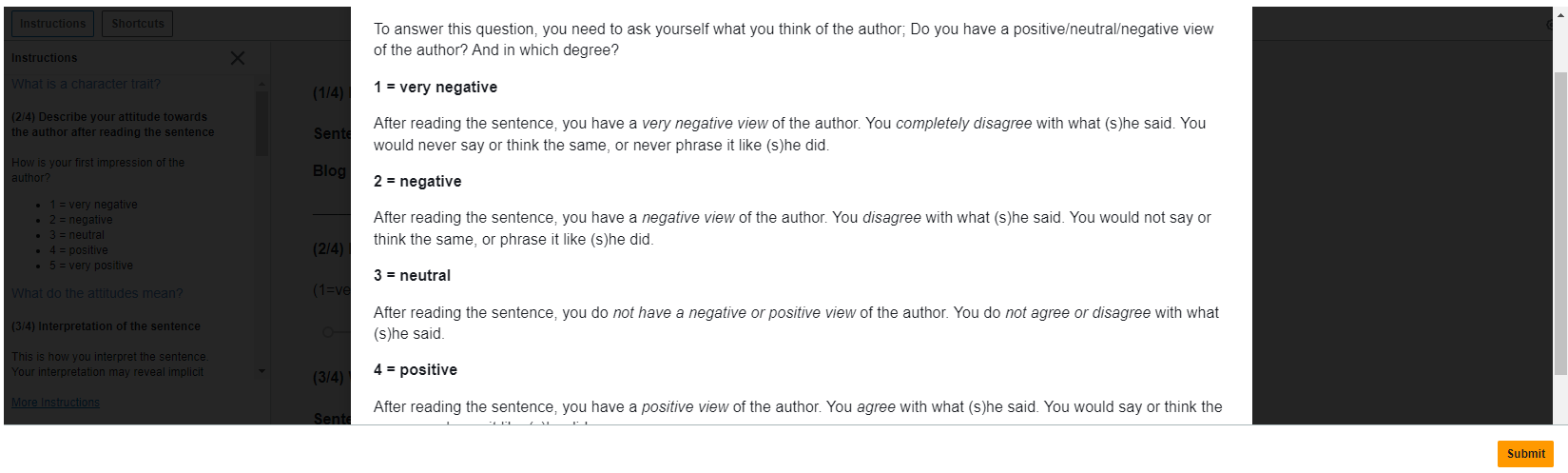}
     \end{subfigure}
     \hfill
     \begin{subfigure}[b]{\textwidth}
         \centering
         \includegraphics[width=\textwidth]{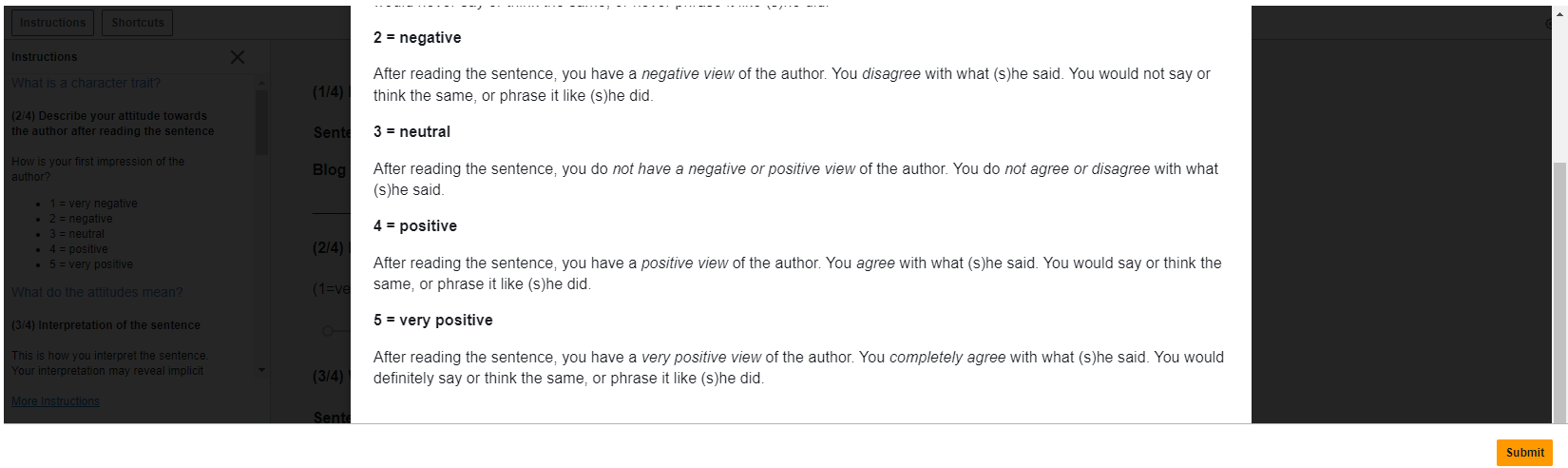}
     \end{subfigure}
    \caption{Overview of the different attitude types and their description given in the instructions.}
    \label{fig:attitudes-definitions}
\end{figure*}

\subsection{Task 2: Formulate Interpretation of Sentence} \label{fomulate-interpretation}

The annotators write their interpretation of the given sentence in their own words. We stress that they should formulate their own interpretation by underlining \underline{\textit{your interpretation}} in the task description. They are also explicitly asked to not simply copy the sentence. The instructions additionally state that your interpretation may reveal implicit information contained within the sentence. 

\subsubsection{\textit{Instructions}: Examples of Interpretations} \label{instruct-what-is-interpretation}

\begin{figure*}[ht!]
    \centering
    \includegraphics[width=\textwidth]{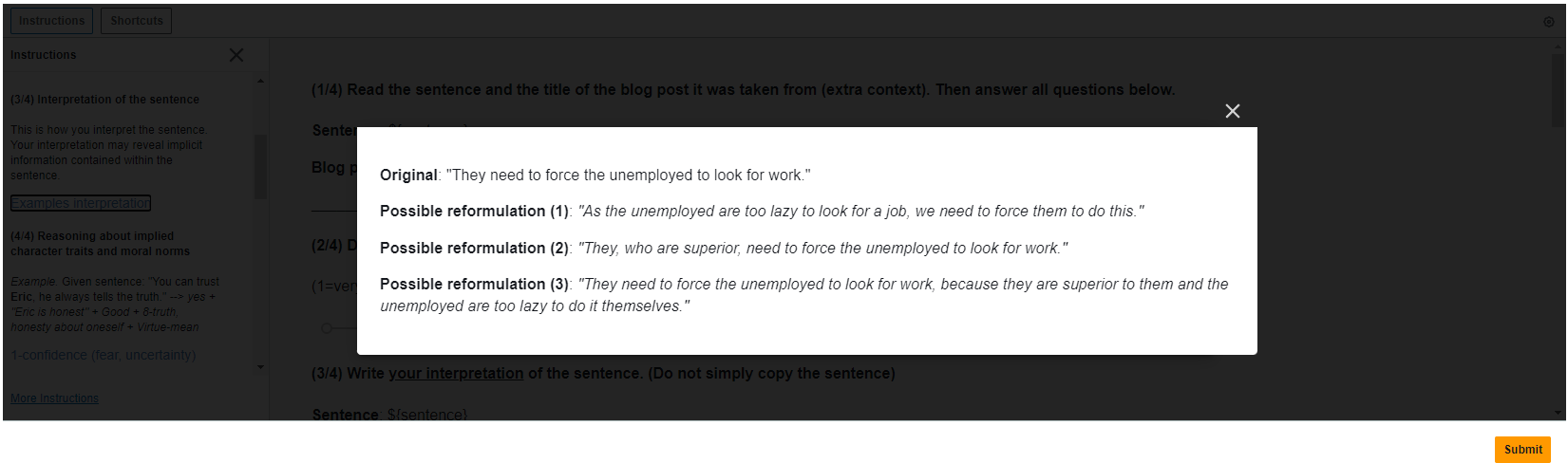}
    \caption{Examples of possible interpretations of a given sentence given in the instructions.}
    \label{fig:interpretation-examples}
\end{figure*}

We provide the annotators with several examples of possible interpretations for an example sentence, which can be accessed by clicking on the clickable blue phrase \textit{``Examples interpretations''} in the instructions (Figure \ref{fig:interpretation-examples}). The examples appear in a pop-up screen. We phrase the interpretations in such a way so that the underlying messages and hidden moral judgments surface. 
\\ \\
Given sentence: \textit{``They need to force the unemployed to look for work."}
\begin{itemize}
    \item Possible interpretation (1): \textit{``As the unemployed are too lazy to look for a job, we need to force them to do this.''}
    \item Possible interpretation (2): \textit{``They, who are superior, need to force the unemployed to look for work.''}
    \item Possible interpretation (3): \textit{``They need to force the unemployed to look for work, because they are superior to them and the unemployed are too lazy to do it themselves.''}
\end{itemize}

\subsection{Task 3: Categorize Implied Character Trait for Each Entity in Sentence} \label{categorise-entity}

The annotators need to answer the following questions for each entity that was previously marked in the first annotation round:
\begin{itemize}
    \item Does the author seem to imply a character trait of this entity? (Task 3.1)
    \item If Task 3.1 = yes:
    \begin{itemize}
        \item What is the character trait? (Task 3.2)
        \item How does society evaluate this character trait? (Task 3.3)
        \item Where is the character positioned in Virtue Ethics in terms of Sphere of Action and contextual appropriateness? (Task 3.4)
    \end{itemize}
\end{itemize}

\begin{figure*}[ht!]
    \centering
    \begin{subfigure}[b]{\textwidth}
         \centering
         \includegraphics[width=\textwidth]{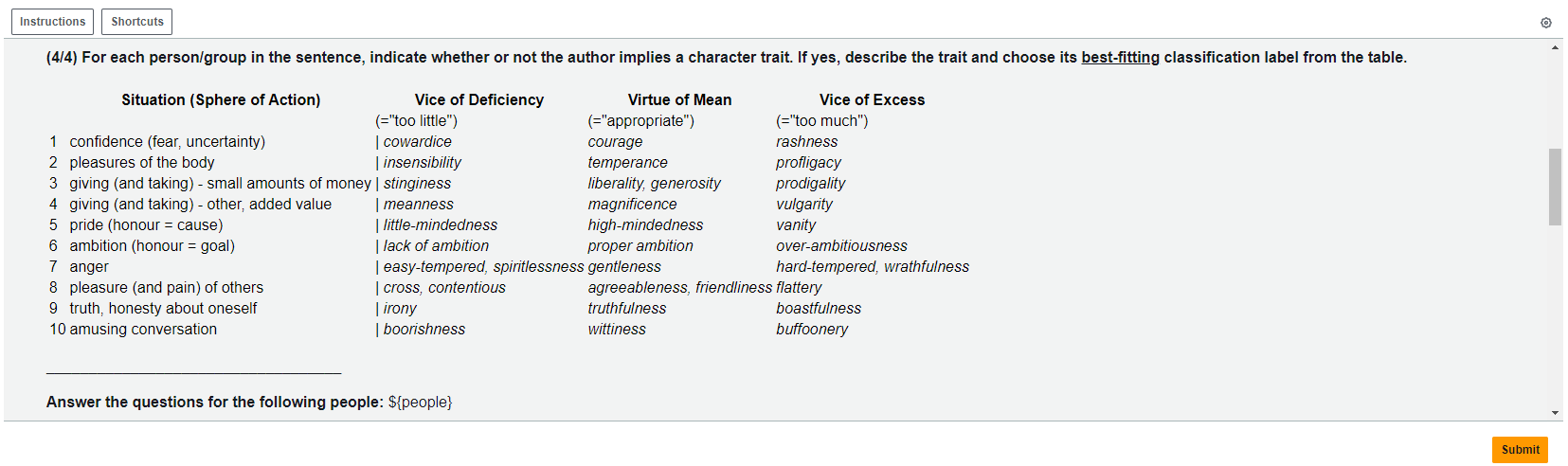}
         \caption{Virtue Ethics classification table.}
         \label{fig:virtue-ethics-table}
     \end{subfigure}
     \hfill
     \begin{subfigure}[b]{\textwidth}
         \centering
         \includegraphics[width=\textwidth]{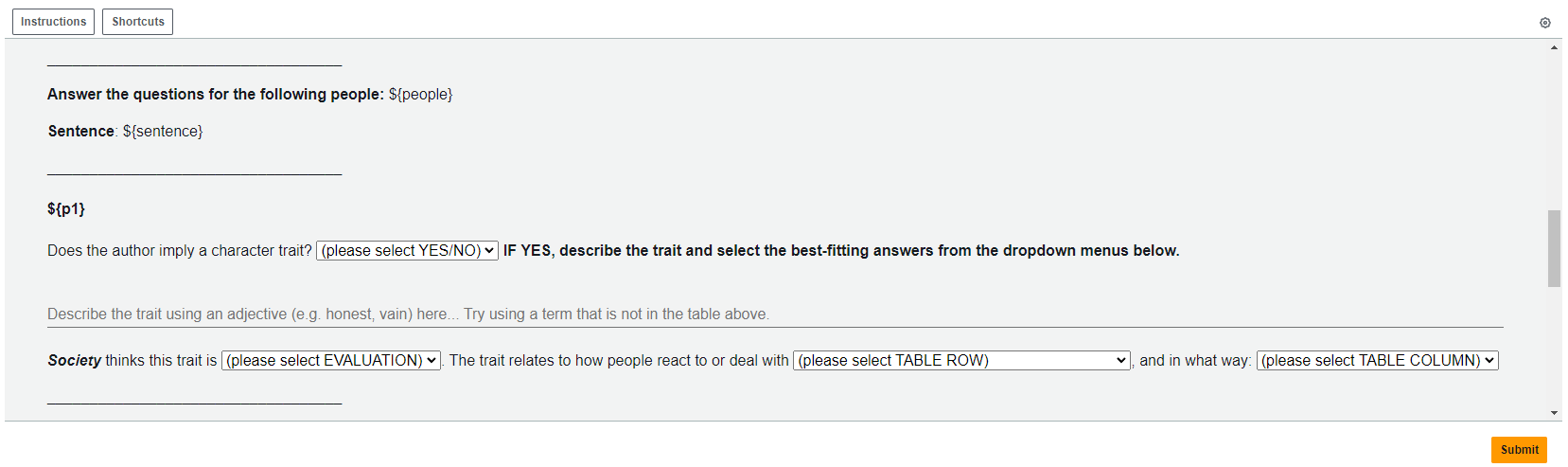}
         \caption{Entity annotation block for the first people entity (\$\{p1\}) in the people entities list (\$\{people\}).}
         \label{fig:entity-annotation-block}
     \end{subfigure}
    \caption{Interface for tackling Task 3.}
    \label{fig:task-3-interface}
\end{figure*}

The interface presents the classification table for classifying the character traits in Virtue Ethics right below the task description (Figure \ref{fig:virtue-ethics-table}). The annotators are then given the list of people entities for which they need to provide annotations (\$\{people\}). The interface then contains entity annotation blocks, with the number of blocks equal to the number of entities in the people entities list (Figure \ref{fig:entity-annotation-block}). Each block first presents the sentence such that the annotator does not need to scroll up to read the sentence again. The block then contains Task 3.1, 3.2, 3.3, and 3.4. The tasks are framed as sentences, and the annotators need to select the labels from drop-down lists.

\subsubsection{Task 3.1: Indicate Presence Implied Character Trait of Entity} \label{indicate-presence-implied-trait-2)}

\begin{figure*}[ht!]
    \centering
    \includegraphics[width=\textwidth]{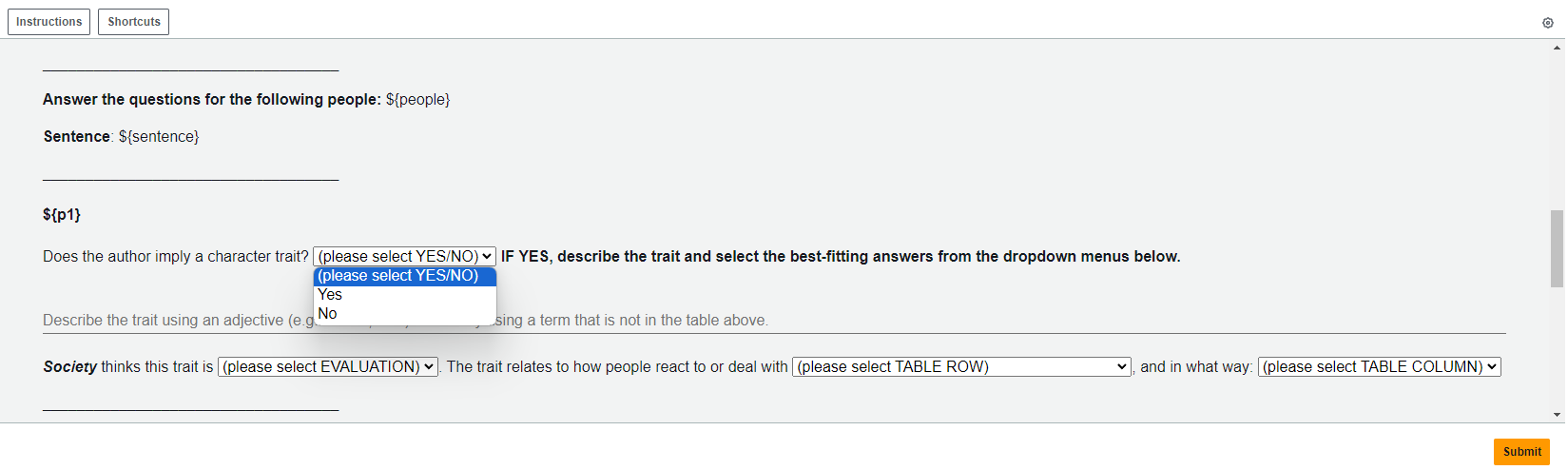}
    \caption{Drop-down list for solving Task 3.1.}
    \label{fig:entity-presence-trait}
\end{figure*}

The annotators need to indicate whether or not the author implies a character trait of the given people entity. They do this by choosing between \textit{yes} and \textit{no} from a drop-down list (Figure \ref{fig:entity-presence-trait}). If they indicate yes, they need to continue with the annotation for this entity. This is also stated in bold after the drop-down list. Otherwise, they can continue to annotate the next entity or, in case this entity is the last one, submit the task form.

\paragraph{\textit{Instructions}: Definition of Character Trait} \label{instruct-definition-trait-2}

The annotators are given the same definition of character trait given in the first annotation round, see § \ref{instruct-definition-character-trait}. The definition can be accessed by clicking on the clickable blue sentence \textit{``What is a character trait?''} in the instructions, which then appears in a pop-up screen.

\subsubsection{Task 3.2: Identify Implied Character Trait of Entity} \label{identify-trait}

\begin{figure*}[ht!]
    \centering
    \includegraphics[width=\textwidth]{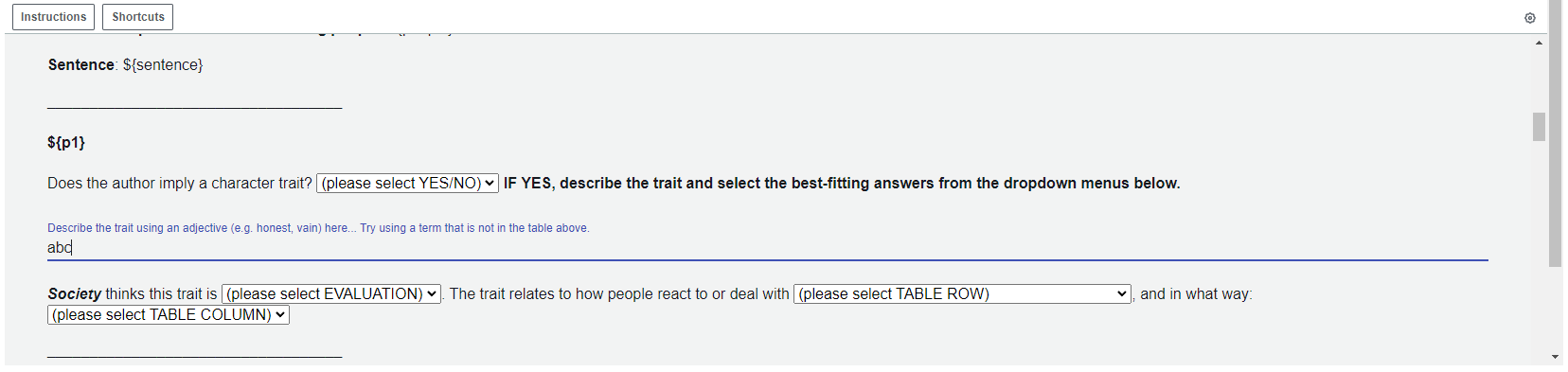}
    \caption{Input for Task 3.2. is provided as free text.}
    \label{fig:entity-describe-trait}
\end{figure*}

The annotators describe the implied character trait in their own words in the free-text text box (Figure \ref{fig:entity-describe-trait}).

\paragraph{\textit{Instructions}: How to Describe Trait} \label{instruct-how-describe-trait}

The annotators are encouraged to describe the trait using an adjective (\textit{``Describe the trait using an adjective (e.g., honest, vain) here.''}) and not to use an adjective that is presented in the Virtue Ethics classification table given above (\textit{``Try using a term that is not in the table able.''}), see Figure \ref{fig:entity-describe-trait}. When the annotators start describing the trait, the two instructions appear above the free-text box. 

\subsubsection{Task 3.3: Evaluate Implied Character Trait of Entity in Society} \label{evaluate-trait}

\begin{figure*}[ht!]
    \centering
    \includegraphics[width=\textwidth]{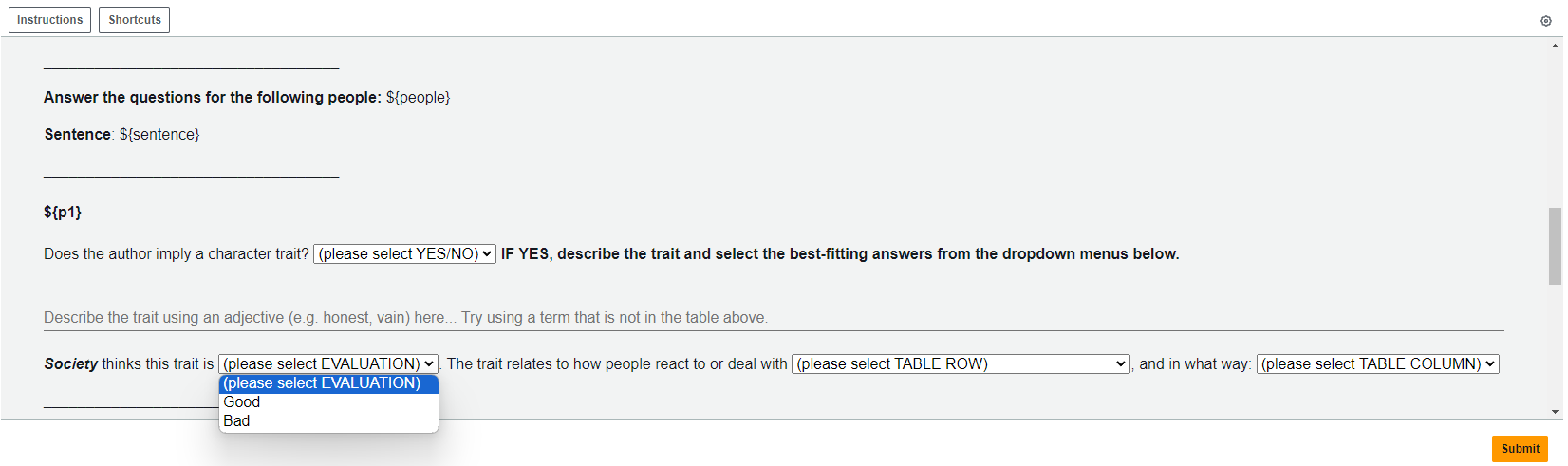}
    \caption{Drop-down list for solving Task 3.3.}
    \label{fig:entity-evaluation}
\end{figure*}

The annotators decide whether society thinks this is a \textit{good} (\textit{positive}) or \textit{bad} (\textit{negative}) trait (Figure \ref{fig:entity-evaluation}). Or in other words, does society praise people for having this trait or frown upon or even reject this trait, respectively. 

\paragraph{\textit{Instructions}: Focus on What Society Thinks} 

Since we are interested in how the character trait is evaluated in society, we mark \textit{society} in bold in the task description (\textit{``\textbf{Society} thinks this trait is''}). This way, we encourage annotators to think beyond their own evaluations of character trait and position the trait in the society they are part of.   

\subsubsection{Task 3.4: Classify Implied Character Trait of Entity in Virtue Ethics} \label{virtue-ethics-trait}

The annotators classify the implied character trait in Virtue Ethics by selecting the best-fitting labels from the drop-down lists (Figure \ref{fig:entity-soa-degree}). A table containing all the Virtue Ethics labels is given below the task description so that the annotators can easily consult the possible labels. They first classify the trait in a Sphere of Action (table rows) and then by its contextual appropriateness (table columns). When clicking on the instructions tab at the top left corner, the annotators see an overview of all the Spheres of Action with a brief description for each Sphere of Action. 

Before starting the annotation task, candidate annotators had to pass an instruction test in which the Virtue Ethics and its labels were explained and participants had to select the appropriate contextual appropriateness label for ten sentences, with each sentence related to a different Sphere of Action. The annotators were allowed to start the annotation task if they correctly labeled nine sentences.

\begin{figure*}[ht!]
    \centering
    \begin{subfigure}[b]{\textwidth}
         \centering
         \includegraphics[width=\textwidth]{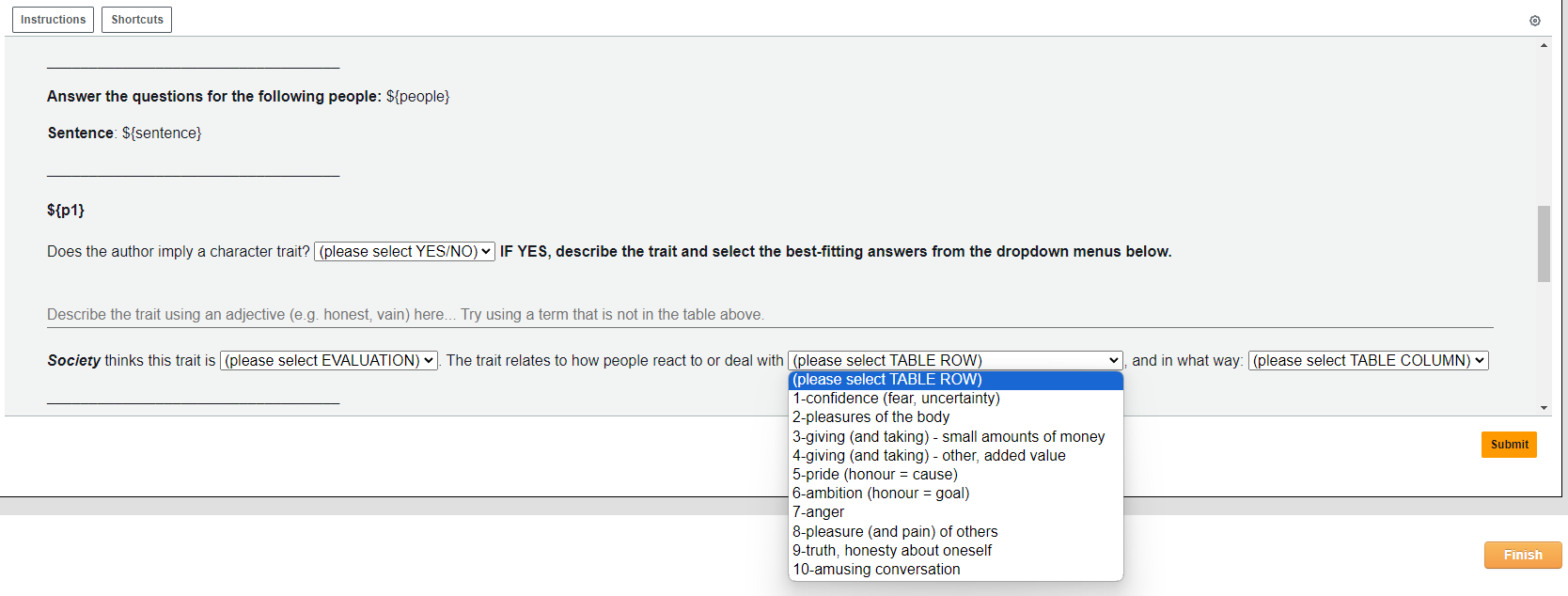}
         \caption{Drop-down list for selecting Sphere of Action labels.}
         \label{fig:entity-soa}
     \end{subfigure}
     \hfill
     \begin{subfigure}[b]{\textwidth}
         \centering
         \includegraphics[width=\textwidth]{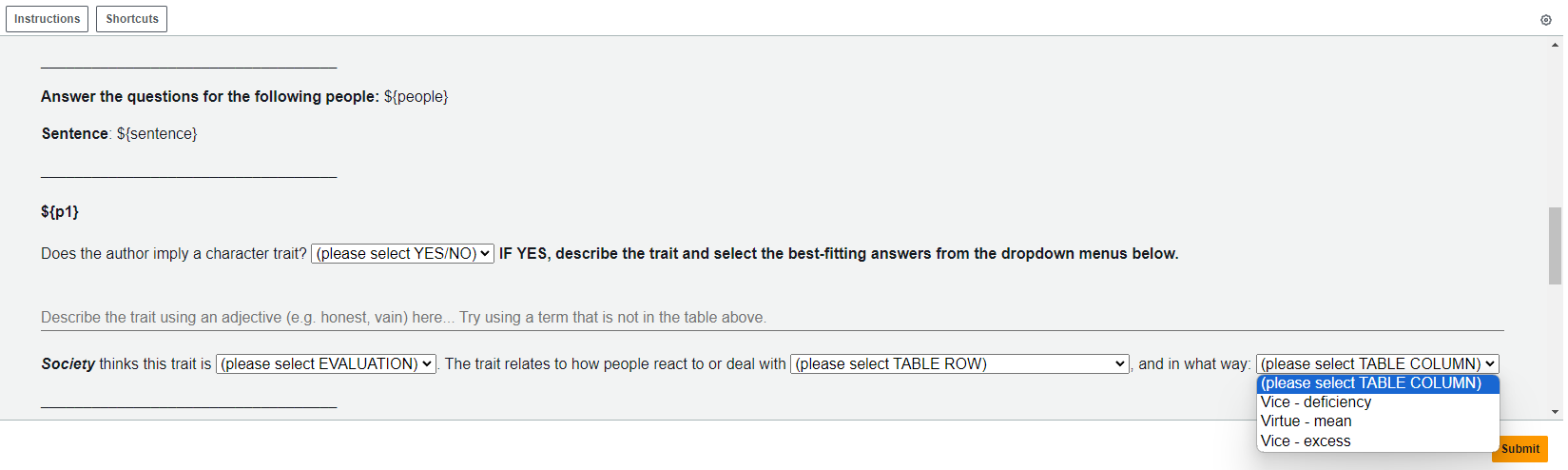}
         \caption{Drop-down list for selecting contextual appropriateness labels.}
         \label{fig:entity-degree}
     \end{subfigure}
    \caption{Drop-down lists for solving Task 3.4.}
    \label{fig:entity-soa-degree}
\end{figure*}

\begin{figure*}[ht!]
    \centering
    \begin{subfigure}[b]{\textwidth}
         \centering
         \includegraphics[width=\textwidth]{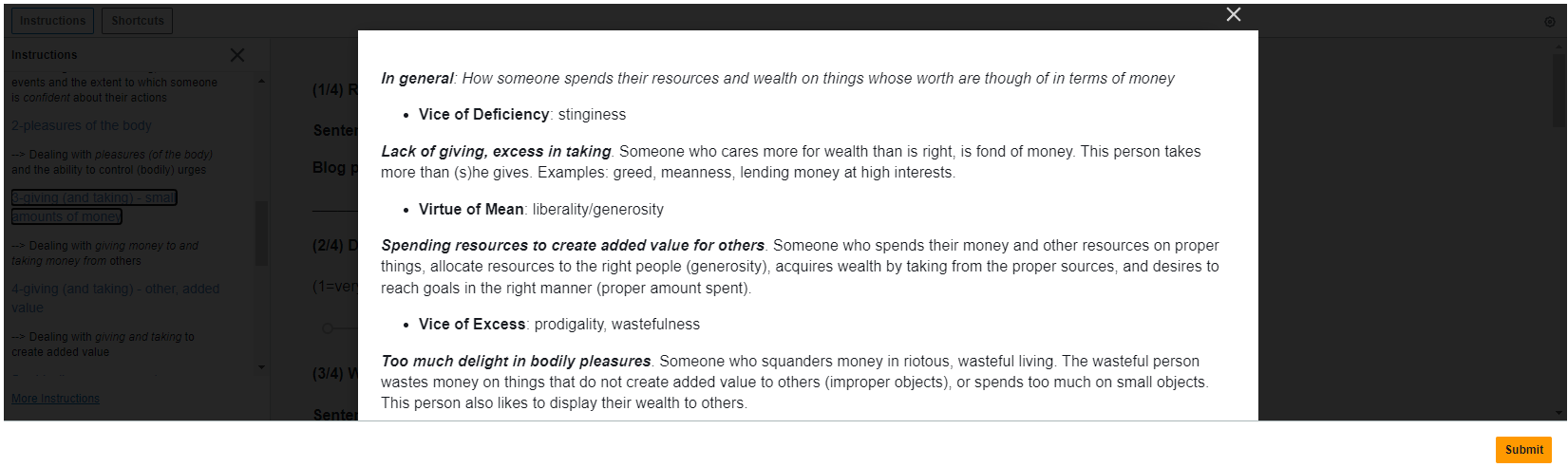}
         \caption{Definition of \textit{Giving (and taking) - small amounts of money}.}
         \label{fig:entity-giving-definition}
     \end{subfigure}
     \hfill
     \begin{subfigure}[b]{\textwidth}
         \centering
         \includegraphics[width=\textwidth]{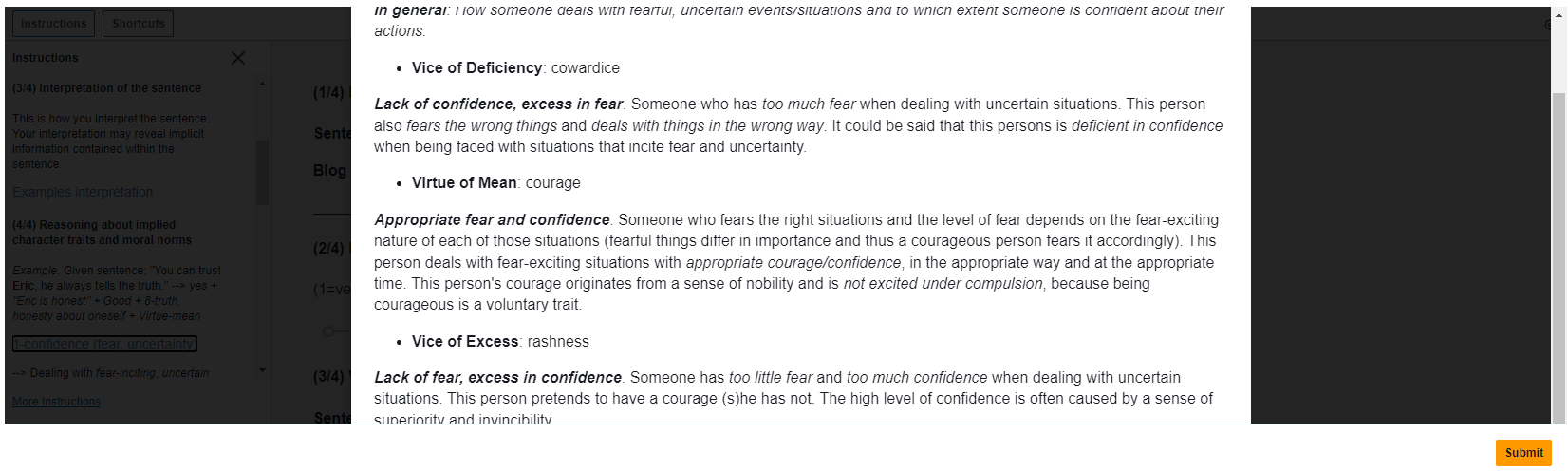}
         \caption{Definition of \textit{Confidence (fear, uncertainty)}.}
         \label{fig:entity-confidence-definition}
     \end{subfigure}
    \caption{Definitions of two Spheres of Action that appear when clicking on the clickable blue phrases \textit{1-confidence (fear, uncertainty)} (a) and \textit{3-giving (and taking - small amounts of money} (b).}
    \label{fig:entity-soa-degree}
\end{figure*}

\paragraph{\textit{Instructions}: Virtue Ethics}
We adopt the well-established Aristotelian Ethics \citep{hursthouse1999virtue}, more specifically Nichomachean Ethics, to label people's moral virtues and vices. Aristotle argued that a virtue/vice is a character trait or a state of someone's character/personality. A Virtue of trait causes a person to act in a manner that is appropriate and desired by society in a given context. It is not because someone acts in a Virtue of manner that that person is actually virtuous. It is important that the action is voluntary and within the control of the individual. By contrast, a vicious trait causes a person to act in a manner that is frowned upon, disapproved or even forbidden by society. 

Aristotle identified several \textbf{Spheres of Action (SoA)} that cause people to act in various ways. 
Those different kinds of actions are put on an axis according to their degree or \textbf{contextual appropriateness}: deficiency, mean and excess. Aristotle stated that a Virtue of person acts in a moderate and contextually appropriate manner (= mean), while a morally faulty person resides to morally deficient (= deficiency) or exceeding actions (= excess). We therefore speak of \textbf{Vice of Deficiency}, \textbf{Virtue of Mean}, and \textbf{Vice of Excess}. Table \ref{tab:soa-and-virtue-vices} presents an overview of the Spheres of Action and the contextual appropriateness labels. We take the descriptions of the SoAs from \citet{hursthouse1999virtue}.

\begin{table}[]
    \small
    \centering
    \begin{tabular}{p{3.2cm}|p{2.4cm}p{2.4cm}p{2.4cm}}
        \toprule
        \textbf{SoA or Feeling} & \textbf{Vice of Deficiency} & \textbf{Virtue of Mean} & \textbf{Vice of Excess} \\
        \midrule
        Confidence (fear, uncertainty) & cowardice & courage & rashness \\
        Pleasures of the body & insensibility & temperance & profligacy \\
        Giving (and taking) - small amounts of money & stinginess & liberality, generosity & prodigality \\
        Giving (and taking) - other, added value & meanness & magnificence & vulgarity \\
        Pride (honour = cause) & little-mindedness & high-mindedness & vanity \\
        Ambition (honour = goal) & lack of ambition & proper ambition & over-ambitiousness \\
        Anger & easy-tempered, spiritlessness & gentleness & hard-tempered, wrathfulness \\
        Pleasure (and pain) of others & cross, contentious & agreeableness, friendliness & flattery \\
        Truth, honesty about oneself & irony & truthfulness & boastfulness \\
        Amusing conversation & boorishness & wittiness & buffoonery \\
        \bottomrule
    \end{tabular}
    \caption{Overview of Spheres of Actions (SoA) with their the virtue and vices.}
    \label{tab:soa-and-virtue-vices}
\end{table}

\paragraph{Confidence (Fear, Uncertainty)}

How someone deals with fearful, uncertain events/situations and to which extent someone is confident about their actions.

\begin{itemize}
    \item \textbf{Cowardice [Vice of Deficiency]} \textit{Lack of confidence, excess in fear}
\\
Someone who has \textit{too much fear} when dealing with uncertain situations when dealing with uncertain situations. This person also \textit{fears the wrong things} and \textit{deals with things in the wrong way}. It could be said that this persons is \textit{deficient in confidence} when being faced with situations that excite fear and uncertainty. 
    \item \textbf{Courage [Virtue of Mean]} \textit{Appropriate fear and confidence}
\\
Someone who fears the right situations and the level of fear depends on the fear-exciting nature of each of those situations (fearful things differ in importance and thus a courageous person fears it accordingly). This person deals with fear-exciting situations with \textit{appropriate} courage/confidence, in the appropriate way, and at the appropriate time. This person's courage originates from a sense of nobility and is \textit{not excited under compulsion} because being courageous is a voluntary trait. 
    \item \textbf{Rashness [Vice of Excess]} \textit{Lack of fear, excess in confidence}
\\
Someone has \textit{too little fear} and \textit{too much confidence} when dealing with uncertain situations. This person pretends to have a courage (s)he has not. The high level of confidence is often caused by a sense of superiority and invincibility.
\end{itemize}

\paragraph{Pleasures of the Body}

How someone deals with and is able to control and fulfil the needs and pleasures of the body.

\begin{itemize}
    \item \textbf{Insensibility [Vice of Deficiency]} \textit{Too little delight in bodily pleasures}
\\
Someone who takes less delight in such pleasures than ought to. This is rather uncommon, almost non-existent as these urges are innate and attached to our animal nature.
    \item \textbf{Temperance [Virtue of Mean]} \textit{Self-control, appropriate desire to fulfill bodily pleasures}
\\
Someone who has temperance and self-control. This person desires moderately, in the appropriate way, at the appropriate time and at the appropriate place. The absence of these pleasant things or the abstinence from them is not painful to a temperate person.
    \item \textbf{Profligacy [Vice of Excess]} \textit{Too much delight in bodily pleasures}
\\
Someone who is a slave to bodily pleasures and loves these delights/pleasures too much. For example, heavy drinking (drinking more than is good for you), obesity, and gluttony. This person also enjoys things that (s)he should not enjoy, in the wrong manner, and more than necessary. The life of a profligate person is led by his/her appetites. Other examples: adultery and outrage.
\\
Example:
\begin{itemize}
    \item \textit{There was \underline{one guy who had a younger brother}. \underline{He} ($\Rightarrow$ one guy who had a younger brother), ahem, tried to take advantage of me when I was drunk.}
\end{itemize}
\end{itemize}

\paragraph{Giving (and Taking) - Small Amounts of Money}

How and to which someone spends their resources (monetary and non-monetary)

\begin{itemize}
    \item \textbf{Stinginess [Vice of Deficiency]} \textit{Lack of giving, excess in taking}
\\
Someone who cares more for wealth than is right, is fond of money. This person takes more than (s)he gives. Examples: greed, meanness, lending money at high interests. The mean person never does anything without thinking twice and always considers how things can be done at the least possible cost. 
    \item \textbf{Liberality, Generosity [Virtue of Mean]} \textit{Spending resources to create added value for others}
\\
Someone who spends their money and other resources on proper things, allocate resources to the right people (generosity), acquires wealth by taking from the proper sources, and desires to reach goals in the right manner (proper amount spent). The liberal person invests in public objects and other people to create added value, which is not always valued in money. This person does go for the cheapest or most expensive way to do things (focus on means), but instead focuses on the goal. 
\\
Example:
\begin{itemize}
    \item \textit{However, \underline{he} has a history of charity that goes far beyond this instance. \underline{He} invested over one billion in a poor region in Russia, transforming the local economy and greatly raising the standard of living there.} (he)
\end{itemize}
    \item \textbf{Prodigality [Vice of Excess]} \textit{Excess in giving (lack in taking)}
\\
Someone who squanders money in riotous, wasteful living. The wasteful person wastes money on things that do not create added value to others (improper objects), or spends too much on small objects. This person also likes to display their wealth to others.  
\end{itemize}

\paragraph{Giving (and Taking) - Other, Added Value}

How someone spends their wealth to create added value and to which things (their worth is not always measurable in terms of money), spending/giving according to their wealth.

\begin{itemize}
    \item \textbf{Meanness [Vice of Deficiency]} \textit{Lack of giving, excess in taking}
    \\
    The mean person likes to do things at the lowest possible cost.
    \item \textbf{Magnificence [Virtue of Mean]} \textit{Spending resources to create added value for others}
    \\
    The magnificent person invests in public objects and other people to create added value, which is not always valued in money. This person does not go for the cheapest or most expensive way to do things (focus on means), but instead focuses on the goal.
    \item \textbf{Prodigality, Wastefulness [Vice of Excess]} \textit{Excess in giving, lack in taking}
    \\
    Someone who squanders money in riotous, wasteful living. The wasteful person wastes money on things that do not create added value to others (improper objects), or spends too much on small objects. This person also likes to display their wealth to others.
\end{itemize}

\paragraph{Pride (Honour = Cause)}

How someone communicates about and acts according to one's honour/pride. 

\begin{itemize}
    \item \textbf{Little-mindedness [Vice of Deficiency]} \textit{Lacks pride, false modesty}
\\
Someone who claims less than (s)he deserves and deprives him/herself of what (s)he deserves. This should not be confused with modesty as this person actively deprives himself of honour (false modesty). This person is too retiring.
    \item \textbf{High-mindedness [Virtue of Mean]} \textit{Proper sense and display of honour}
\\
Someone who claims much but also deserves much. This differs from modesty as a modest person has little to claim about - high-mindedness implies some kind of greatness. The high-minded person also does not seek honour from just everyone (does not look for praise) or on trivial grounds. This person is not easily moved to admiration, does not gossip or speak evil of others. Brooding on the past or blaming others is not something a high-minded person would do.
    \item \textbf{Vanity [Vice of Excess]} \textit{Excess in pride, loves oneself}
\\
Someone who claims much without deserving it. The vain person considers him/herself the better person (sense of superiority) and believes that (s)he deserves greater things. This sense of superiority is unjust and unjustified. This person is ignorant about his/her honour, likes to talk about his/herself, looks down on others and speaks evil of others. At a professional level, they tend to take positions for which they are unfit.
\\
Example:
\begin{itemize}
    \item \textit{There are very low quality \underline{psychologists} who go around trying to tell people what to do and fix their problems but they are not qualified because they have studied a shell major, which is psychology.}
\end{itemize}
\end{itemize}

\paragraph{Ambition (Honour = Goal)}

How someone builds towards more honour, how someone works on becoming a more Virtue of person. 

\begin{itemize}
    \item \textbf{Lack of ambition [Vice of Deficiency]} \textit{Lacks ambition}
\\
Someone who has little desire to improve one's character or condition. Example: laziness. This person also tends to blame others for his/her lack of ambition.
\\
Example:
\begin{itemize}
    \item \textit{\underline{Psychologist} is just someone who cannot man up and study medicine so they go for the sad mockery (...) \underline{People} ($\Rightarrow$ psychologists) study it as a wild card/placeholder major.} (psychologists)
\end{itemize}
    \item \textbf{Proper ambition [Virtue of Mean]} \textit{Healthy ambition}
\\
Someone who is ambitious in the proper manner, in the proper situations and for reaching appropriate goals. 
    \item \textbf{Over-ambitiousness [Vice of Excess]} \textit{Too much ambition}
\\
Someone who desires too much - often at the expense of others. For this person, the goal justifies the means.
\end{itemize}

\paragraph{Anger}

How someone deals with emotions, especially negative emotions such as anger. 

\begin{itemize}
    \item \textbf{Easy-tempered, spiritlessness [Vice of Deficiency]} \textit{Too little anger}
\\
Someone who does not sufficiently deal with their negative emotions, quite emotionless. The easy-tempered person also lacks the spirit to defend him/herself and, more importantly, others. 
    \item \textbf{Gentleness [Virtue of Mean]} \textit{Deals with negative emotions such as anger in a proper way.}
\\
Someone who is angry on the right occasions, with the right persons, in the right manner, for the right reasons and for the right length of time. The gentle person does not lose his/her balance and does not get carried away by emotions. This person is ready to forgive and is not eager to take vengeance.
    \item \textbf{Hard-tempered, wrathfulness [Vice of Excess]} \textit{Too much anger}
\\
Someone who is easily angered by anything and anyone on any occasion. This hard-tempered person's bursts of emotions can be either very short or very long (vengeance and punishment). This person is also prone to violence, vengeance and punishment. (S)he does not easily forgive. The extremes are choleric people.
\\
Example:
\begin{itemize}
    \item \textit{I understand \underline{Putin} is insane.} (Putin)
\end{itemize}
\end{itemize}

\paragraph{Pleasure (and Pain) of Others}

How someone agrees (pleasure of other) or disagrees (pain of other) with his/her interlocutor in any type of interaction. 

\begin{itemize}
    \item \textbf{Cross, Contentious [Vice of Deficiency]} \textit{too much pain of other, disagreement}
\\
Someone who objects to and disagrees with everything and anyone. The cross person does not consider the (emotional) pain they give to people, sets his/her face against everything.
\\
Example:
\begin{itemize}
    \item \textit{``\underline{My mother} took some pictures and posted them on Facebook, where they were quickly filled with horrified comments from \underline{various Americans who} \underline{she is friends with}."} (various Americans who she is friends with)
\end{itemize}
    \item \textbf{Agreeableness, friendliness [Virtue of Mean]} \textit{appropriate mix of agreement and disagreement}
\\
Someone who (dis)agrees when she oughts to and not because of love or hate. The agreeable person fits their behaviour to the context and the person they are talking to. 
    \item \textbf{Flattery [Vice of Excess]} \textit{Too much pleasure of other, agreement}
\\
Someone who pleases people by praising everything and not objecting to anything. This ``people-pleaser" feels the need to avoid discomfort and maximize the pleasure of others. When this person has a goal that leads to this pleasing, then we call it flattery. 
\end{itemize}

\paragraph{Truth, Honesty About Oneself}

How truthful someone is in their interaction with others: what they say and how they say it.

\begin{itemize}
    \item \textbf{Irony [Vice of Deficiency]} \textit{Minimizing truthfulness by disclaiming and depreciating the truth}
\\
Someone who disclaims what (s)he has or depreciates it. This depreciatory way of speaking allows him/her to avoid parade.  The character of ironic person should not be confused with modesty and irony (as a figure of speech). 
    \item \textbf{Truthfulness [Virtue of Mean]} \textit{truthful in word and deed}
\\
Someone who is truthful in word and deed. The truthful person never exaggerates or diminishes the truth, and shuns falsehood as a base thing (does not take pleasure in it). This person is more inclined towards understatement than overstatement of the truth.
\\
Example:
\begin{itemize}
    \item \textit{I also understand more thoughtful \underline{conservatives who voted for Trump}, because regardless of their opinions on \underline{him} ($\Rightarrow$ Trump) as a person, they knew he would support conservative causes, which he did. }(Trump) 
\end{itemize}
    \item \textbf{Boastfulness [Vice of Excess]} \textit{Pretension and twisting the truth}
\\
Someone who likes to pretend to be someone they are not or pretend to hold ideas that others esteem even though they do not. Examples: liars and hypocrites.
\\
Example:
\begin{itemize}
    \item \textit{I will make it very clear, I'm not pro Russian or pro Ukraine, I think both sides are to be blamed for this war and mostly we are seeing only one side of the story from \underline{the western media}.} (the western media)
\end{itemize}
\end{itemize}

\paragraph{Amusing Conversation}

How someone behaves during \textit{amusing} conversation and mixing with others

\begin{itemize}
    \item \textbf{Boorishness [Vice of Deficiency]} \textit{Too boring and serious}
\\
Someone who never says anything laughable, contributes little to the conversation, is too serious and takes everything in ill part. 
    \item \textbf{Wittiness [Virtue of Mean]} \textit{Graceful jest and tact}
\\
Someone who has a lot of tact, jests gracefully, does not ridicule anyone.
    \item \textbf{Buffoonery [Vice of Excess]} \textit{Excess in ridicule}
\\
Someone who does ridicule things just for a laugh, not elegant in their wit, easily resides to offense. The ridicule person likes to make jests of people, vilify them or laugh with them.
\end{itemize}

\subsection{Qualification Test: Round 2} \label{round-2-qualification-test}

Candidate annotators first have to pass a qualification test before they can start the annotation round. They were automatically admitted to the annotation round if they answered all questions or all but one question in the qualification test correctly. The qualification test starts by briefly explaining the three contextual appropriateness labels, called behaviour types in the test for simplicity, after which the annotators have to select the correct description of the types (radio button). It then goes over all Spheres of Action, called situations in the test for simplicity. An example sentence and a person of interest is given for each situation and the annotators are expected to select the best-fitting behaviour type for the person of interest as implied in the sentence (radio button). The sentences present rather clear examples of the behaviour type. We indicate the correct answer with \textbf{(X)}.

Test description: \textit{In this project, you will be asked to describe IMPLIED (= indirectly suggested) CHARACTER TRAITS and classify them given a predefined set of situations and behaviors. We go over the ten different SITUATIONS (= spheres of action), each having three BEHAVIOR types ('vice of deficiency', 'virtue of mean', 'vice of excess').}
\\ \\
(3 BEHAVIOR TYPES) Each situation incites certain behaviors. Those behaviors can either be desired/positive (= virtue) or undesired/negative (= vice).
\\ \\
- A desired behavior (= virtue) is APPROPRIATE to the situation and people involved. It is often a moderate behavior. We therefore classify this type of desired behavior as 'VIRTUE OF MEAN'.
\\ \\
- An undesired behavior (= vice) is disproportionate to the situation and people involved. If it is EXCESSIVE, or 'too much', we call this type of undesired behavior as 'VICE OF EXCESS'. The behavior can also be DEFICIENT, or 'too little'. We then call this undesired behavior 'VICE OF DEFICIENCY'.
\begin{itemize}
    \item Pick the correct description for the three behavior types: VICE OF DEFICIENCY - VIRTUE OF MEAN - VICE OF EXCESS.
    \begin{itemize}
        \item "too much" - appropriate - "too little"
        \item \textbf{(X)} "too little" - appropriate - "too much"
    \end{itemize}
\end{itemize}
(10 SITUATIONS) For each situation (S), we give an example sentence and a person of interest (-- Person). You need to pick the BEST-FITTING behavior type for the person of interest, as IMPLIED in the sentence.
\begin{itemize}
    \item (S1) CONFIDENCE (FEAR, UNCERTAINTY): how someone deals with fear-inciting, uncertain events and the extent to which someone is confident about their actions.
    \begin{itemize}
        \item "They made increasingly reckless investments." -- They
        \begin{itemize}
            \item Vice of Deficiency $\Rightarrow$ cowardice, or "too little" confidence (= too much fear/uncertainty)
            \item Virtue of Mean $\Rightarrow$ courage, or "appropriate" confidence, fear and uncertainty
            \item \textbf{(X)} Vice of Excess $\Rightarrow$ rashness, or "too much" confidence (= too little fear/uncertainty)
        \end{itemize}
    \end{itemize}
    \item (S2) PLEASURES OF THE BODY: how someone deals with and is able to control and fulfill needs and pleasures, especially those of the body.
    \begin{itemize}
        \item "He, ahem, tried to take advantage of me when I was drunk." -- He
        \begin{itemize}
            \item Vice of Deficiency $\Rightarrow$ insensibility, or "too little" delight in bodily pleasures
            \item Virtue of Mean $\Rightarrow$ temperance, or "appropriate" delight (self-control)
            \item \textbf{(X)} Vice of Excess $\Rightarrow$ profligacy, or "too much" delight
        \end{itemize}
    \end{itemize}
    \item (S3) GIVING (AND TAKING) - small amounts of money: how someone spends their wealth on things whose worth are though of in terms of MONEY.
    \begin{itemize}
        \item "Don't ask Carla for any money, she would only lend it to you at large interests." -- Carla
        \begin{itemize}
            \item \textbf{(X)} Vice of Deficiency $\Rightarrow$ stinginess, or "too little" giving (= too much taking)
            \item Virtue of Mean $\Rightarrow$ liberality/generosity, or "appropriate" giving (spending on the appropriate things)
            \item Vice of Excess $\Rightarrow$ prodigality/wastefulness, or "too much" giving (= too little taking; squandering)
        \end{itemize}
    \end{itemize}
    \item (S4) GIVING (AND TAKING) - other, added value: how someone spends their wealth to create added value and to which things (their worth is not always measurable in terms of money), spending/giving according to their wealth.
    \begin{itemize}
        \item "Anna supports the animal shelter in our city financially and volunteers there twice a week." -- Anna
        \begin{itemize}
            \item Vice of Deficiency $\Rightarrow$ meanness, or "too little" giving (little added value)
            \item \textbf{(X)} Virtue of Mean $\Rightarrow$ magnificence, or "appropriate" giving (creating added value)
            \item Vice of Excess $\Rightarrow$ vulgarity, or "too much" giving (making great display on the wrong occasions and in the wrong way, little added value)
        \end{itemize}
    \end{itemize}
    \item (S5) PRIDE (HONOUR = CAUSE): how someone communicates about and acts according to one's honour/pride. Honour is the CAUSE of their behavior.
    \begin{itemize}
        \item "There are very low quality psychologists who go around trying to tell people what to do and fix their problems but they are not qualified." -- psychologists
        \begin{itemize}
            \item Vice of Deficiency $\Rightarrow$ little-mindedness, or "too little" pride (false modesty)
            \item Virtue of Mean $\Rightarrow$ high-mindedness, or "appropriate" pride
            \item \textbf{(X)} Vice of Excess $\Rightarrow$ vanity, or "too much" unjustified pride (sense of superiority)
        \end{itemize}
    \end{itemize}
    \item (S6) AMBITION (HONOUR = GOAL): How someone builds towards more honour and a more virtuous self. Honour is the GOAL of their behavior.
    \begin{itemize}
        \item "John's parents had to force him to look for a job." -- him (= John)
        \begin{itemize}
            \item \textbf{(X)} Vice of Deficiency $\Rightarrow$ lack of ambition, or "too little" ambition
            \item Virtue of Mean $\Rightarrow$ proper ambition, or "appropriate" ambition
            \item Vice of Excess $\Rightarrow$ over-ambitiousness, or "too much" ambition (toxic)
        \end{itemize}
    \end{itemize}
    \item (S7) ANGER: How someone deals with emotions, especially negative emotions such as anger.
    \begin{itemize}
        \item "Serena has a list of everyone who has ever wronged her." -- Serena
        \begin{itemize}
            \item Vice of Deficiency $\Rightarrow$ easy-tempered/spiritlessness, or "too little" anger (emotionless, careless)
            \item Virtue of Mean $\Rightarrow$ gentleness, "appropriate" anger (led by reason, not by emotions)
            \item \textbf{(X)} Vice of Excess $\Rightarrow$ hard-tempered, or "too much" anger (wrathfulness, irritability)
        \end{itemize}
    \end{itemize}
    \item (S8) PLEASURE (AND PAIN) OF OTHERS: general pleasantness in life, or how someone agrees (pleasure of other) or disagrees (pain of other) with his/her interlocutor in any type of interaction.
    \begin{itemize}
        \item "My mother posted some of her poems on Facebook, but they were quickly filled with mean and discouraging comments from many of her so-called friends." -- many of her so-called friends
        \begin{itemize}
            \item \textbf{(X)} Vice of Deficiency $\Rightarrow$ cross/contentious, or "too little" pleasure/agreement (= too much pain)
            \item Virtue of Mean $\Rightarrow$ agreeableness/friendliness, or "appropriate" mix of (dis)agreement
            \item Vice of Excess $\Rightarrow$ flattery, or "too much" agreement (= too little pain; people pleaser, often for own advantage)
        \end{itemize}
    \end{itemize}
    \item (S9) TRUTH, HONESTY ABOUT ONESELF: How truthful someone is in their interaction with others; what they say and how they say it.
    \begin{itemize}
        \item "To me, it sounds weird to think that Jeremy is lying." -- Jeremy
        \begin{itemize}
            \item Vice of Deficiency $\Rightarrow$ irony, or "too little" truthfulness by disclaiming/depreciating the truth (pretense as understatement)
            \item \textbf{(X)} Virtue of Mean $\Rightarrow$ truthfulness, truthful in word and deed
            \item Vice of Excess $\Rightarrow$ boastfulness, "too much" so-called truth (pretense as exaggeration)
        \end{itemize}
    \end{itemize}
    \item (S10) AMUSING CONVERSATION: How someone behaves during amusing conversation and mixing with others.
    \begin{itemize}
        \item "Hannah never says anything remotely funny, she's always standing there with a long face." -- Hannah
        \begin{itemize}
            \item \textbf{(X)} Vice of Deficiency $\Rightarrow$ boorishness, or "too little" amusement (too boring, serious)
            \item Virtue of Mean $\Rightarrow$ wittiness, "appropriate" amusement
            \item Vice of Excess $\Rightarrow$ buffoonery, "too much" amusement (ridicule)
        \end{itemize}
    \end{itemize}
\end{itemize}

\section{Demographic Information on Human Evaluators} \label{appendix:demographics-evaluators}

\begin{itemize}
    \item Gender: female, female
    \item Age: 22, 23
    \item Education level: master, master
    \item Nationality: Belgian, Japanese
\end{itemize}